\documentclass{article}

\usepackage{PRIMEarxiv}

\usepackage[utf8]{inputenc} 
\usepackage[T1]{fontenc}    
\usepackage{url}            
\usepackage{booktabs}       
\usepackage{amsfonts}       
\usepackage{nicefrac}       
\usepackage{microtype}      
\usepackage{xcolor}         
\usepackage{lipsum}
\usepackage{fancyhdr}       
\usepackage{graphicx}       
\graphicspath{{media/}}     
\usepackage{multirow}
\usepackage[linesnumbered,ruled]{algorithm2e}
\usepackage{caption}
\usepackage{makecell}
\usepackage{pifont}

\usepackage{enumitem}
\usepackage{amsmath}
\usepackage{natbib}

\definecolor{linkblue}{RGB}{0,0,255}

\title{REATS: LLM Reasoning-based Ensemble Learning for Adaptive Time Series Forecasting
}

\author{
  Xu Zhang\thanks{Work done by Xu Zhang during his research internship at Microsoft Research Asia.}\\
Fudan University  \\
  Shanghai, China \\
  \texttt{xuzhang@alu.fudan.edu.cn} \\
  \And
  Chang Xu\thanks{Corresponding authors: Chang Xu and Li Zhao.} \\
  Microsoft Research \\
  Beijing, China \\
  \texttt{chanx@microsoft.com} \\
     \And
  Hui Sun \\
  Nankai University\\
  Tianjin, China\\
  \texttt{sunh@nbjl.nankai.edu.cn} \\
    \And    
  Nan Ma \\
  Microsoft Research \\
 Cambridge, USA\\
  \texttt{nama1@microsoft.com} \\
    \And
  Zijian Zhang \\
  Jilin University\\
 Jilin, China\\
  \texttt{zhangzijian@jlu.edu.cn} \\
     \And
      Peng Wang\\
  Fudan University \\
  Shanghai, China \\
  \texttt{pengwang5@fudan.edu.cn} \\
  \And
    Wei Wang\\
Fudan University  \\
  Shanghai, China \\
  \texttt{weiwang1@fudan.edu.cn} \\
  \And
Li Zhao\footnotemark[2] \\
  Microsoft Research \\
  Beijing, China \\
  \texttt{lizo@microsoft.com} \\
}

\usepackage{hyperref}

\hypersetup{
    colorlinks=true,
    linkcolor=linkblue,      
    citecolor=black,      
    urlcolor=magenta     
}

\begin{document}
\maketitle

\begin{abstract}
  Recent advances in time series forecasting (TSF) have led to the development of numerous powerful models, each characterized by distinct design principles and strengths in capturing temporal dynamics. However, due to the diversity of real-world time series, relying on a single model often makes it difficult to effectively handle the complexity patterns exhibited by different samples. Ensemble learning offers a promising solution by combining the complementary strengths of multiple forecasting models.
  However, existing methods rely on fixed rules or black-box models based solely on numerical inputs, failing to leverage LLM reasoning to improve ensemble effectiveness or explain weighting decisions.
In this paper, we propose REATS, which leverages LLM reasoning capabilities as an intelligent ensemble router that jointly processes textual temporal pattern descriptions and numerical features to produce interpretable, sample-adaptive ensemble weights through chain-of-thought reasoning. To enable effective LLM-based ensembling, we study its key design choices and propose: (i) a structured input pipeline that transforms raw time series into hybrid textual numerical representations  with fixed token cost regardless of input length, facilitating LLM reasoning and enabling rule-based chain-of-thought construction without API dependency, augmented with retrieved similar-sample priors; 
(ii) a diverse multi-row weight supervision scheme to enrich training signals, coupled with a token-efficient percentage-table format that reduces numerical complexity and mitigates LLM hallucinations; and (iii) a two-stage fine-tuning framework combining SFT for structured reasoning acquisition and GRPO with a reciprocal reward mapping that transforms the continuous unbounded reward range into bounded signals with amplified near-oracle sensitivity, addressing the uniform sensitivity and outlier-dominated advantage compression in mixed-quality rollout groups inherent in naive reward designs for regression-based GRPO. Experiments on eight benchmarks demonstrate that REATS outperforms competitive ensemble baselines on foundation model and small model candidate groups respectively, while providing natural language explanations for its decisions and demonstrating strong transfer learning capability and out-of-domain generalization to unseen candidate models. The code is available at the link \url{https://github.com/microsoft/TimeCraft/tree/main/REATS}.
\end{abstract}

\section{Introduction}
Time series forecasting has wide applications in fields such as energy management~\cite{uremovic2022new}, financial analysis~\cite{zhang2025multi}, and transportation planning~\cite{wachs1987forecasts}. In recent years, a large number of deep learning-based forecasting models have been proposed, including methods based on different architectures such as Transformers~\cite{zhang2025multi}, linear models~\cite{liu2022combined,zhang2025lightweight}, and convolutional networks~\cite{zhu2023drcnn}. 
Different models are designed with distinct approaches to address various characteristics of time series. 
Intuitively, no single model can consistently achieve optimal performance across all datasets and forecasting scenarios. 
However, essentially, they all attempt to use a single modeling strategy to handle all samples, which limits the forecasting performance.
Fortunately, ensemble learning can adequately address this issue to further improve the forecasting accuracy, with its core challenge being how to dynamically assign appropriate model weights for different time series samples.

Traditional ensemble methods typically adopt fixed weights or static strategies based on validation errors~\cite{chen2022rrmse,gruber2015ensemble}, making it difficult to adapt to the complex and dynamic temporal patterns in time series data. Some studies attempt to train small neural networks to achieve dynamic weight allocation~\cite{fu2022reinforcement}, but such approaches can only utilize numerical inputs without leveraging textual semantic understanding, limiting the expressiveness of their ensemble decisions. Moreover, they need to be retrained when the candidate models change, resulting in limited flexibility. Table~\ref{tab:ensemble_comparison} summarizes the key differences among existing ensemble paradigms and our proposed LLM-based approach.


\begin{table}[h!]
\centering
\scriptsize
\caption{Design-property comparison of ensemble-routing paradigms for time series forecasting. Check marks indicate whether the method explicitly supports the property by design.}
\label{tab:ensemble_comparison}
\setlength{\tabcolsep}{4pt}

\begin{tabular}{l|cccc}
\toprule
\textbf{Method} & \textbf{Multimodal input} & \textbf{Sample-adaptive} & \textbf{Unseen model generalization} & \textbf{Human-readable reasoning} \\
\midrule
Static ensemble & \ding{55} & \ding{55} & \ding{55} & \ding{55} \\
NN-based ensemble & \ding{55} & \ding{51} & \ding{55} & \ding{55} \\
\textbf{LLM ensemble (REATS)} & \ding{51} & \ding{51} & \ding{51} & \ding{51} \\
\bottomrule
\end{tabular}

\end{table}

The rapid development of Large Language Models (LLMs) provides a new perspective for addressing these limitations. Compared with existing ensemble approaches, LLMs offer three key advantages for time series ensembling: (1) \textbf{Multimodal information fusion}: LLMs can integrate textual descriptions of temporal patterns (such as trends, seasonality, and volatility) with numerical data for comprehensive reasoning, unlike neural networks limited to numerical inputs. (2) \textbf{Flexible scalability}: by modifying model descriptions in the prompts, candidate models can be flexibly adjusted without retraining. (3) \textbf{Inherent interpretability}: LLMs can generate natural language explanations clarifying why certain models are more suitable for the given temporal characteristics, facilitating human-in-the-loop collaboration.

However, applying LLMs to time series ensemble learning poses three key challenges: (1) raw time series incur token costs that scale linearly with length, requiring a fixed-budget representation that preserves temporal semantics; (2) constructing scalable and informative supervision is challenging: relying on LLM API calls to generate CoT for massive samples is prohibitively expensive and demands a scalable alternative, while a single optimal weight vector provides limited supervision diversity and less informative training signal feedback; and (3)  typical GRPO applications~\cite{shao2024deepseekmath} employ verifiable binary or bounded rewards (e.g., correct/incorrect), which are inadequate for continuous MSE-based optimization, where the naive mapping $r{=}-\delta$ provides uniform sensitivity that struggles to discriminate near-optimal candidates, and its unbounded range allows a single outlier generation to dominate group variance, compressing the reward gaps among the remaining good candidates.  Crucially, this compression persists in GRPO's advantage normalization: since the outlier inflates the group standard deviation $\sigma_r$, the relative spacing among near-optimal candidates collapses, requiring a nonlinear bounded mapping that reshapes the within-group reward geometry.

Based on the above insights, we propose REATS: LLM Reasoning-based Ensemble Learning for Adaptive Time Series Forecasting. Our contributions are as follows:

\begin{itemize}[leftmargin=*]
   \item We propose an ensemble learning framework, REATS, that leverages the reasoning capabilities of LLMs for TSF{, and study the key techniques that make this paradigm effective: how to represent time series for an LLM router, how to construct reasoning and weight supervision at scale, and how to adapt GRPO to a continuous forecasting objective}. The framework includes a systematic sample construction pipeline (e.g., retrieval-augmented prior knowledge, diverse weight supervision, and token-efficient weight table format) and a two-stage fine-tuning process, achieving strong performance in both accuracy and interpretability. 

    \item We design a hybrid textual\textendash numerical input representation with fixed token cost that activates semantic reasoning and enables rule-based CoT construction without API dependency. The rule-based CoT significantly enhances both in-domain and out-of-domain ensemble performance, demonstrating that expensive API-generated CoT is not always necessary for effective LLM fine-tuning. Besides, the designed multi-row weight supervision enriches fine-tuning signals during SFT and GRPO stages.

    \item  We design a reciprocal reward mapping function that adapts GRPO to continuous regression tasks by addressing two limitations of the naive $r{=}-\delta$: uniform sensitivity that struggles to discriminate near-optimal candidates, and unbounded range that allows outlier-dominated advantage compression in mixed-quality rollout groups. The resulting bounded, nonlinear mapping enables effective late-stage policy refinement.  This reward-level solution outperforms several recent GRPO algorithmic variants that target reward sparsity through optimization modifications.

    \item  Experiments on eight benchmarks show that REATS consistently surpasses the competitive baselines while demonstrating robust OOD generalization and enhanced transfer learning.

\end{itemize}

\section{Related work}

We focus our discussion on ensemble learning for time series forecasting, as it is most relevant to our work. A comprehensive review of individual forecasting models is provided in Appendix~\ref{sec:related_tsf}.

Recent ensemble learning research has increasingly focused on NLP and LLM-centered scenarios~\cite{lv2024specfuse,yun2025ensemble}, e.g., multiple LLM-generated textual responses are ranked or fused to improve generation quality~\citep{jiang2023llm}.
Such text-oriented ensemble mechanisms are difficult to directly apply to time series forecasting, where candidate models output continuous numerical trajectories rather than natural language responses.

For numerical forecasting models, ensemble learning still largely relies on traditional strategies.
The first is \textit{traditional machine learning-based methods}, which adopt fixed or static weighting strategies based on training or validation performance~\cite{bertsimas2023ensemble,chen2022rrmse,gruber2015ensemble}, such as inverse-error weighting or uniform averaging. These methods are simple and efficient but lack adaptability to varying input patterns. The second is \textit{neural network based methods}, which employ neural networks to extract temporal features and leverage reinforcement learning algorithms to learn ensemble weights through forecasting-oriented reward signals~\cite{fu2022reinforcement}, offering dynamic adaptability to input-specific patterns.
However, despite the demonstrated reasoning capabilities of LLMs in various domains, their potential for ensemble learning in time series forecasting has yet to be fully investigated, leaving considerable room for integrating powerful reasoning abilities into adaptive model combination.

\section{LLM Reasoning-based Ensemble Learning
for Adaptive Time Series Forecasting}

\subsection{Problem definition}

Given a univariate historical time series $\mathbf{X} = [x_1, x_2, \ldots, x_T]$ and forecast horizon $H$, a set of $N$ candidate base forecasting models $\mathcal{M} = \{M_1, M_2, \ldots, M_N\}$ each produces a prediction $\hat{\mathbf{Y}}_i = M_i(\mathbf{X}) = [\hat{y}^i_{T+1}, \ldots, \hat{y}^i_{T+H}]$. The final ensemble prediction is obtained by:
\begin{equation}
\hat{\mathbf{Y}} = \sum_{i=1}^{N} w_i \cdot \hat{\mathbf{Y}}_i, \quad \text{s.t.} \sum_{i=1}^{N} w_i = 1, \ w_i \geq 0
\end{equation}
where $\mathbf{w} = [w_1, w_2, \ldots, w_N]$ denotes the ensemble weights. Our goal is to fine-tune an LLM that, given a structured prompt constructed from temporal pattern features, retrieved similar series, and candidate model descriptions, can produce ensemble weights through explicit chain-of-thought reasoning to minimize the forecasting error.

\subsection{Overview}

\begin{figure*}[bt]
\centerline{\includegraphics[width=\linewidth]{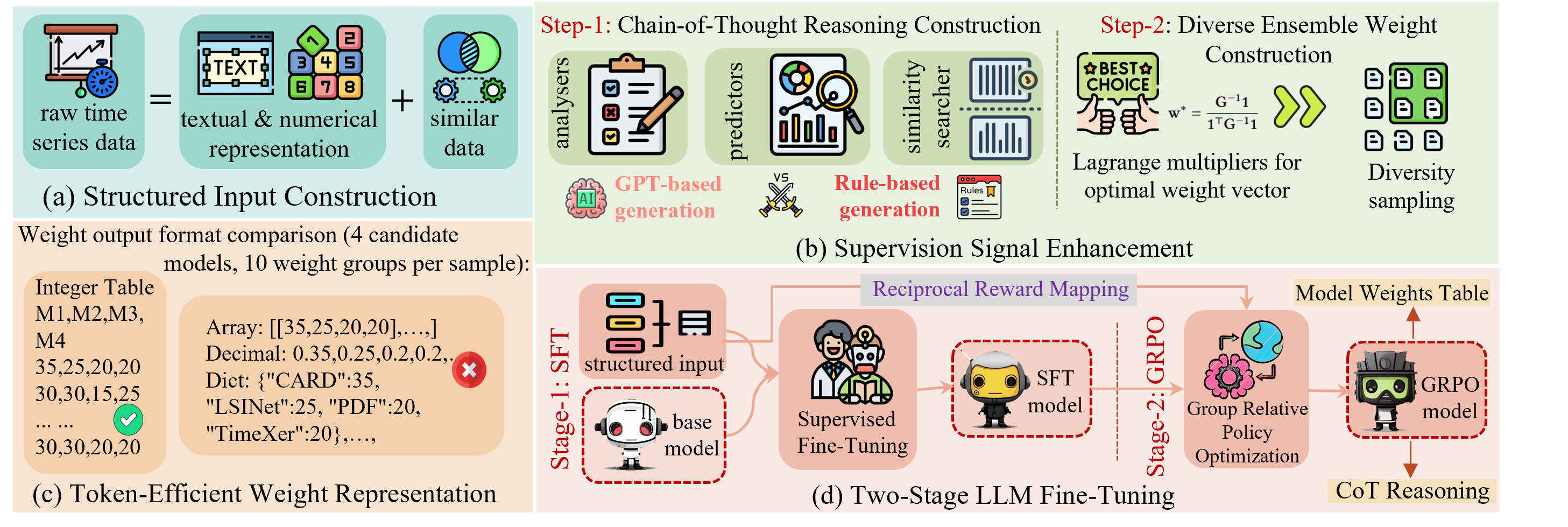} }
\caption{Overall framework of REATS.}
\label{fig:frame_work}
\end{figure*}

Figure~\ref{fig:frame_work} illustrates the overall framework of REATS, which consists of four design components:
\textbf{(1) Structured Input Construction} transforms raw time series into hybrid textual--numerical descriptions augmented with retrieved similar-sample priors.
\textbf{(2) Supervision Signal Enhancement} constructs chain-of-thought reasoning and diverse weight supervision.
\textbf{(3) Token-Efficient Weight Representation} formats weights as integer percentage tables.
\textbf{(4) Two-Stage Fine-Tuning} performs SFT for structured reasoning, followed by GRPO with a reciprocal reward mapping that extends it to unbounded regression task for direct forecasting optimization.
Components (1)--(3) are detailed in Section~\ref{sec:input} and component (4) in Section~\ref{sec:training}.

\subsection{Structured LLM finetuning data construction}
\label{sec:input}
\begin{figure*}[bt]
\centerline{\includegraphics[width=\linewidth]{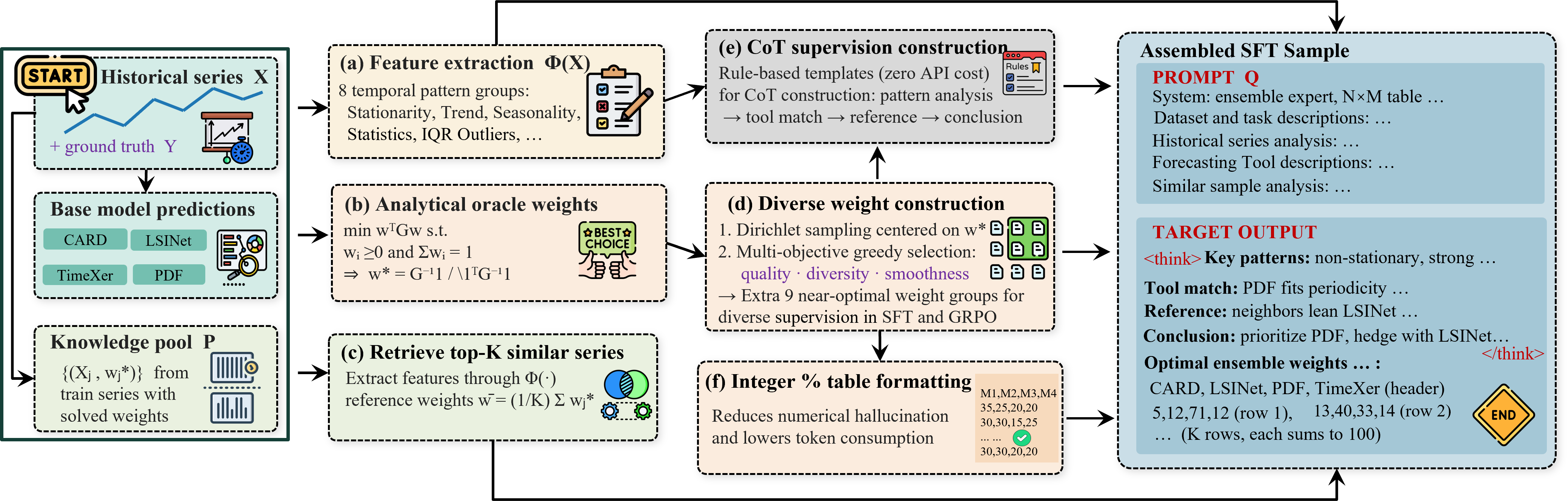} }
\caption{Structured LLM fine-tuning data construction pipeline.}
\label{fig:data_construc}
\end{figure*}

Figure~\ref{fig:data_construc} illustrates the overall data construction pipeline, covering input construction (feature extraction and retrieval augmentation) and output construction (CoT reasoning, diverse weights, and weight table format).

\noindent\textbf{Hybrid textual–numerical feature extraction.}
How to effectively represent time series for LLM-based reasoning is a key design choice. Directly tokenizing raw values as text causes token counts to grow linearly with series length (e.g., 1656$\to$7480 from length 96 to 512) and increases hallucination risk, while encoding series through a separate MLP projector sacrifices interpretability and OOD generalization (Table~\ref{tab:input_ablation_small}).
We transform raw time series into structured textual--numerical descriptions rather than feeding high-precision numerical sequences directly into LLMs. Given a historical time series $\mathbf{X}$, we extract eight groups of temporal pattern features $\Phi(\mathbf{X}) = \{\phi_{\text{sta}},\allowbreak \phi_{\text{noise}},\allowbreak \phi_{\text{trend}},\allowbreak \phi_{\text{sea}},\allowbreak \phi_{\text{ac}},\allowbreak \phi_{\text{stat}},\allowbreak \phi_{\text{out}},\allowbreak \phi_{\text{dist}}\}$, covering stationarity, noise, trend, seasonality, autocorrelation, statistical properties, IQR-based outliers, and distribution (details in Appendix Table~\ref{tab:ts_statistics_summary_full}).
Each feature group uses a two-level structure: an outer group header (e.g., ``[trend]'') followed by inner key-value pairs (e.g., ``Trend slope: -0.00454'', ``Trend strength: Strong''), preserving numerical information while providing semantic interpretations. All values are rounded to three significant figures to reduce token consumption.
Our hybrid representation addresses both issues:
\textbf{(1)} the token count is fixed regardless of input length, activating the LLM's textual reasoning to improve ensemble effectiveness;
\textbf{(2)} the structured format supports rule-based CoT construction, where each feature group serves as an explicit reasoning anchor, achieving comparable quality to GPT-generated CoT without API dependency (Table~\ref{tab:cot_combined}(a)).

\noindent\textbf{Retrieval-augmented prior knowledge construction.}
We construct a knowledge pool $\mathcal{P} = \{(\mathbf{X}_j, \mathbf{w}_j)\}_{j=1}^{|\mathcal{P}|}$ from training set samples only, where $\mathbf{w}_j \in \mathbb{R}^N$ stores the optimal ensemble weights. Given a query series $\mathbf{X}$, we retrieve the top-$K$ most similar samples (excluding the query itself during training to prevent label leakage):
\begin{equation}
\mathcal{R}(\mathbf{X}) = \underset{j \in \mathcal{P}}{\mathrm{arg\,min\text{-}K}} \; \|\mathbf{X} - \mathbf{X}_j\|_2
\end{equation}
The retrieved samples provide a reference weight distribution $\bar{\mathbf{w}} = \frac{1}{K}\sum_{j \in \mathcal{R}} \mathbf{w}_j$ and are averaged at the raw series level into a single prototype, which is then converted into the same hybrid representation as the query. This aggregation avoids K-fold token expansion while capturing representative temporal patterns of the local neighborhood. The retrieved prior knowledge grounds the LLM's reasoning in empirical model performance, strengthening its reasoning quality and improving ensemble accuracy.

\noindent\textbf{Chain-of-Thought reasoning construction.}
\label{sec:supervision}
We construct CoT supervision signals following a four-step format: (1) identify dominant temporal features; (2) analyze each candidate model's suitability; (3) interpret retrieved similar-sample weights; (4) justify the final weight allocation. The CoT is wrapped in \texttt{<think>...</think>} followed by ensemble weights, forming a ``reason-then-decide'' output. Note that Rule-CoT is constructed only during training-time data preparation using training set oracle weights. At inference time, the fine-tuned LLM autonomously generates its own CoT reasoning and weights from input features alone, without access to any oracle information. Instead of relying on expensive GPT API calls to generate CoT, we propose a lightweight rule-based generation approach that achieves comparable quality without API dependency (validated in Table~\ref{tab:cot_combined}(a)). The process follows: known oracle weights $\rightarrow$ reverse-engineer plausible explanations via rule-matching between temporal features and tool category (candidate model) strengths $\rightarrow$ generate natural language through template variants.

\noindent The overview is given in Algorithm~\ref{alg:rule_cot}, and the detailed version with category-specific rule examples is provided in Appendix Algorithm~\ref{alg:rule_cot_detail}. A complete generated CoT example is shown in Appendix~\ref{sec:sft_example}  ``Model output'' Part.

\begin{algorithm}[t]
\caption{Rule-Based Chain-of-Thought Generation (Detailed in the Appendix Algorithm~\ref{alg:rule_cot_detail})}
\label{alg:rule_cot}

\KwIn{Temporal features $\Phi(\mathbf{X})$, candidate models $\mathcal{M}$ (termed \emph{tools} in the prompt) with model category labels, oracle weights $\mathbf{w}^*$, RAG reference $\Phi(\mathbf{X}_{\text{rag}})$, $\mathbf{w}_{\text{rag}}$}
\KwOut{CoT text}

$\bar{\mathbf{w}} \gets$ average of $K'$ oracle weight rows\;

\tcp{Step 1: Key patterns}
$s_1 \gets$ template phrases for each parsed attribute of $\Phi(\mathbf{X})$\;

\tcp{Step 2: Oracle-guided tool matching}
\ForEach{tool $t_i$ sorted by $\bar{w}_i$ descending}{
    Match tool category against temporal attributes via rules\;
    Select fit-level template guided by $\bar{w}_i$: high ($>$0.35) / mid ($>$0.15) / low\;
}
$s_2 \gets$ join tool-match sentences\;

\tcp{Step 3: RAG reference comparison}
$s_3 \gets$ template noting agreement/divergence between $\bar{\mathbf{w}}$ and $\mathbf{w}_{\text{rag}}$\;

\tcp{Step 4: Allocation conclusion}
$s_4 \gets$ conclusion template naming dominant/moderate/minor tools\;

\Return $s_1 \oplus s_2 \oplus s_3 \oplus s_4$\;
\end{algorithm}

\noindent\textbf{Diverse ensemble weight construction.}
\label{sec:supervision_diverse_w}
We first derive the optimal (oracle) weight vector $\mathbf{w}^*$ by minimizing ensemble MSE $= \mathbf{w}^\top \mathbf{G} \mathbf{w}$ (where $G_{ij} = \frac{1}{T}\sum_t e_{i,t} e_{j,t}$) subject to $\sum_i w_i = 1$ and $w_i \geq 0$. Without the non-negativity constraint, the closed-form solution is:
\begin{equation}
\label{equ:oracle_w}
    \mathbf{w}^* = \frac{\mathbf{G}^{-1} \mathbf{1}}{\mathbf{1}^\top \mathbf{G}^{-1} \mathbf{1}}
\end{equation}
The full derivation is shown in Appendix~\ref{app:oracle_derivation}.
In practice, we solve the full simplex-constrained quadratic program $\mathbf{w}^* = \arg\min_{\mathbf{w} \geq 0,\, \mathbf{1}^\top\mathbf{w}=1} \mathbf{w}^\top (\mathbf{G} + \epsilon \mathbf{I}) \mathbf{w}$ to obtain the exact non-negative oracle weights.
To avoid overfitting to a single supervision target, we further generate 9 diverse weight vectors via Dirichlet sampling and multi-objective greedy selection, combined with the oracle (placed as the first row) to form $K'{=}10$ rows per sample (Appendix~\ref{app:diverse_weight}). The diverse rows serve as training-time auxiliary tasks. (1) During SFT, producing multiple valid allocations forces the CoT to capture broader temporal pattern understanding rather than memorizing a single solution. (2) During GRPO, they provide additional reward signals that mitigate reward sparsity and encourage diverse exploration in weight space.

\noindent\textbf{Token-efficient weight representation.}
\label{sec:weight_format}
We represent weights as integer percentages formatted as a compact comma-separated table, e.g., \texttt{CARD,SEMixer,TimeXer,LSINet} (header) \texttt{/ 35,25,20,20} (row 1) \texttt{/ 30,30,15,25} (row 2).
This reduces numerical hallucination and lowers token consumption (empirically validated in Table~\ref{tab:ablation_small}(b) and (c)).

\subsection{Two-Stage Fine-Tuning with SFT and GRPO}
\label{sec:training}

We adopt a two-stage strategy to train a lightweight LLM (1.7B parameters): supervised fine-tuning (SFT) for structured reasoning imitation, followed by Group Relative Policy Optimization (GRPO) to directly optimize forecasting MSE.

\noindent\textbf{Stage 1: Supervised Fine-Tuning.}
We fine-tune the LLM using next-token prediction on the constructed dataset. Each sample pairs an input prompt $Q$ with a target output: a CoT block in \texttt{<think>...</think>} tags followed by a $K'$-row weight table. SFT teaches the model to (1) follow the ``reason-then-decide'' format, (2) produce pattern-aware reasoning, and (3) learn coarse weight allocations by imitating the oracle weight distribution. However, token-level loss does not directly optimize downstream MSE, leaving the weights suboptimal.

\noindent\textbf{Stage 2: Group Relative Policy Optimization.}
We further fine-tune using GRPO~\cite{shao2024deepseekmath} with a composite reward. For each prompt, the policy generates $G$ candidate outputs, each parsed into $K'$ weight rows. The reward consists of:

\noindent\textit{Relative MSE Reward.} Let $\mathcal{L}(\mathbf{w})$ denote the ensemble forecasting MSE under weight $\mathbf{w}$, and $\delta_i = \mathcal{L}(\mathbf{w}_i) - \mathcal{L}(\mathbf{w}_i^*)$ be the MSE gap between the $i$-th predicted row and its oracle.

\noindent\textit{Oracle Proximity Reward.} Computes weight-space MSE between predicted and oracle vectors, serving as auxiliary gradient when MSE signals are sparse.

\noindent\textit{Format Penalty.} A negative penalty activated only upon format violations (negative weights or sum $\neq$ 100\%).

Standard GRPO applications (e.g., math, code) use binary or bounded rewards (correct/wrong) that naturally provide well-separated signals for advantage computation. However, our task produces a continuous MSE gap $\delta \geq 0$ as the raw signal. The most direct adaptation, $r = -\delta$, is unbounded: a single outlier generation with large $\delta$ inflates the group standard deviation $\sigma_r$, causing the normalized advantages $\hat{A}_i{=}(r_i{-}\mu_r)/\sigma_r$ among near-optimal candidates to collapse to near-identical values, preventing GRPO from distinguishing better from best. These issues require a dedicated reward mapping (Section~\ref{sec:reward_discussion}).
While recent GRPO variants (e.g., DAPO~\cite{yu2026dapo}, DrGRPO~\cite{liuunderstanding}, GSPO~\cite{zheng2025group}, SAPO~\cite{gao2025soft}) attempt to mitigate reward sparsity through modifications to the policy optimization procedure (e.g., reward normalization, sample filtering, and soft advantage computation), they do not address the reward signal itself and thus remain limited when the underlying reward is continuous and unbounded (see Section~\ref{sec:reward_discussion}).

We adopt a reciprocal mapping $r = 1/(1 + k\delta)$, which compresses $\delta$ into $[0,1]$ while concentrating sensitivity near the oracle: in early training it distinguishes ``poor'' from ``worse'' (avoiding reward collapse when all generations are far from oracle), and in later training it distinguishes ``good'' from ``better'' (avoiding saturation when generations converge near the oracle). Crucially, because the reciprocal is a \emph{nonlinear} bounded transformation, it reshapes the relative spacing of rewards within a group, compressing outlier gaps while amplifying near-oracle differences in mixed-quality rollout groups, and this geometric change in relative reward positions is preserved after advantage normalization $\hat{A}_i{=}(r_i{-}\mu_r)/\sigma_r$, unlike a simple linear rescaling which would be canceled. Detailed discussions and comparisons with alternative mappings are in Section~\ref{sec:reward_discussion} and Appendix~\ref{sec:reward_mapping_appendix}--\ref{sec:naive_reward_analysis}.

The overall reward is:
\begin{equation}
    r = \lambda_1 \cdot r_{\text{mse}} + \lambda_2 \cdot r_{\text{agg}} + \lambda_3 \cdot r_{\text{oracle}} + p_{\text{format}}
\end{equation}
where $r_{\text{mse}} = \frac{1}{K'}\sum_{i=1}^{K'} \frac{1}{1 + k \cdot \delta_i}$, $r_{\text{oracle}} = \frac{1}{K'}\sum_{i=1}^{K'} \frac{1}{1 + k \cdot \text{MSE}(\mathbf{w}_i, \mathbf{w}_i^*)}$, and $r_{\text{agg}} = \frac{1}{1 + k \cdot \mathrm{MSE}(\sum_j \bar{w}_j \hat{\mathbf{y}}_j, \mathbf{y})}$ with $\bar{\mathbf{w}} = \frac{1}{K'}\sum_i \mathbf{w}_i$.

This two-stage design is complementary: SFT provides a strong initialization with structured reasoning, format compliance, and coarse ensemble weights, while GRPO further refines the weight allocation by directly optimizing forecasting accuracy through MSE reward-driven exploration beyond the imitation targets.

\section{Experiments and results}

\subsection{Experimental settings}

\noindent\textbf{Datasets and metrics.}
We evaluate on eight benchmarks~\cite{zhang2026semixer,nie2022time_patchformer,zhang2025lightweight}: ETTh1/h2, ETTm1/m2, Exchange, Weather, Electricity, and Traffic, covering energy, finance, meteorology, and transportation domains. All experiments are conducted on standard univariate time series forecasting, and both input and prediction length are 96, with train/val/test split ratio of 7:1:2. We report MSE and MAE. Dataset statistics are summarized in Appendix Table~\ref{tab:dataset_stat}.

\noindent\textbf{Candidate model pool for ensemble.}
We construct a diverse pool of forecasting models spanning two categories: (1) \textit{Small specialized models}: TimeXer~\cite{wangtimexer}, LSINet~\cite{zhang2025lightweight}, CARD~\cite{zhou2024card}, TimeMixer~\cite{wangtimemixer}, ModernTCN~\cite{luo2024moderntcn}, SEMixer~\cite{zhang2026semixer}, DLinear~\cite{zeng2023transformers_linear}, PDF~\cite{dai2024periodicity}, PatchTST~\cite{nie2022time_patchformer}, and MLF~\cite{zhang2025multi}; (2) \textit{Foundation models}: MOMENT~\cite{goswamimoment}, Sundial~\cite{liu2025sundial}, Timer~\cite{liutimer}, TIME-MOE~\cite{shi2025time}, TimesFM~\cite{dasdecoder}, MOIRAI~\cite{woo2024unified}, TimerXL~\cite{liu2024timerXL} and Chronos~\cite{ansari2024chronos}. These models also serve as individual baselines.  The corresponding pre-trained weight sources are provided in Appendix~\ref{sec:model_descriptions}.

\noindent\textbf{Ensemble baselines.}
We compare against four categories of ensemble methods: (1) \textit{Heuristic}: uniform averaging and random weighting; (2) \textit{Error-based}: inverse MSE weighting and optimal fixed weighting (OptW), where OptW is obtained by solving the same non-negative simplex-constrained quadratic program used for REATS oracle construction. The subscripts tr/val indicate whether the fixed weights are estimated on the training or validation set and then applied to all test samples; (3) \textit{Neural network-based}: RLMC~\cite{fu2022reinforcement}, which follows its original setup using raw historical series while sharing the same data splits, candidate forecasts, and oracle supervision as REATS; (4) \textit{Prompt-based}: LLM zero-shot ensemble without fine-tuning (GPT-5.2/5.5, Codex, DeepSeek-V3.2, Grok-4), using the same prompt components (temporal features, tool descriptions, and RAG references) and reason-then-decide output format as REATS for fair comparison. Additionally, to validate our reward mapping design, we compare against four recent GRPO algorithmic variants (DAPO~\cite{yu2026dapo}, DrGRPO~\cite{liuunderstanding}, GSPO~\cite{zheng2025group}, SAPO~\cite{gao2025soft}) that address reward sparsity through optimization-level modifications (Section~\ref{sec:reward_discussion}).

\noindent\textbf{Implementation details.}
We use Qwen3-1.7B as the base LLM. Retrieval uses $K{=}3$ similar samples, with $K'{=}10$ weight rows per sample (1 oracle + 9 diverse). Default ensemble size is 4 candidates (scalability to 2/6/8 also evaluated).
For the reciprocal reward mapping, we set $k{=}20$ (chosen to jointly ensure non-vanishing far-range rewards and strong near-oracle discrimination; see Appendix~\ref{sec:reward_mapping_appendix}), $(\lambda_1, \lambda_2, \lambda_3){=}(0.8, 0, 0.2)$, and format penalty $-0.5$. At inference, only the first output row is used as the final ensemble weights, since it is trained against the QP oracle and receives the strongest optimization pressure. Training is on NVIDIA A100 GPUs. All oracle weights used for SFT supervision and GRPO reward computation are derived from training set ground truth only.


\subsection{Comparison of ensemble forecasting performance}

\noindent\textbf{Foundation model candidates.}
As shown in Table~\ref{tab:main_res_combined}(a), no single foundation model dominates across all datasets, confirming the necessity of ensemble learning. Zero-shot LLM ensembles (e.g., Codex: 0.1594) perform comparably to the best traditional baselines (OptW$_{\text{val}}$: 0.1597). REATS-GRPO achieves the lowest average MSE of \textbf{0.1384}, reducing error by 13.3\% over OptW$_{\text{val}}$ and winning on all eight datasets. The consistent SFT$\to$GRPO gain (0.1455$\to$0.1384) confirms that MSE-based reward effectively refines weights beyond imitation learning.

\noindent\textbf{Small model candidates.}
Table~\ref{tab:main_res_combined}(b) presents a more challenging setting where individual models exhibit extreme variance (e.g., TimeXer avg MSE 1.6664), causing naive averaging (0.3173) to underperform the best single model. Notably, zero-shot LLM ensembles (best: Codex at 0.1709) fall behind traditional baselines (OptW$_{\text{tr}}$: 0.1352), indicating that without fine-tuning, LLMs struggle to handle highly heterogeneous candidates. REATS-GRPO achieves \textbf{0.1080}, reducing error by 20.1\% over OptW$_{\text{tr}}$, demonstrating that fine-tuned LLM reasoning can effectively identify complementary strengths even among weak and unreliable models.

\noindent\textbf{Statistical significance.} REATS-GRPO achieves the lowest average MSE in both model groups, with per-dataset win rates statistically significant under the sign test ($p{<}0.05$).

\noindent\textbf{Generalization to unseen candidate models.}
We train REATS on small model candidates and evaluate on unseen foundation models (Table~\ref{tab:ood_foundation}). In OOD settings, the RAG knowledge pool and candidate descriptions are reconstructed for the new candidate models using their training-set predictions, without access to test labels. REATS-GRPO achieves \textbf{0.1442}, outperforming the best traditional baseline (OptW$_{\text{val}}$: 0.1564) by 7.8\% and the best zero-shot LLM (DeepSeek-V3.2: 0.1626) by 11.3\%. REATS also consistently achieves the best overall performance across additional OOD settings reported in the appendix, including generalization to mixed candidates from both small and foundation model pools (Table~\ref{tab:ood_mixed}) and within-group generalization in small-to-unseen-small and foundation-to-unseen-foundation settings (Table~\ref{tab:ood_within_combined}). These results suggest that REATS learns transferable reasoning about temporal characteristics rather than memorizing specific model identities.

\noindent\textbf{Generalization to different numbers of candidate models.}
We evaluate REATS with N=2, 4, 6, and 8 foundation model candidates. As shown in Figure~\ref{fig:scalability}, REATS generally achieves the lowest average MSE across all settings, with the advantage tending to increase as more candidates are available.

\begin{table*}[t]
    \centering
    \caption{Ensemble forecasting results (MSE $\downarrow$).  Dataset abbreviations: Exch=Exchange, H1/H2=ETTh1/ETTh2, M1/M2=ETTm1/ETTm2, Wea=Weather, Elec=Electricity, Traf=Traffic. The MAE results are shown in appendix Table~\ref{tab:main_res_mae_combined}.}
    \label{tab:main_res_combined}
    
    \vspace{-6pt} 
    \begin{minipage}[t]{0.46\textwidth}
    \vspace{0pt}
    \centering
    {\small\textbf{(a) Foundation Model Candidates}}\\[2pt]
    \setlength{\tabcolsep}{1.2pt}
    \tiny
    \begin{tabular}{l|cccccccc|c}
        \toprule
        \textbf{Method} & \textbf{Exch} & \textbf{H1} & \textbf{H2} & \textbf{M1} & \textbf{M2} & \textbf{Wea} & \textbf{Elec} & \textbf{Traf} & \textbf{Avg} \\
        \midrule
        \multicolumn{10}{l}{\textit{Individual Models}} \\
       MOIRAI & .1675 & .2161 & .2413 & .1273 & .3164 & .0269 & .5908 & .1489 & .2294 \\
        MOMENT & .3037 & .1966 & .2765 & .1124 & .1707 & .5683 & .7635 & .9884 & .4225 \\
        TimeMoE & .2359 & .1768 & .1996 & .0997 & .1416 & .0752 & .3902 & .0831 & .1753 \\
        TimesFM & .1845 & .1712 & .2212 & .1112 & .2862 & .0039 & .4770 & .0864 & .1927 \\
        \midrule
        \multicolumn{10}{l}{\textit{Traditional Ensemble}} \\
        Ens$_{\text{avg}}$ & .1932 & .1624 & .2077 & .0944 & .1771 & .0766 & .4593 & .1508 & .1902 \\
        Ens$_{\text{rand}}$ & .1990 & .1680 & .2132 & .0979 & .1875 & .0939 & .4786 & .1857 & .2030 \\
        InvMSE$_{\text{tr}}$ & .1831 & .1621 & .2061 & .0940 & .1605 & .0560 & .4383 & .0831 & .1729 \\
        InvMSE$_{\text{val}}$ & .1915 & .1621 & .2062 & .0940 & .1602 & .0082 & .4379 & .0825 & .1678 \\
        NN$_{RLMC}$ & .1864 & .1617 & .2047 & .0915 & .1615 & .0032 & .3900 & .0821 & .1602 \\
        OptW$_{\text{tr}}$ & .1703 & .1613 & .1988 & .0929 & .1429 & .0433 & .3932 & .0821 & .1606 \\
        OptW$_{\text{val}}$ & .1967 & .1618 & .2013 & .0928 & .1429 & .0069 & .3930 & .0823 & .1597 \\
        \midrule
        \multicolumn{10}{l}{\textit{LLM-based Ensemble}} \\
        GPT-5.2 & .1617 & .1491 & .1936 & .0874 & .1579 & .0237 & .4095 & .0942 & .1596 \\
        Codex & .1605 & .1471 & .1924 & .0860 & .1490 & .0424 & .4064 & .0914 & .1594 \\
        GPT-5.5 & .1650 & .1529 & .1973 & .0890 & .1644 & .0745 & .4191 & .1036 & .1707 \\
        DS-V3.2 & .1535 & .1448 & .1919 & .0852 & .1506 & .0392 & .4086 & .1084 & .1603 \\
        Grok-4 & .1681 & .1521 & .1986 & .0897 & .1626 & .0836 & .4337 & .1143 & .1753 \\
        \midrule
        \multicolumn{10}{l}{\textit{REATS (Ours)}} \\
        REATS-SFT & .1466 & .1403 & .1864 & .0831 & .1381 & .0050 & .3839 & .0806 & .1455 \\
        \textbf{REATS-GRPO} & \textbf{.1368} & \textbf{.1336} & \textbf{.1763} & \textbf{.0806} & \textbf{.1320} & \textbf{.0030} & \textbf{.3688} & \textbf{.0760} & \textbf{.1384} \\
        \bottomrule
    \end{tabular}
    \end{minipage}
    \hfill
    \begin{minipage}[t]{0.52\textwidth}
    \vspace{0pt}
    \centering
    {\small\textbf{(b) Small Model Candidates}}\\[2pt]
    \setlength{\tabcolsep}{0.7pt}
    \tiny
    \begin{tabular}{l|cccccccc|c}
        \toprule
        \textbf{Method} & \textbf{Exch} & \textbf{H1} & \textbf{H2} & \textbf{M1} & \textbf{M2} & \textbf{Wea} & \textbf{Elec} & \textbf{Traf} & \textbf{Avg} \\
        \midrule
        \multicolumn{10}{l}{\textit{Individual Models}} \\
  CARD & .1690 & .1647 & .1944 & .0979 & .1147 & .4372 & .4442 & .0667 & .2111 \\
        LSINet & .5319 & .1558 & .1839 & .0815 & .0839 & 1.2352 & .2607 & 2.1674 & .5875 \\
        PDF & .6344 & .1515 & .1825 & .0748 & .0729 & .8325 & \textbf{.1472} & 2.1738 & .5337 \\
        TimeXer & 1.4382 & 1.9124 & 1.9351 & 1.8879 & 1.9464 & .4429 & 1.6963 & 2.0720 & 1.6664 \\
        \midrule
        \multicolumn{10}{l}{\textit{Traditional Ensemble}} \\
        Ens$_{\text{avg}}$ & .3146 & .2420 & .2730 & .1753 & .1892 & .1926 & .3168 & .8353 & .3173 \\
        Ens$_{\text{rand}}$ & .3918 & .3127 & .3439 & .2476 & .2636 & .3026 & .3813 & .9927 & .4045 \\
        InvMSE$_{\text{tr}}$ & .1879 & .1517 & .1810 & .0781 & .0783 & .2081 & .1757 & .0750 & .1420 \\
        InvMSE$_{\text{val}}$ & .3064 & .1591 & .2058 & .0866 & .0783 & .1932 & .1869 & .0733 & .1612 \\
        RLMC & .3875 & .1572 & .3029 & .0970 & .0838 & .1666 & .1549 & \textbf{.0666} & .1771 \\
        OptW$_{\text{tr}}$ & \textbf{.1561} & .1549 & .1795 & .0753 & .0740 & .2072 & .1486 & .0859 & .1352 \\
        OptW$_{\text{val}}$ & .2629 & .1599 & .1911 & .0928 & .0733 & .1928 & .1540 & .0858 & .1516 \\
        \midrule
        \multicolumn{10}{l}{\textit{LLM-based Ensemble}} \\
        GPT-5.2 & .2497 & .1857 & .2112 & .1103 & .1140 & .1130 & .2127 & .2889 & .1857 \\
        Codex & .2397 & .1724 & .2014 & .0994 & .1059 & .1240 & .2016 & .2231 & .1709 \\
        GPT-5.5 & .2574 & .1904 & .2155 & .1150 & .1182 & .1378 & .2255 & .3821 & .2052 \\
        DS-V3.2 & .2435 & .1708 & .2008 & .1050 & .1077 & .1168 & .1943 & .2439 & .1729 \\
        Grok-4 & .2617 & .1862 & .2191 & .1164 & .1253 & .1809 & .2232 & .4550 & .2210 \\
        \midrule
        \multicolumn{10}{l}{\textit{REATS (Ours)}} \\
        REATS-SFT & .1888 & .1492 & .1769 & .0785 & .0792 & .0424 & .1605 & .0925 & .1210 \\
        \textbf{REATS-GRPO} & .1587 & \textbf{.1409} & \textbf{.1700} & \textbf{.0725} & \textbf{.0720} & \textbf{.0206} & .1476 & .0819 & \textbf{.1080} \\
        \bottomrule
    \end{tabular}
    \end{minipage}
\end{table*}

\begin{table*}[t]
    \centering
    \caption{ Left: Trained on small model candidates from Table~\ref{tab:main_res_combined}(b), evaluated on 4 unseen foundation models (mapping in Appendix Table~\ref{tab:ood_model_mapping}). Right: Ablation study on small model candidates. Default REATS uses CoT + RAG + integer percentage table. ID = in-distribution, OOD = out-of-distribution. MAE results are  in appendix Table~\ref{tab:ood_combined_mae}. Other OOD results with similar conclusions are in Table~\ref{tab:ood_within_combined}.}
    \label{tab:ood_foundation}
    \vspace{-6pt}
    \begin{minipage}[t]{0.49\textwidth}
    \vspace{0pt}
    \centering
    {\small\textbf{(a) OOD to Foundation Models (MSE $\downarrow$)}}\\[2pt]
    \setlength{\tabcolsep}{2pt}
    \resizebox{\textwidth}{!}{
    \begin{tabular}{l|cccccccc|c}
        \toprule
        \textbf{Method} & \textbf{Exch} & \textbf{H1} & \textbf{H2} & \textbf{M1} & \textbf{M2} & \textbf{Wea} & \textbf{Elec} & \textbf{Traf} & \textbf{Avg} \\
        \midrule
        \multicolumn{10}{l}{\textit{Individual Models}} \\
        Model 1 & .3037 & .1966 & .2765 & .1124 & .1707 & .5683 & .7635 & .9884 & .4225 \\
        Model 2 & .2359 & .1742 & .2022 & .0987 & .1229 & .4397 & .3740 & .0893 & .2171 \\
        Model 3 & .1845 & .1712 & .2212 & .1112 & .1416 & .0752 & .5569 & .1800 & .2052 \\
        Model 4 & .1675 & .2161 & .2413 & .1273 & .3164 & \textbf{.0039} & .5908 & .0831 & .2183 \\
        \midrule
        \multicolumn{10}{l}{\textit{Traditional Ensemble}} \\
        Ens$_{\text{avg}}$ & .1930 & .1618 & .2090 & .0946 & .1391 & .1670 & .4765 & .1725 & .2017 \\
        Ens$_{\text{rand}}$ & .1994 & .1674 & .2141 & .0981 & .1489 & .1837 & .4957 & .2056 & .2141 \\
        InvMSE$_{\text{tr}}$ & .1831 & .1615 & .2076 & .0943 & .1290 & .1083 & .4485 & .0906 & .1779 \\
        InvMSE$_{\text{val}}$ & .1915 & .1615 & .2079 & .0943 & .1291 & .0072 & .4481 & .0896 & .1662 \\
        RLMC & .1697 & .1619 & .2106 & .0902 & .1228 & .2903 & .3740 & \textbf{.0795} & .1874 \\
        OptW$_{\text{tr}}$ & .1703 & .1608 & .2013 & .0930 & .1182 & .0761 & .3797 & .0839 & .1604 \\
        OptW$_{\text{val}}$ & .1967 & .1619 & .2067 & .0927 & .1184 & .0109 & .3797 & .0842 & .1564 \\
        \midrule
        \multicolumn{10}{l}{\textit{LLM-based Ensemble}} \\
        GPT-5.2 & .1656 & .1486 & .1965 & .0881 & .1152 & .0771 & .4158 & .1122 & .1649 \\
        Codex & .1598 & .1453 & .1952 & .0862 & .1126 & .0973 & .4157 & .1013 & .1642 \\
        GPT-5.5 & .1710 & .1546 & .2035 & .0912 & .1260 & .1157 & .4356 & .1259 & .1779 \\
        DeepSeek-V3.2 & .1587 & .1449 & .1940 & .0860 & .1080 & .1047 & .4032 & .1011 & .1626 \\
        Grok-4 & .1734 & .1531 & .2021 & .0897 & .1258 & .1641 & .4438 & .1353 & .1859 \\
        \midrule
        \multicolumn{10}{l}{\textit{REATS (Ours)}} \\
        \textbf{REATS-GRPO} & \textbf{.1388} & \textbf{.1350} & \textbf{.1821} & \textbf{.0854} & \textbf{.1043} & .0062 & \textbf{.3613} & .1409 & \textbf{.1442} \\
        \bottomrule
    \end{tabular}
    }
    \end{minipage}
    \hfill
    \begin{minipage}[t]{0.49\textwidth}
    \vspace{0pt}
    \centering
    {\small\textbf{(b) Component \& Format Ablation (MSE $\downarrow$)}}\label{tab:ablation_small}\\[2pt]
    \setlength{\tabcolsep}{2pt}
    \resizebox{\textwidth}{!}{
    \begin{tabular}{cl|cccccccc|c}
        \toprule
        \textbf{Cat.} & \textbf{Variant} & \textbf{Exch} & \textbf{H1} & \textbf{H2} & \textbf{M1} & \textbf{M2} & \textbf{Wea} & \textbf{Elec} & \textbf{Traf} & \textbf{Avg} \\
        \midrule
        \multirow{4}{*}{\rotatebox[origin=c]{90}{\scriptsize CoT}} & REATS-SFT (ID) & \textbf{.1888} & \textbf{.1492} & \textbf{.1769} & \textbf{.0785} & \textbf{.0792} & \textbf{.0424} & .1605 & \textbf{.0925} & \textbf{.1210} \\
         & w/o CoT (ID) & .2415 & .1587 & .1886 & .0833 & .0839 & .0447 & \textbf{.1582} & .0929 & .1315 \\
        \cline{2-11}
         & REATS-SFT (OOD) & \textbf{.1881} & \textbf{.2125} & \textbf{.3408} & \textbf{.1051} & \textbf{.1705} & .0653 & .2001 & \textbf{.0924} & \textbf{.1719} \\
         & w/o CoT (OOD) & .2412 & .2639 & .3689 & .2070 & .7563 & \textbf{.0292} & \textbf{.1809} & .0926 & .2675 \\
        \midrule
        \multirow{4}{*}{\rotatebox[origin=c]{90}{\scriptsize RAG}} & REATS-SFT (ID) & \textbf{.1888} & \textbf{.1492} & \textbf{.1769} & \textbf{.0785} & .0792 & \textbf{.0424} & \textbf{.1605} & \textbf{.0925} & \textbf{.1210} \\
         & w/o RAG (ID) & .2388 & .1532 & .1815 & .0790 & \textbf{.0792} & .1464 & .1629 & .0935 & .1418 \\
        \cline{2-11}
         & REATS-SFT (OOD) & \textbf{.1881} & \textbf{.2125} & \textbf{.3408} & \textbf{.1051} & \textbf{.1705} & \textbf{.0653} & \textbf{.2001} & \textbf{.0924} & \textbf{.1719} \\
         & w/o RAG (OOD) & .3499 & .2378 & .7690 & .3823 & .3524 & .1413 & .4445 & .0990 & .3470 \\
        \midrule
        \multirow{8}{*}{\rotatebox[origin=c]{90}{\scriptsize Weight Fmt}} & REATS-SFT (ID) & \textbf{.1888} & .1492 & \textbf{.1769} & .0785 & .0792 & .0424 & .1605 & \textbf{.0925} & \textbf{.1210} \\
         & Decimal (ID) & .2182 & .1511 & .1776 & \textbf{.0769} & \textbf{.0779} & .0429 & .1597 & .0931 & .1247 \\
         & Dict (ID) & .2060 & \textbf{.1487} & .1775 & .0781 & .0786 & .0567 & .1613 & .0933 & .1250 \\
         & Array (ID) & .2236 & .1500 & .1788 & .0779 & .0785 & \textbf{.0309} & \textbf{.1575} & .0926 & .1237 \\
        \cline{2-11}
         & REATS-SFT (OOD) & \textbf{.1881} & .2125 & .3408 & .1051 & .1705 & .0653 & \textbf{.2001} & \textbf{.0924} & .1719 \\
         & Decimal (OOD) & .1950 & .2164 & \textbf{.3043} & \textbf{.1034} & .1682 & .0720 & .2069 & .0979 & \textbf{.1705} \\
         & Dict (OOD) & .1905 & .2223 & .3293 & .1165 & \textbf{.1481} & .0737 & .2105 & .0947 & .1732 \\
         & Array (OOD) & .2797 & \textbf{.1966} & .3595 & .2233 & .2627 & \textbf{.0511} & .2061 & .0925 & .2089 \\
        \bottomrule
    \end{tabular}
    }
    \par\vspace{6pt}
    {\small\textbf{(c) Token Efficiency \& Inference Speed (s/sample)}}\\[2pt]
    \setlength{\tabcolsep}{2pt}
    \resizebox{\textwidth}{!}{
    \begin{tabular}{l|l|c|c|c|c}
        \hline
        Format & Example Row & Tokens & vs Table & Infer & vs Table \\
        \hline
        \hline
        \textbf{Table} & \texttt{40,30,20,10} & \textbf{103} & --- & \textbf{1.10} & --- \\
        Array & \texttt{[40, 30, 20, 10],...,} & 132 & +28\% & 1.21 & +10\% \\
        Decimal & \texttt{0.40,0.30,0.20,0.10,...,} & 182 & +77\% & 1.42 & +29\% \\
        Dict & \makecell[l]{\texttt{\{"CARD":40, "LSINet":30,...,}} & 292 & +183\% & 2.05 & +86\% \\
        \hline
    \end{tabular}
    }
    \end{minipage}
\end{table*}

\subsection{Discussion of GRPO reward mapping}
\label{sec:reward_discussion}

\begin{figure*}[h!]
\centerline{\includegraphics[width=1\linewidth]{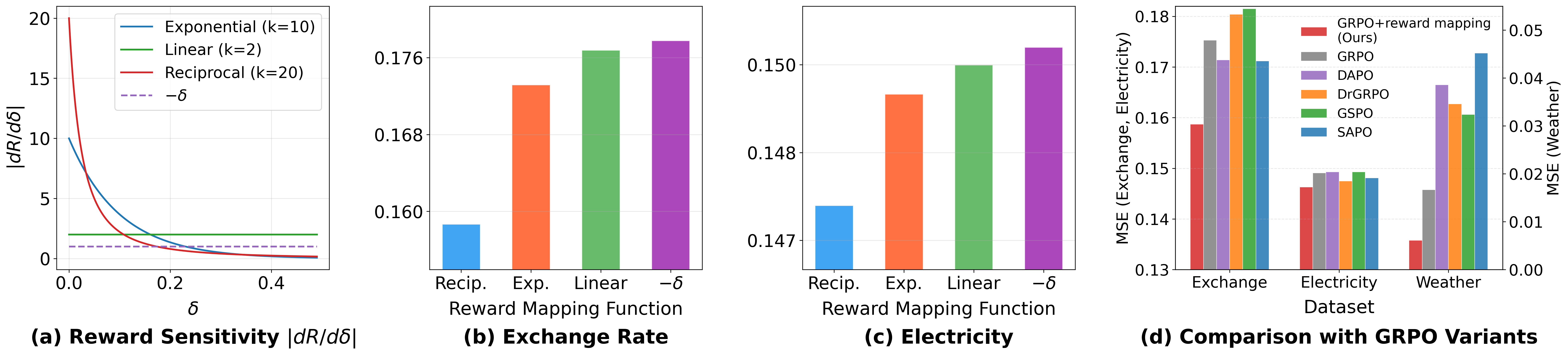} }
\caption{Reward mapping analysis. (a) Raw reward sensitivity $|dr/d\delta|$ of four mappings. (b)(c) Empirical MSE on Exchange and Electricity under different reward mappings. (d) Comparison with recent GRPO variants (all using naive $r{=}-\delta$). Full results in Appendix Table~\ref{tab:grpo_variants}.}
\label{fig:reward_combined}
\end{figure*}

\noindent\textbf{Comparison of bounded mapping functions.}
We compare four reward mapping functions, including three bounded mappings (reciprocal $r=1/(1+k\delta)$, exponential $r=\exp(-k\delta)$, and linear $r=\max(0,1-k\delta)$) along with the naive unbounded baseline $r=-\delta$, and analyze them from both empirical and theoretical perspectives. Since GRPO generates diverse candidates within each group, the reward mapping must provide adequate signal across the full $\delta$ range to avoid gradient vanishing. Under this constraint, the reciprocal mapping (with polynomial decay) maintains effective sensitivity across a wide range while also achieving strong discrimination when $\delta$ is small (i.e., when the predicted weights are close to optimal), making it well-suited for GRPO optimization, as shown in Figure~\ref{fig:reward_combined}(a). Crucially, since $r{=}-\delta$ is unbounded, a single outlier generation with large $\delta$ inflates the group standard deviation $\sigma_r$, compressing the normalized advantages $\hat{A}_i{=}(r_i{-}\mu_r)/\sigma_r$ among near-optimal candidates to near-identical values. The reciprocal mapping addresses this through two mechanisms. (1) \emph{Boundedness}, where outlier rewards are compressed near zero, limiting the influence of any single generation on $\sigma_r$. (2) \emph{Nonlinear curvature}, where it reshapes the relative spacing of rewards within a group, and this geometric change in relative positions is preserved after advantage normalization (unlike a linear rescaling, which would be canceled).

Details on selecting the recommended $k$ are provided in Appendix~\ref{sec:reward_mapping_appendix}. Appendix~\ref{sec:reward_sensitivity} analyzes reward sensitivity, including why $\mathrm{d}R/\mathrm{d}\delta$ serves as a reliable sensitivity measure. Appendix~\ref{sec:reward_theory} provides theoretical proofs that the reciprocal mapping maintains adequate sensitivity across a wide $\delta$ range.
Figure~\ref{fig:reward_combined}(b)(c) confirms that the reciprocal mapping consistently achieves the lowest MSE across both datasets, while the naive $r{=}-\delta$ performs comparably to or worse than the bounded alternatives. Beyond the uniform sensitivity visible in (a), the unbounded range of $r{=}-\delta$ also allows outlier generations with large $\delta$ to dominate group variance $\sigma_r$, compressing normalized advantages among near-optimal candidates in mixed-quality rollout groups (detailed analysis in Appendix~\ref{sec:naive_reward_analysis}).

\noindent\textbf{Comparison with GRPO variants.}

Reward signal sparsity is a recognized challenge in GRPO, and several recent variants (DAPO~\cite{yu2026dapo}, DrGRPO~\cite{liuunderstanding}, GSPO~\cite{zheng2025group}, and SAPO~\cite{gao2025soft}) address it through optimization-level modifications, such as reward normalization, sample filtering, and soft advantage computation. To compare reward-level and optimization-level solutions, we evaluate standard GRPO equipped with our reciprocal mapping against these variants, each using the shared naive reward $r{=}-\delta$. As shown in Figure~\ref{fig:reward_combined}(d), standard GRPO with our reciprocal mapping consistently outperforms the evaluated optimization-level variants, suggesting that when rewards are continuous and unbounded, directly improving the reward signal quality is also a promising direction worth exploring (full results in Appendix Table~\ref{tab:grpo_variants}).

\begin{figure*}[h!]
    \centering
    \setlength{\abovecaptionskip}{2pt}
    \begin{minipage}[t]{0.55\textwidth}
    \vspace{-8pt}
    \centering
    \includegraphics[width=\textwidth]{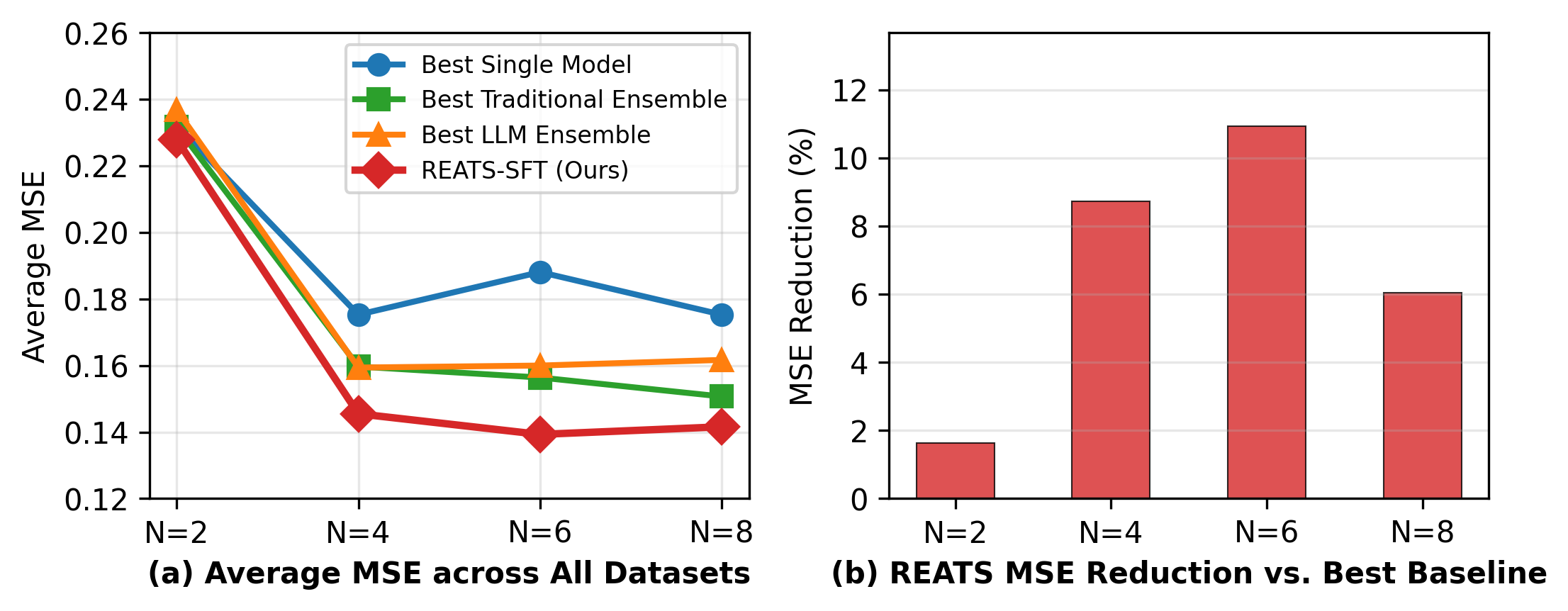}
    \caption{Scalability to different numbers of candidate models. N=4 results correspond to Table~\ref{tab:main_res_combined}(a). Full results for other N values are in Appendix Table~\ref{tab:scalability_zeroshot_combined} (MSE) and Table~\ref{tab:scalability_zeroshot_combined_mae} (MAE).}
    \label{fig:scalability}
    \par\vspace{3pt}
    \captionof{table}{{Effectiveness of the diverse weight table, where $K'$ denotes the number of weight rows.}}
    \label{tab:ablation_rows}
    \setlength{\tabcolsep}{3pt}
    {
    \resizebox{\textwidth}{!}{
    \begin{tabular}{c|cccccccc|c}
        \hline
         & Exch & H1 & H2 & M1 & M2 & Wea & Elec & Traf & Avg \\
        \hline
        $K'{=}1$  & .1753 & .1410 & .1785 & .0740 & .0739 & .0745 & .1491 & .0855 & .1190 \\
        $K'{=}10$ & \textbf{.1587} & \textbf{.1409} & \textbf{.1700} & \textbf{.0725} & \textbf{.0720} & \textbf{.0206} & \textbf{.1476} & \textbf{.0819} & \textbf{.1080} \\
        \hline
    \end{tabular}
    }
    }
    \end{minipage}
    \hfill
    \begin{minipage}[t]{0.42\textwidth}
    \vspace{0pt}
    \centering
    \includegraphics[width=\textwidth]{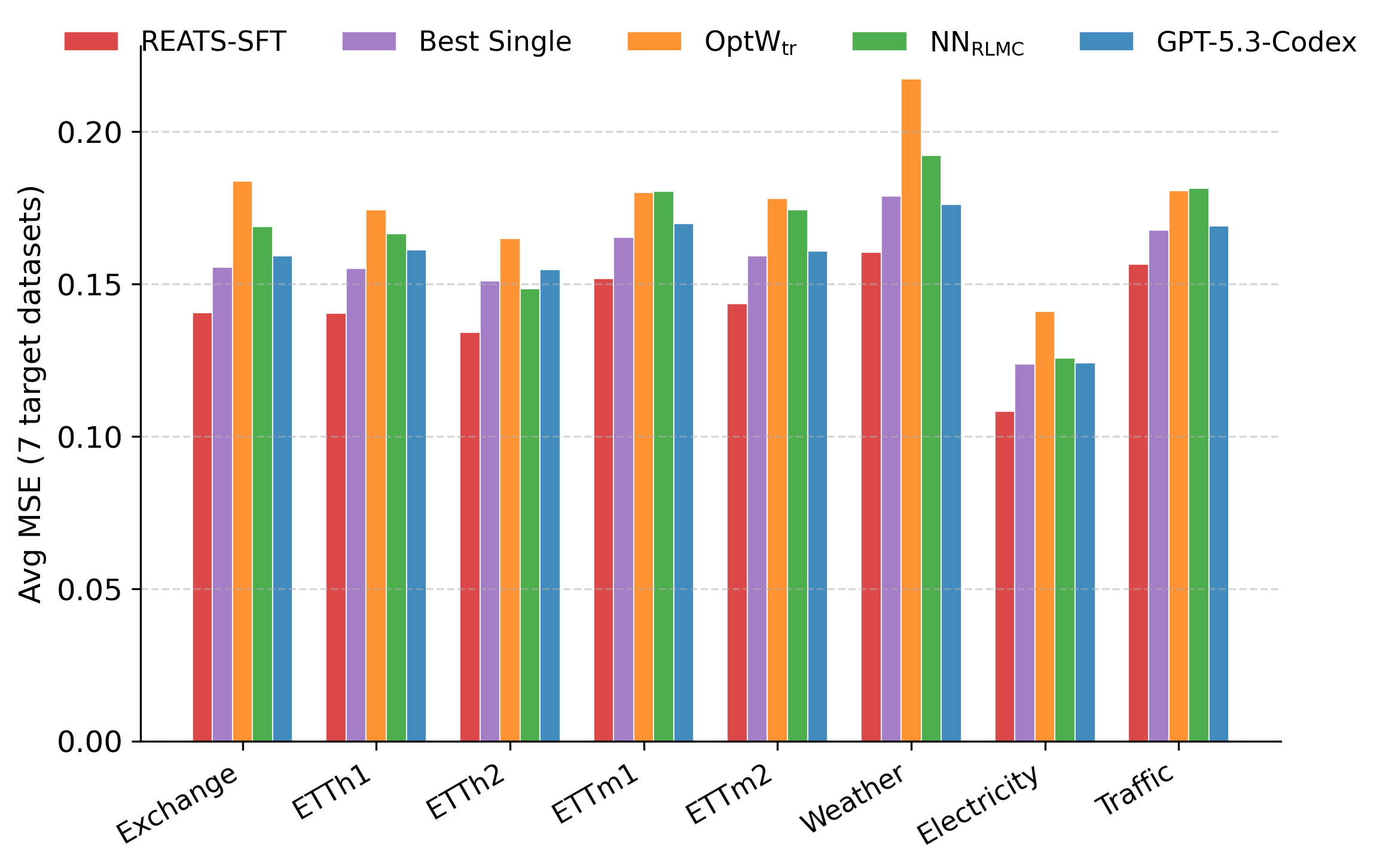}
    \caption{Transfer learning comparison (zero-shot model group, MSE $\downarrow$). Each source dataset (X-axis) is used for training, and the average MSE across the other 7 target datasets (Y-axis) is reported. Baselines are the best method from each category.}
    \label{fig:transfer_zeroshot}
    \end{minipage}
\end{figure*}

\subsection{Ablation study}
We ablate three design choices in REATS-SFT using the small model candidates (Table~\ref{tab:ablation_small}(b)).

\textbf{Chain-of-Thought (CoT).} CoT activates the LLM's reasoning capability by requiring it to explicitly analyze temporal patterns and candidate model suitability before deciding weights (``reason-then-decide''). On ID data, CoT improves over direct prediction (MSE: 0.1210 vs.\ 0.1315, +8.7\%). On OOD data, reasoning becomes more critical: without CoT, MSE degrades from 0.1719 to 0.2675 (+55.6\%). \textbf{This is because when facing unseen candidates, the model must reason about input series characteristics, retrieved references, and each candidate's design properties}. Without CoT, the model can only rely on memorized input-weight mappings that fail to generalize to unseen candidates.

\textbf{Retrieval-Augmented Generation (RAG).} RAG retrieves historically similar forecasting cases as reference anchors for weight allocation. Removing RAG degrades both ID and OOD performance, with particularly large OOD degradation, suggesting that retrieved examples help the model ground its reasoning in unfamiliar scenarios.

\textbf{Weight output format.} We compare four formats: integer percentage table, decimal, dictionary, and array. The integer table achieves the best ID accuracy (0.1210 vs.\ 0.1247 for decimal) and near-best OOD accuracy (0.1719 vs.\ 0.1705 for decimal), while being the most token-efficient (103 vs.\ 292 tokens for Dict) and fastest at inference (1.10 vs.\ 2.05 s/sample). Considering the combined advantage in accuracy, efficiency, and parsing simplicity, integer table is our default format, as shown in Table~\ref{tab:ablation_small}(b).

\noindent\textbf{Number of supervision weight rows $K'$.}
Table~\ref{tab:ablation_rows} ablates the diverse weight table during GRPO. Reducing $K'$ from 10 to 1 (oracle row only) raises the average MSE from 0.1080 to 0.1190 (+10.2\%), degrading on all eight datasets. The reason is that $r_{\text{mse}}{=}\frac{1}{K'}\sum_i 1/(1{+}k\delta_i)$ evaluates each rollout on $K'$ allocations rather than one, which mitigates reward sparsity and yields better-separated rewards within a rollout group. Moreover, the $K'$ rows form a neighborhood in weight space, where the first row targets the oracle weights $\mathbf{w}^*$ and the remaining $K'{-}1$ rows spread over diverse near-optimal allocations. The reward therefore reflects the quality of a whole region of the weight simplex, and the gradient indicates which direction around the oracle is preferable, giving GRPO a richer learning signal than a single-row comparison.

\textbf{Input representation.} We compare three input formats: hybrid textual--numerical (ours), raw time series as text, and raw series via MLP encoder (Table~\ref{tab:input_ablation_small}). Across all input-representation variants, we keep the model backbone, CoT, and weight supervision, and training configuration unchanged, modifying only the input representation.
On ID data, textual--numerical and the MLP encoder achieve comparable best accuracy, both outperforming raw-TS text. 
On OOD data, textual--numerical generalizes best, as textual descriptions activate the LLM's semantic reasoning rather than relying on position-dependent numerical patterns. Additionally, textual--numerical uses a fixed token budget (1085 tokens) regardless of input length, while raw-TS scales linearly (e.g., 1656$\to$7480 from length 96 to 512, Table~\ref{tab:token_count}). Crucially, only the structured textual format enables rule-based CoT construction (Section~\ref{sec:supervision}), which is key to enhance ensemble performance.


\subsection{Discussion on hyperparameters}

\begin{table*}[t]
    \centering
    \begin{minipage}[t]{0.49\textwidth}
    \centering
    \captionof{table}{RAG top-$k$ analysis (MSE $\downarrow$, small model group).}
    \label{tab:ablation_topk_small}
    \vspace{-6pt}
    \setlength{\tabcolsep}{3pt}
    \resizebox{\textwidth}{!}{
    \begin{tabular}{c|cccccccc|c}
        \hline
           & Exch & H1 & H2 & M1 & M2 & Wea & Elec & Traf & Avg \\
        \hline
        $k{=}25$  & .2132 & .1508 & .1805 & \textbf{.0780} & \textbf{.0791} & .0535 & .1612 & .0935 & .1262 \\
        $k{=}3$  & \textbf{.1888} & \textbf{.1492} & \textbf{.1769} & .0785 & .0792 & \textbf{.0424} & \textbf{.1605} & \textbf{.0925} & \textbf{.1210} \\
        \hline
    \end{tabular}
    }
    \par\vspace{6pt}
    \includegraphics[width=\textwidth]{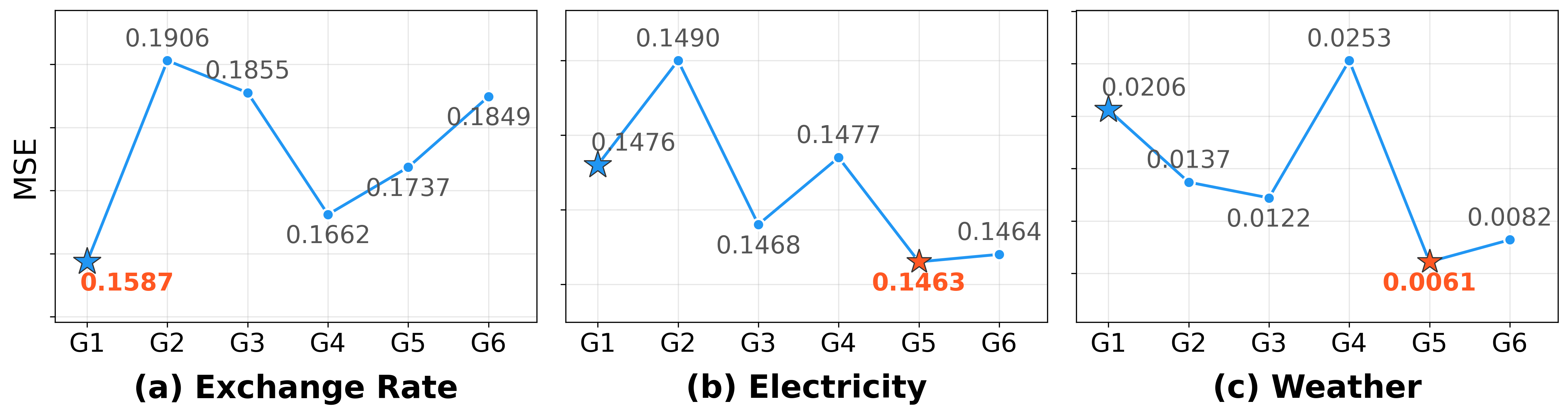}
    \captionof{figure}{GRPO reward coefficient sensitivity across datasets. G1--G6 correspond to $(\lambda_1, \lambda_2, \lambda_3)$ = (0.8,0,0.2), (0.5,0,0.5), (0,0.8,0.2), (0.3,0.3,0.4), (1,0,0), (0,1,0).}
    \label{fig:hyper_ablation}
    \end{minipage}
    \hfill
    \begin{minipage}[t]{0.49\textwidth}
    \centering
    \captionof{table}{Further analysis on CoT (MSE $\downarrow$). The settings are same in Table~\ref{tab:main_res_combined}.}
    \label{tab:cot_combined}
    \vspace{-4pt}
    {\small\textbf{(a) Rule-CoT vs GPT-CoT on SFT.}}\\[2pt]
    \setlength{\tabcolsep}{2pt}
    \resizebox{\textwidth}{!}{
    \begin{tabular}{ll|cccccccc|c}
        \toprule
        \textbf{Group} & \textbf{Method} & \textbf{Exch} & \textbf{H1} & \textbf{H2} & \textbf{M1} & \textbf{M2} & \textbf{Wea} & \textbf{Elec} & \textbf{Traf} & \textbf{Avg} \\
        \midrule
        \multirow{2}{*}{Small}
         & GPT-CoT        & .2152 & .1591 & .1822 & .0793 & .0815 & \textbf{.0294} & \textbf{.1589} & .1025 & .1260 \\
         & Rule-CoT       & \textbf{.1888} & \textbf{.1492} & \textbf{.1769} & \textbf{.0785} & \textbf{.0792} & .0424 & .1605 & \textbf{.0925} & \textbf{.1210} \\
        \midrule
        \multirow{2}{*}{TSFM}
         & GPT-CoT        & .1486 & .1406 & .1866 & \textbf{.0828} & \textbf{.1359} & \textbf{.0045} & .3851 & \textbf{.0805} & .1456 \\
         & Rule-CoT       & \textbf{.1466} & \textbf{.1403} & \textbf{.1864} & .0831 & .1381 & .0050 & \textbf{.3839} & .0806 & \textbf{.1455} \\
        \bottomrule
    \end{tabular}
    }
    \par\vspace{6pt}
    {\small\textbf{(b) Effectiveness of CoT on GRPO}}\\[2pt]
    \setlength{\tabcolsep}{3pt}
    \resizebox{\textwidth}{!}{
    \begin{tabular}{ll|cccccccc|c}
        \toprule
        \textbf{Group} & \textbf{Method} & \textbf{Exch} & \textbf{H1} & \textbf{H2} & \textbf{M1} & \textbf{M2} & \textbf{Wea} & \textbf{Elec} & \textbf{Traf} & \textbf{Avg} \\
        \midrule
        \multirow{2}{*}{Small}
         & w/o CoT & .1936 & .1424 & .1704 & .0725 & .0735 & .0359 & .1491 & .0835 & .1151 \\
         & w/ CoT  & \textbf{.1587} & \textbf{.1409} & \textbf{.1700} & \textbf{.0725} & \textbf{.0720} & \textbf{.0206} & \textbf{.1476} & \textbf{.0819} & \textbf{.1080} \\
        \bottomrule
    \end{tabular}
    }
    \end{minipage}
\end{table*}

\noindent\textbf{Number of neighbors in Retrieval-Augmented Generation (RAG).}
We compare $k{=}3$ and $k{=}25$ retrieved neighbors in Table~\ref{tab:ablation_topk_small}. The smaller $k{=}3$ achieves lower average MSE (0.1210 vs.\ 0.1262), suggesting that a few highly similar reference cases provide more precise anchoring than many less-relevant ones, which may introduce noise into the reasoning process.

\noindent\textbf{GRPO reward coefficients.}
We compare six configurations in Figure~\ref{fig:hyper_ablation}. Key findings: \textbf{(1)} Per-row $r_{\text{mse}}$ evaluates each of the $K'$ weight rows independently, providing implicit reward augmentation with finer-grained feedback, while $r_{\text{agg}}$ aggregates all rows before computing a single reward. G5 (pure $r_{\text{mse}}$) achieves lower MSE than G6 (pure $r_{\text{agg}}$) across all datasets, confirming per-row evaluation is more effective. \textbf{(2)} $r_{\text{oracle}}$ helps stabilize early training on small datasets (Exchange: G1 outperforms G5), but on larger datasets (Electricity, Weather), pure $r_{\text{mse}}$ (G5) yields the best results without conflicting objectives. Excessive oracle weight (G2) consistently hurts. \textbf{(3)} Combining all three (G4) degrades performance by inheriting the less effective $r_{\text{agg}}$ and the limiting $r_{\text{oracle}}$. Our default G1 ($\lambda_1{=}0.8, \lambda_3{=}0.2$) balances MSE optimization with oracle guidance across varying dataset sizes.

\subsection{Discussion on CoT}

\noindent\textbf{Rule CoT VS LLM CoT.} Both Rule-CoT and GPT-CoT are constructed using the same temporal features, candidate-model information, RAG references, and explicitly provided oracle weights. They differ only in how the reasoning text is generated.
As shown in Table~\ref{tab:cot_combined}(a), Rule-CoT outperforms GPT-CoT on 6/8 datasets in the small group (avg 0.1210 vs.\ 0.1260) and achieves near-identical results in the TSFM group (0.1455 vs.\ 0.1456). Rule-CoT produces structurally consistent reasoning that is easier for SFT to learn, whereas GPT-CoT may introduce hallucinated or inconsistent analysis across similar inputs. Given comparable or better performance at zero API cost, we adopt Rule-CoT as default.

\noindent\textbf{Impact of CoT on GRPO.}
Table~\ref{tab:cot_combined}(b) shows that CoT improves GRPO across all 8 datasets, reducing average MSE from 0.1151 to 0.1080 (6.2\% relative gain). 
CoT forces the model to produce explicit reasoning before outputting weights; 
during GRPO sampling, different reasoning paths (e.g., emphasizing trend vs.\ 
seasonality) naturally lead to diverse weight assignments, enriching exploration 
and yielding richer reward signals. Without CoT, the model directly generates 
weight vectors that lack diversity, resulting in sparse reward variance.

\noindent\textbf{Interpretability and GRPO reasoning improvement.}
A key advantage of REATS over traditional ensemble methods is its inherent interpretability: the model generates structured chain-of-thought reasoning (series pattern analysis $\rightarrow$ tool-to-pattern matching $\rightarrow$ RAG reference $\rightarrow$ allocation conclusion) before producing weights, enabling users to understand \emph{why} specific tools receive higher allocations. Furthermore, GRPO improves not only ensemble accuracy but also reasoning quality: on the Weather dataset, SFT defaults to PDF (periodicity-based) for an autocorrelation-dominated series, while GRPO correctly identifies CARD (attention-based) as the primary tool with new reasoning paths, aligning with oracle weights. A detailed case study is provided in Appendix~\ref{sec:interpretability}.

\subsection{Transfer learning comparison}

To evaluate cross-dataset transferability, we train REATS on a single source dataset and test on the remaining 7 target datasets. As shown in Figure~\ref{fig:transfer_zeroshot} (zero-shot model group), REATS-SFT generally achieves the lowest average MSE across source datasets. Traditional baselines rely solely on numerical inputs (raw forecasts and statistics), making their learned weight mappings tightly coupled to source-domain patterns and prone to overfitting. REATS, in contrast, combines textual descriptions of temporal features with numerical inputs and leverages the LLM's unique CoT reasoning capability, which provides a more transferable representation: the text-based reasoning generalizes across datasets because it operates on semantic-level temporal characteristics (e.g., trend strength, seasonality type) rather than dataset-specific numerical distributions. Results on the small model group show a similar trend (Appendix Figure~\ref{fig:transfer_small}).

\subsection{Efficiency analysis}

Compared to general-purpose LLMs such as GPT, Grok, and DeepSeek, REATS uses a 1.7B model that can be fully fine-tuned and deployed on a single lightweight GPU. To further reduce costs, we replace LLM-generated CoT with rule-based CoT, which significantly lowers dataset construction and training overhead without sacrificing accuracy (Table~\ref{tab:cot_combined}(a)). The integer percentage weight table additionally reduces token consumption and inference latency (Table~\ref{tab:ablation_small}(c)). While REATS is not as fast as traditional ensemble methods, it leverages LLM reasoning to achieve higher ensemble learning accuracy and interpretability, offering a favorable trade-off between efficiency and performance.

\section{Conclusion}

We presented REATS, a framework that repurposes LLM reasoning for ensemble learning and TSF{, with a study of the key techniques this paradigm requires, spanning input representation, scalable CoT and weight supervision, output format, and reward design}. By proposing hybrid textual--numerical inputs, chain-of-thought reasoning, and a two-stage SFT-to-GRPO fine-tuning pipeline, REATS enables a lightweight 1.7B LLM to produce sample-adaptive, interpretable ensemble weights that generally outperform both traditional and zero-shot LLM ensemble baselines across the evaluated settings. Key empirical findings include: (1) GRPO with reciprocal reward mapping effectively refines weights beyond SFT imitation, and outperforms recent GRPO algorithmic variants in the continuous unbounded reward setting{, with multi-row diverse weight supervision providing a denser reward signal that is complementary to the mapping}; (2) rule-based CoT achieves comparable or better performance than API-generated CoT at zero cost, and is critical for OOD generalization, enabling the model to reason about unseen candidates rather than relying on memorized patterns; and (3) LLM-based semantic reasoning enables superior transfer learning across datasets compared to numerical-only baselines.

\bibliographystyle{unsrt}  
\bibliography{main} 

\appendix
\clearpage
\newpage

\section{Appendices}

\begin{algorithm}[h!]
\caption{Rule-Based Chain-of-Thought Generation (Detailed)}
\label{alg:rule_cot_detail}

\KwIn{Temporal features $\Phi(\mathbf{X})$, tool set $\mathcal{T}=\{t_1,...,t_N\}$ each with category $c_i \in \{\text{linear, transformer, FFT, decomp., CNN, foundation}\}$, oracle weights $\mathbf{w}^* \in \mathbb{R}^{K \times N}$, RAG features $\Phi(\mathbf{X}_{\text{rag}})$ and weights $\mathbf{w}_{\text{rag}} \in \mathbb{R}^{K' \times N}$}
\KwOut{Chain-of-thought text $\textit{CoT}$}

$\bar{\mathbf{w}} \gets \frac{1}{K'}\sum_{k=1}^{K'} \mathbf{w}^*_k$ \tcp*{Average oracle weights across $K'$ rows}

\tcp{Step 1: Summarize key temporal patterns}
Parse $\Phi(\mathbf{X})$ into: trend strength/direction, noise level, ACF value, seasonality (yes/no), stationarity, outliers, distribution type\;
\ForEach{attribute $a$}{
    Sample phrase from predefined template pool $\mathcal{P}_a$ \tcp*{e.g., (strong, upward) $\to$ ``pronounced upward drift''}
}
$s_1 \gets$ \texttt{``Key patterns: ''} $+$ concatenate all sampled phrases\;

\tcp{Step 2: Oracle-guided tool-to-pattern matching}
\ForEach{tool $t_i$ in descending order of $\bar{w}_i$}{
    Retrieve category $c_i$ and its strength descriptors\;
    \tcp{Category-specific rule matching}
    \uIf{$c_i = \text{linear}$ \textbf{and} trend is strong}{
        reason $\gets$ ``strong trend favors linear decomposition''\;
    }
    \uElseIf{$c_i = \text{FFT}$ \textbf{and} seasonality detected}{
        reason $\gets$ ``FFT decomposition captures periodic structure''\;
    }
    \ElseIf{$c_i = \text{transformer}$ \textbf{and} ACF is high}{
        reason $\gets$ ``attention can exploit autocorrelation structure''\;
    }
    \tcp{... (other category-attribute rules)}
    \uIf{$\bar{w}_i > 0.35$}{
        $\text{sent}_i \gets$ sample from high-fit templates with reason \tcp*{``\{tool\} fits best because ...''}
    }
    \uElseIf{$\bar{w}_i > 0.15$}{
        $\text{sent}_i \gets$ sample from mid-fit templates \tcp*{``\{tool\} adds complementary value via ...''}
    }
    \Else{
        $\text{sent}_i \gets$ sample from low-fit templates with mismatch reason \tcp*{``\{tool\} has limited relevance ...''}
    }
}
$s_2 \gets$ \texttt{``Tool match: ''} $+$ join all $\text{sent}_i$\;

\tcp{Step 3: Compare with RAG reference}
$\bar{\mathbf{w}}_{\text{rag}} \gets \frac{1}{K}\sum_k \mathbf{w}_{\text{rag},k}$\;
$t_{\text{dom}} \gets \arg\max_i \bar{w}_i$; \quad $t_{\text{rag}} \gets \arg\max_i \bar{w}_{\text{rag},i}$\;
\uIf{$t_{\text{dom}} = t_{\text{rag}}$}{
    $s_3 \gets$ ``consistent with the current allocation direction''\;
}
\Else{
    Compare $\Phi(\mathbf{X})$ vs $\Phi(\mathbf{X}_{\text{rag}})$ to identify feature differences\;
    $s_3 \gets$ ``similar-series ensembles lean toward $t_{\text{rag}}$, but current series differs in \{differences\}, shifting preference toward $t_{\text{dom}}$''\;
}
$s_3 \gets$ \texttt{``Reference: ''} $+ \; s_3$\;

\tcp{Step 4: State allocation conclusion}
$\text{dom} \gets \{t_i : \bar{w}_i > 0.35\}$; \quad $\text{mid} \gets \{t_i : 0.12 < \bar{w}_i \leq 0.35\}$; \quad $\text{minor} \gets$ rest\;
$s_4 \gets$ \texttt{``Conclusion: ''} $+$ sample from conclusion templates naming dom/mid/minor\;

\Return $\textit{CoT} = s_1 \oplus s_2 \oplus s_3 \oplus s_4$\;
\end{algorithm}

\subsection{Related work: Time series forecasting}
\label{sec:related_tsf}

In recent years, time series forecasting has shifted from traditional statistical methods to deep learning–based approaches, which can be broadly categorized into Transformer-based models, convolutional networks, and lightweight linear models.

Transformers have been widely adopted due to their strong sequence modeling capabilities. Early works directly applied them along the temporal dimension, while later studies introduced task-specific designs. For example, PatchTST~\cite{nie2022time_patchformer} segments time series into patches and models channels independently. TimeXer~\cite{wangtimexer} adopts a decoder-only architecture with causal masking to capture temporal dependencies, while CARD~\cite{zhou2024card} focuses on cross-channel correlations via channel-aligned attention.  TimeMixer~\cite{wangtimemixer} and SEMixer~\cite{zhang2026semixer} model temporal dynamics across multiple scales via decomposition and mixing strategies. MLF~\cite{zhang2025multi} extends this idea by introducing multi-period modeling with adaptive patching and weighted fusion. In addition, efficient alternatives have been explored. ModernTCN~\cite{luo2024moderntcn} uses large-kernel convolutions to model dependencies without attention, while DLinear~\cite{zeng2023transformers_linear} and LSINet~\cite{zhang2025lightweight} demonstrate that simple linear architectures can achieve competitive performance with high efficiency.

Finally, inspired by large language models, time series foundation models learn general temporal representations via large-scale pretraining for cross-task generalization. Encoder-based methods such as MOMENT~\cite{goswamimoment} and MOIRAI~\cite{woo2024unified} adopt masked modeling with patch tokenization to handle diverse and multivariate data, while decoder-based approaches including Timer~\cite{liutimer}, TimesFM~\cite{dasdecoder}, Sundial~\cite{liu2025sundial}, and TIME-MOE~\cite{shi2025time} unify forecasting and related tasks under an autoregressive or generative paradigm, incorporating designs such as flexible patching, probabilistic modeling, and mixture-of-experts to improve scalability and generalization.

\subsection{Datasets}

The 8 public datasets used in this paper are extensively used for long-term TSF algorithm evaluation,
covering multiple fields including industry (4 ETT datasets), climate (Weather), energy (Electricity), transportation (Traffic), and economy (Exchange). The detailed descriptions are as follows:
\begin{enumerate}
\item Electricity dataset\footnote{\url{https://archive.ics.uci.edu/dataset/321/electricity}} collects the electricity consumption (kWh) every 15 minutes of 321 clients from 2012 to 2014.
\item ETT datasets\footnote{\url{https://github.com/zhouhaoyi/Informer2020}} comprises two sub-datasets, ETT1 and ETT2, collected from two separate counties. Each sub-dataset offers two versions with varying sampling resolutions (15 minutes and 1 hour). ETT dataset includes multiple time series of electrical loads and a single time sequence of oil temperature.

\item Weather dataset\footnote{\url{https://www.bgc-jena.mpg.de/wetter/}} contains 21 meteorological indicators, such as air temperature, humidity, etc, recorded every 10 minutes for the entirety of 2020.
\item Exchange dataset\footnote{\url{https://github.com/laiguokun/multivariate-time-series-data}} contains the current exchange of eight countries.

\item Traffic records hourly road occupancy rates measured by 862 sensors of the San Francisco Bay area freeways in 2 years. 

\end{enumerate}

\begin{table}[h!] 
    \setlength{\tabcolsep}{3.5pt}
    \centering
    \caption{Statistics of the eight benchmark datasets. \textit{Time points} denotes the total number of observations. \textit{Split} denotes train/validation/test sizes. \textit{Frequency} denotes the sampling interval.}
    \label{tab:dataset_stat}
    { \small
    \begin{tabular}{c|c|c|c|c}
        \hline
        \multirow{1}{*}{\shortstack{Datasets}}  &  \multicolumn{1}{c|}{Variable} & \multicolumn{1}{c|}Time points &  \multicolumn{1}{c|}{Split (train/val/test)}  & \multicolumn{1}{c}{Frequency} \\
         \midrule[0.5pt]
         \multirow{1}{*}{ETTh1,ETTh2}  &7 &17,420 &\shortstack{(8545, 2881, 2881)} &Hourly\\
        \midrule[0.5pt]
         \multirow{1}{*}{ETTm1,ETTm2}  &7 &69,680 &\shortstack{(34465, 11521, 11521)}  &15min\\
        \midrule[0.5pt]
        \multirow{1}{*}{Weather}  &21 &52,696 &\shortstack{(36792, 5271, 10540)} &10min\\
        \midrule[0.5pt]
        \multirow{1}{*}{Exchange rate}  &8 &7,588 &\shortstack{(5120, 665, 1422)} &Daily\\
                \midrule[0.5pt]
        \multirow{1}{*}{Electricity }  &321 &26,304 &\shortstack{(18317, 2633, 5261)}&Hourly\\

                \midrule[0.5pt]
        \multirow{1}{*}{Traffic }  &862 &17,544 &\shortstack{(12185, 1757, 3509)}&Hourly\\
        \midrule[0.5pt]
    \end{tabular}}
\end{table}

\subsection{Forecasting Model Descriptions}
\label{sec:model_descriptions}

We use two groups of forecasting models as ensemble candidates. All models operate in a univariate setting with input length 96 and prediction length 96.

\subsubsection{Foundation model weight sources}

\begin{itemize}[noitemsep, topsep=2pt]
    \item \textbf{MOIRAI}: A universal time series forecasting transformer pre-trained on large-scale data with masked token prediction. Weights: \texttt{Salesforce/moirai-1.1-R-small}\footnote{\url{https://huggingface.co/Salesforce/moirai-1.1-R-small}}.
    \item \textbf{MOMENT}: A family of foundation models for time series, pre-trained with masked reconstruction on diverse time series corpora. Weights: \texttt{AutonLab/MOMENT-1-large}\footnote{\url{https://huggingface.co/AutonLab/MOMENT-1-large}}.
    \item \textbf{TimeMoE}: A decoder-only Mixture-of-Experts model for time series, using sparse expert routing for multi-horizon forecasting. Weights: \texttt{Maple728/TimeMoE-50M}\footnote{\url{https://huggingface.co/Maple728/TimeMoE-50M}}.
    \item \textbf{TimesFM}: A decoder-only foundation model from Google, pre-trained on 100B real-world time points with patched input tokenization. Weights: \texttt{google/timesfm-2.5-200m-pytorch}\footnote{\url{https://huggingface.co/google/timesfm-2.5-200m-pytorch}}.
    \item \textbf{Timer}: A generative pre-trained Transformer for time series with unified next-token prediction across forecasting tasks. Weights: \texttt{Timer\_forecast\_1.0.ckpt}\footnote{\url{https://github.com/thuml/OpenLTM}}.
    \item \textbf{TimerXL}: An extended version of Timer with larger model capacity. Weights: \texttt{TimerXL\_forecast.pth}\footnote{\url{https://github.com/thuml/OpenLTM}}.
    \item \textbf{Sundial}: A causal language model for time series that discretizes continuous values via adaptive binning. Weights: \texttt{thuml/sundial-base-128m}\footnote{\url{https://huggingface.co/thuml/sundial-base-128m}}.
    \item \textbf{Chronos}: A language modeling framework for time series using quantization-based tokenization, built on T5 architecture. Weights: \texttt{amazon/chronos-t5-base}\footnote{\url{https://huggingface.co/amazon/chronos-t5-base}}.
\end{itemize}

\subsection{Supplementary experimental results}

\subsubsection{Comparison with GRPO algorithm variants}

\begin{table}[t]
    \centering
    \caption{Comparison of GRPO algorithm variants on three datasets (MSE $\downarrow$). All variants use the naive reward ($-\delta$). Visualized in Figure~\ref{fig:reward_combined}(d).}
    \label{tab:grpo_variants}
    { \small
    \begin{tabular}{l|cccccc}
        \toprule
        \textbf{Dataset} & \textbf{GRPO with Reward Mapping} & \textbf{GRPO} & \textbf{DAPO} & \textbf{DrGRPO} & \textbf{GSPO} & \textbf{SAPO} \\
        \midrule
        Exchange & \textbf{0.1587} & 0.1753 & 0.1714 & 0.1804 & 0.1815 & 0.1712 \\
        Electricity & \textbf{0.1463} & 0.1503 & 0.1493 & 0.1475 & 0.1493 & 0.1481 \\
        Weather & \textbf{0.0061} & 0.0167 & 0.0386 & 0.0346 & 0.0324 & 0.0452 \\

        \bottomrule
    \end{tabular}
    }
\end{table}

\begin{table}[t]
\centering
\caption{Average token count of the human input across different data representations.}
\label{tab:token_count}
  
\begin{tabular}{lcc}
\toprule
\textbf{Representation} & \textbf{Input=96} & \textbf{Input=512} \\
\midrule
Raw TS (time series as text) & 1656 & 7480 \\
Textual--Num (structured analysis) & 1085 & 1085 \\
\bottomrule
\end{tabular}
\end{table}

\begin{table*}[t]
    \centering
    \caption{Input ablation study (REATS-SFT, MSE $\downarrow$). (a) In-distribution evaluation on small model candidates. (b) OOD evaluation: trained on small models, tested on foundation models. Best in \textbf{bold}.}
    \label{tab:input_ablation_small}

    \vspace{-6pt}
    \begin{minipage}[t]{0.48\textwidth}
      
    \vspace{0pt}
    \centering
    {\small\textbf{(a) Small Model Candidates (ID)}}\\[2pt]
    \setlength{\tabcolsep}{1.2pt}
    \scriptsize
    \begin{tabular}{l|cccccccc|c}
        \toprule
        \textbf{Method} & \textbf{Exch} & \textbf{H1} & \textbf{H2} & \textbf{M1} & \textbf{M2} & \textbf{Wea} & \textbf{Elec} & \textbf{Traf} & \textbf{Avg} \\
        \midrule
        Textual--Num (\textbf{Ours}) & \textbf{.1888} & .1492 & \textbf{.1769} & .0785 & .0792 & \textbf{.0424} & .1605 & .0925 & \textbf{.1210} \\
        Raw TS (text) & .1988 & .1487 & .1787 & .0770 & .0790 & .0475 & .1595 & .0935 & .1228 \\
        Raw TS (tsenc) & .1968 & \textbf{.1451} & .1778 & \textbf{.0756} & \textbf{.0760} & .0480 & \textbf{.1562} & \textbf{.0922} & \textbf{.1210} \\
        \bottomrule
    \end{tabular}
    \end{minipage}
    \hfill
    \begin{minipage}[t]{0.48\textwidth}
    \vspace{0pt}
    \centering
    {\small\textbf{(b) Small $\to$ Zeroshot (OOD)}}\\[2pt]
    \setlength{\tabcolsep}{1.2pt}
    \scriptsize
  
    \begin{tabular}{l|cccccccc|c}
        \toprule
        \textbf{Method} & \textbf{Exch} & \textbf{H1} & \textbf{H2} & \textbf{M1} & \textbf{M2} & \textbf{Wea} & \textbf{Elec} & \textbf{Traf} & \textbf{Avg} \\
        \midrule
        Textual--Num & .1579 & \textbf{.1442} & \textbf{.1925} & .0892 & .1165 & .0098 & .3972 & \textbf{.3640} & \textbf{.1839} \\
        Raw TS (text) & .1583 & .1450 & .1937 & .0894 & .1183 & .0211 & \textbf{.3967} & .3677 & .1863 \\
        Raw TS (tsenc) & \textbf{.1501} & .1448 & .1945 & \textbf{.0874} & \textbf{.1118} & \textbf{.0089} & .4042 & .6306 & .2165 \\
        \bottomrule
    \end{tabular}
    \end{minipage}
\end{table*}

\subsubsection{OOD model mapping}

\begin{table}[h!]
    \centering
    \caption{Model mapping for OOD generalization experiments (Table~\ref{tab:ood_foundation}). Model 1--4 are unseen foundation models used for each dataset, with varying candidate sets to cover a broader range of foundation models. \textbf{Bold} indicates models not seen during training.}
    \label{tab:ood_model_mapping}
    \setlength{\tabcolsep}{6pt}
    \small
  
    \begin{tabular}{l|cccc}
        \toprule
        Dataset & Model 1 & Model 2 & Model 3 & Model 4 \\
        \midrule
        Exchange & \textbf{MOMENT} & \textbf{TimeMoE} & \textbf{TimesFM} & \textbf{MOIRAI} \\
        ETTh1 & \textbf{MOMENT} & \textbf{Sundial} & \textbf{TimesFM} & \textbf{MOIRAI} \\
        ETTh2 & \textbf{MOMENT} & \textbf{Sundial} & \textbf{TimesFM} & \textbf{MOIRAI} \\
        ETTm1 & \textbf{MOMENT} & \textbf{Sundial} & \textbf{TimesFM} & \textbf{MOIRAI} \\
        ETTm2 & \textbf{MOMENT} & \textbf{Timer} & \textbf{TimeMoE} & \textbf{MOIRAI} \\
        Weather & \textbf{MOMENT} & \textbf{Timer} & \textbf{TimeMoE} & \textbf{TimesFM} \\
        Electricity & \textbf{MOMENT} & \textbf{Sundial} & \textbf{Timer} & \textbf{MOIRAI} \\
        Traffic & \textbf{MOMENT} & \textbf{Sundial} & \textbf{Timer} & \textbf{TimeMoE} \\
        \bottomrule
    \end{tabular}
\end{table}

\subsubsection{Supplementary OOD experiments (Table~\ref{tab:ood_within_combined} and Table~\ref{tab:ood_within_combined_mae})}

\begin{table}[t]
    \centering
    \caption{OOD generalization to mixed models (MSE $\downarrow$). Trained on small model candidates (CARD, LSINet, PDF, TimeXer), evaluated on mixed candidates (2--3 unseen per dataset). \textbf{Bold} in mapping table indicates unseen models.  The MAE results are shown in appendix Table~\ref{tab:ood_combined_mae}.}
    \label{tab:ood_mixed}
    { \small
    \begin{tabular}{l|cccccccc|c}
        \toprule
        \textbf{Method} & \textbf{Exch} & \textbf{H1} & \textbf{H2} & \textbf{M1} & \textbf{M2} & \textbf{Wea} & \textbf{Elec} & \textbf{Traf} & \textbf{Avg} \\
        \midrule
        \multicolumn{10}{l}{\textit{Individual Models}} \\
        Model 1 & .1690 & \textbf{.1515} & 1.8930 & \textbf{.0815} & 1.9402 & .4372 & 1.7326 & \textbf{.0667} & .8090 \\
        Model 2 & .1675 & 1.9515 & \textbf{.2022} & .1124 & \textbf{.0729} & .7114 & .7635 & .9884 & .6212 \\
        Model 3 & .3037 & 1.9124 & 2.0073 & 2.0260 & 1.9540 & .5683 & \textbf{.1472} & 2.1670 & 1.3857 \\
        Model 4 & 1.5737 & .1746 & 1.9351 & 1.8879 & .2862 & .4397 & 1.7626 & 2.1059 & 1.2707 \\
        \midrule
        \multicolumn{10}{l}{\textit{Traditional Ensemble}} \\
        Ens$_{\text{avg}}$ & .2534 & .4671 & .8089 & .4161 & .4371 & .2702 & .6257 & .7310 & .5012 \\
        Ens$_{\text{rand}}$ & .3149 & .5819 & .9487 & .5392 & .5611 & .3393 & .7218 & .8503 & .6071 \\
        InvMSE$_{\text{tr}}$ & .1688 & .1611 & .2467 & .0860 & .0840 & .2739 & .1914 & .0786 & .1613 \\
        InvMSE$_{\text{val}}$ & .2427 & .1922 & .3573 & .1034 & .0829 & .3331 & .2199 & .0765 & .2010 \\
        RLMC & .2562 & .1739 & .2023 & .1087 & .0736 & .2617 & .1513 & .0667 & .1583 \\
        OptW$_{\text{tr}}$ & .1594 & .1632 & .2094 & .0848 & .0792 & .2598 & .1562 & .0827 & .1493 \\
        OptW$_{\text{val}}$ & .1937 & .1732 & .2103 & .1103 & .0792 & .3172 & .1586 & .0825 & .1656 \\
        \midrule
        \multicolumn{10}{l}{\textit{LLM-based Ensemble}} \\
        GPT-5.2 & .1760 & .3551 & .5593 & .2701 & .2608 & .2747 & .3541 & .2109 & .3076 \\
        Codex & .1671 & .3235 & .5333 & .2039 & .2263 & .1939 & .3156 & .1843 & .2685 \\
        GPT-5.5 & .1964 & .3982 & .5998 & .2758 & .3103 & .3162 & .3746 & .2945 & .3457 \\
        DeepSeek-V3.2 & .1724 & .3556 & .5755 & .1904 & .2783 & .1588 & .3074 & .2001 & .2798 \\
        Grok-4 & .1828 & .3646 & .5754 & .2452 & .3072 & .2771 & .4228 & .2986 & .3342 \\
        \midrule
        \multicolumn{10}{l}{\textit{REATS (Ours)}} \\
        \textbf{REATS-GRPO} & \textbf{.1370} & .1798 & .2568 & .0816 & .0761 & \textbf{.0218} & .1522 & .0791 & \textbf{.1231} \\
        \bottomrule
    \end{tabular}
    }
    \par\vspace{4pt}
    {\scriptsize
    {\small \textbf{Model Mapping}}\\[2pt]
    \setlength{\tabcolsep}{8pt}
    \begin{tabular}{l|cccc}
        \toprule
        Dataset & Model 1 & Model 2 & Model 3 & Model 4 \\
        \midrule
        Exchange & CARD & \textbf{MOIRAI} & \textbf{MOMENT} & \textbf{ModernTCN} \\
        ETTh1 & PDF & \textbf{TimeMixer} & TimeXer & \textbf{Timer} \\
        ETTh2 & \textbf{PatchTST} & \textbf{Sundial} & \textbf{TimeMixer} & TimeXer \\
        ETTm1 & LSINet & \textbf{MOMENT} & \textbf{PatchTST} & TimeXer \\
        ETTm2 & \textbf{ModernTCN} & PDF & \textbf{PatchTST} & \textbf{TimesFM} \\
        Weather & CARD & \textbf{MLF} & \textbf{MOMENT} & \textbf{Timer} \\
        Electricity & \textbf{DLinear} & \textbf{MOMENT} & PDF & \textbf{TimeMixer} \\
        Traffic & CARD & \textbf{MOMENT} & \textbf{SEMixer} & \textbf{TimeMixer} \\
        \bottomrule
    \end{tabular}
    }
\end{table}

\subsubsection{Supplementary results about scalability to varying numbers of candidate Models (Table~\ref{tab:scalability_zeroshot_combined} and Table~\ref{tab:scalability_zeroshot_combined_mae})}

\subsection{Discussion of reward mapping function designs}

\subsubsection{Selection of reward scaling factor $k$}
\label{sec:reward_mapping_appendix}

We evaluate the three bounded reward mapping functions that involve a scaling factor $k$ (the naive $r{=}-\delta$ has no such parameter), where $\delta = \mathcal{L}(\mathbf{w}) - \mathcal{L}(\mathbf{w}^*)$ denotes the forecasting MSE gap between the predicted ensemble weights and the oracle weights. We define a unified scoring framework with three criteria:
\begin{enumerate}[noitemsep, topsep=2pt]
    \item \textbf{Near-oracle sensitivity}: $r(0) - r(0.01) \in [0.05, 0.20]$, ensuring the function can distinguish near-optimal candidates without over-sensitivity.
    \item \textbf{Mid-range signal}: $r(0.1) \in [0.20, 0.50]$, maintaining meaningful learning signal while sufficiently penalizing suboptimal generations.
    \item \textbf{Far-range signal}: $r(0.3) > 0.05$, ensuring that poorly-performing generations still receive non-zero reward differences for policy optimization.
\end{enumerate}

\begin{figure*}[h!]
\centerline{\includegraphics[width=\linewidth]{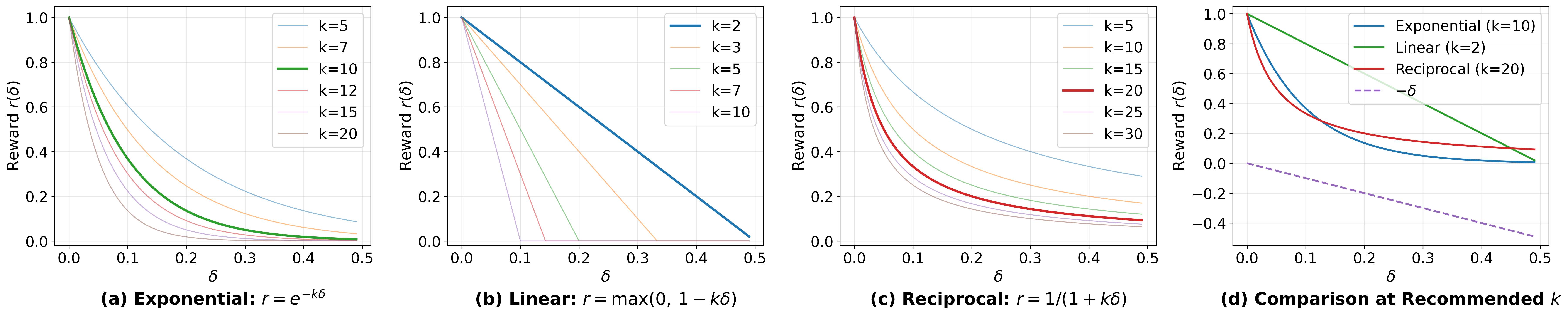} }
\caption{Reward curves under varying $k$ for (a) exponential, (b) linear, and (c) reciprocal mapping functions. Bold lines indicate the selected $k$ values. (d) Comparison of all four mappings at their recommended $k$.}
\label{fig:reward_k_sweep}
\end{figure*}

Figure~\ref{fig:reward_k_sweep} shows the reward curves under varying $k$. By checking which $k$ values simultaneously satisfy all three criteria, we identify the valid range for each function and select $k$ near the center: $k{=}20$ for reciprocal (valid range $[10, 25]$, widest among all three), $k{=}10$ for exponential (valid range $[7, 12]$, narrow), and $k{=}2$ for linear (to maximize coverage before the dead zone at $\delta \geq 1/k$).

\begin{figure}[h!]
\centerline{\includegraphics[width=0.7\linewidth]{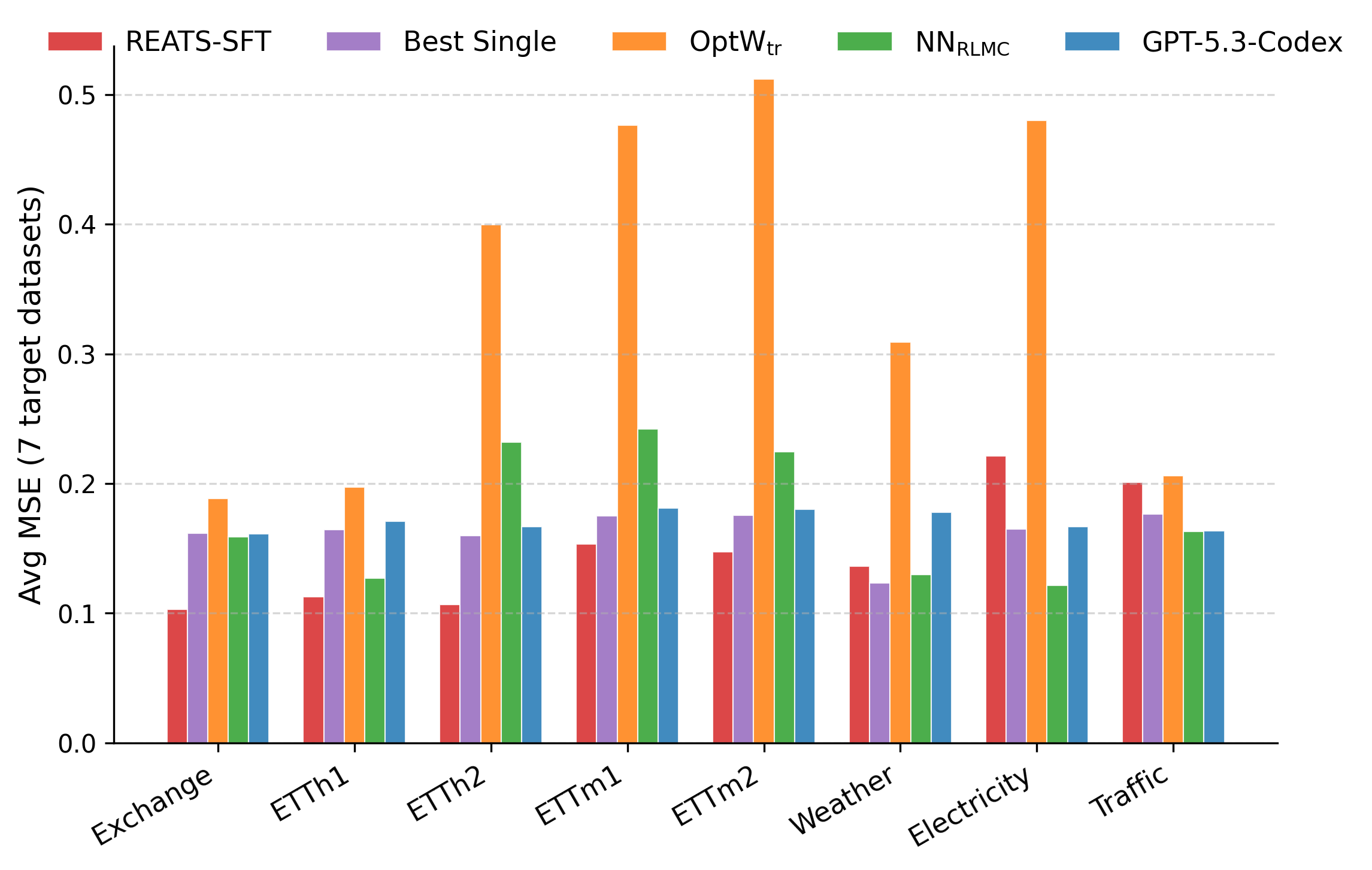} }
\caption{Transfer learning comparison (small model group, MSE $\downarrow$). Each source dataset (X-axis) is used for training, and the average MSE across the other 7 target datasets (Y-axis) is reported. Baselines are the best method from each category.}
\label{fig:transfer_small}
\end{figure}

\subsubsection{Reward Sensitivity Analysis}
\label{sec:reward_sensitivity}
To formalize the discriminative power of each reward function, we analyze their derivatives with respect to $\delta$. By the first-order Taylor expansion:
\begin{equation}
    r(\delta + \Delta\delta) - r(\delta) = r'(\delta)\,\Delta\delta + o(\Delta\delta), \quad \Delta\delta \to 0
\end{equation}
For two nearby points $\delta_1,\delta_2$, the mean value theorem guarantees the existence of $\xi \in (\delta_1,\delta_2)$ such that
\begin{equation}
    r(\delta_1) - r(\delta_2) = r'(\xi)\,(\delta_1 - \delta_2)
\end{equation}
Therefore, the local reward difference can be approximated as:
\begin{equation}
    |r(\delta_1) - r(\delta_2)| \approx \left|\frac{dr}{d\delta}\right| \cdot |\delta_1 - \delta_2|
\end{equation}
This shows that $|dr/d\delta|$ directly quantifies the \emph{local reward sensitivity}: given the same performance difference $|\Delta\delta|$ between two candidates, a larger derivative produces a larger raw reward gap, contributing to better discrimination among candidates in that $\delta$ region.

The derivatives of the three reward functions are:
\begin{align}
    \text{Reciprocal:} \quad \left|\frac{dr}{d\delta}\right| &= \frac{k}{(1 + k\delta)^2} \\
    \text{Exponential:} \quad \left|\frac{dr}{d\delta}\right| &= k \cdot e^{-k\delta} \\
    \text{Linear:} \quad \left|\frac{dr}{d\delta}\right| &= \begin{cases} k & \delta < 1/k \\ 0 & \delta \geq 1/k \end{cases}
\end{align}
At their recommended $k$ values, the three functions exhibit distinct sensitivity profiles:
\begin{itemize}[noitemsep, topsep=2pt]
    \item \textbf{Reciprocal} ($k{=}20$): Maintains effective sensitivity across a wide $\delta$ range (polynomial decay ensures no dead zones even at moderate $\delta$). \textbf{Under GRPO's constraint that the reward mapping should provide signal across the full group diversity, the reciprocal also achieves the highest near-oracle sensitivity} ($\delta < 0.03$), preserving higher raw reward resolution near the oracle as weights approach optimality. In contrast, other functions produce diminished or zero reward signals at moderate $\delta$, limiting their near-oracle $k$ choices.
    \item \textbf{Exponential} ($k{=}10$): Provides stronger discrimination in the mid-range ($\delta \in [0.034, 0.34]$), but its signal vanishes rapidly beyond this range, and weaker near-oracle sensitivity limits the final refinement precision.
    \item \textbf{Linear} ($k{=}2$): Maintains constant sensitivity within its narrow active region $[0, 1/k]$, but offers zero signal beyond the cutoff, severely restricting its effective range.
\end{itemize}

\subsubsection{Theoretical analysis}
\label{sec:reward_theory}

The empirical analysis above demonstrates that the reciprocal mapping outperforms alternatives. In this section, we provide theoretical justification by analyzing how the reward mapping function affects the GRPO learning signal. Since GRPO continues from an SFT-initialized model, the policy starts from a coarse initialization that GRPO must further refine.

\textbf{GRPO signal and reward variance.}
In GRPO, the policy gradient for generation $i$ within a prompt group is scaled by the normalized advantage:
\begin{equation}
    \hat{A}_i = \frac{r_i - \mu_r}{\sigma_r}, \quad \text{where } \mu_r = \frac{1}{G}\sum_j r_j,\;\; \sigma_r = \sqrt{\frac{1}{G}\sum_j (r_j - \mu_r)^2}
\end{equation}
where $G$ is the group size. A necessary condition for reward-driven learning is $\sigma_r > 0$: when all rewards are identical ($\sigma_r = 0$), no reward-driven policy-gradient update arises from that group. To understand how the choice of reward mapping affects $\sigma_r$, we apply a first-order Taylor expansion of $r(\delta)$ around the group mean $\bar{\delta}$:
\begin{equation}
    \sigma_r \approx |r'(\bar{\delta})| \cdot \sigma_\delta
    \label{eq:reward_variance}
\end{equation}
where $\sigma_\delta$ is the standard deviation of $\delta$ values within the generation group (controlled by sampling temperature). This decomposes the GRPO learning signal into two independent factors:
\begin{itemize}[noitemsep, topsep=2pt]
    \item $|r'(\bar{\delta})|$: the reward function's \textbf{local sensitivity} at the current training regime
    \item $\sigma_\delta$: the \textbf{generation diversity} (controlled by sampling hyperparameters)
\end{itemize}
Under the first-order approximation, $|r'(\bar{\delta})| > 0$ and $\sigma_\delta > 0$ yield non-zero local reward dispersion. Moreover, while advantage normalization prevents the raw derivative magnitude from proportionally scaling the learning signal, a mapping with stronger near-oracle sensitivity and bounded range produces better-separated normalized advantages in mixed-quality groups (see Appendix~\ref{sec:naive_reward_analysis} for quantitative analysis).

\textbf{Connection to GRPO's within-group advantage separation.}
The normalized advantage gap between two candidates $i,j$ in a GRPO group is:
\begin{equation}
    \hat{A}_i - \hat{A}_j = \frac{f(\delta_i) - f(\delta_j)}{\sigma_{f(\delta)}}
\end{equation}
For any affine mapping $f(\delta) = -a\delta + b$ ($a>0$), the scaling factor $a$ cancels between numerator and denominator, leaving $\hat{A}_i - \hat{A}_j = (\delta_j - \delta_i)/\sigma_\delta$ regardless of $a$. Thus, affine reward transformations do not alter the normalized advantage gaps within a group. However, for a nonlinear bounded mapping such as $f(\delta) = 1/(1+k\delta)$, both the numerator and $\sigma_{f(\delta)}$ depend nonlinearly on the entire group's $\delta$ distribution, which changes the advantage separation among candidates. In particular, in mixed-quality groups where some $\delta_{\text{bad}} \gg \delta_i, \delta_j$, the unbounded $f{=}-\delta$ allows $\sigma_{f(\delta)}$ to grow without limit, driving $|\hat{A}_i - \hat{A}_j| \to 0$; the bounded reciprocal prevents this compression (see Appendix~\ref{sec:naive_reward_analysis}).

\textbf{Since GRPO generates multiple candidate outputs per prompt via sampling, the $\delta$ values within each group can span a wide range. The reward mapping must maintain adequate sensitivity across this range.} If the mapping's derivative vanishes in certain $\delta$ regions (``dead zones''), two failure cases arise: (1)~if \emph{all} group members fall in the dead zone, all rewards become identical, $\sigma_r = 0$, and no reward-driven learning signal; (2)~if \emph{some} group members fall in the dead zone, their rewards collapse to indistinguishable values, and those generations provide no reward-driven learning signal.

\textbf{Derivative comparison.} We compare the sensitivity $|r'(\delta)|$ of each mapping:
\begin{align}
    |r'_{\text{rec}}(\delta)| &= \frac{k}{(1+k\delta)^2} = O(1/\delta^2) \quad \text{(polynomial decay)} \\
    |r'_{\text{exp}}(\delta)| &= k \cdot e^{-k\delta} = O(e^{-k\delta}) \quad \text{(exponential decay)} \\
    |r'_{\text{lin}}(\delta)| &= \begin{cases} k & \text{if } \delta < 1/k \\ 0 & \text{if } \delta \geq 1/k \end{cases} \quad \text{(hard cutoff)}
\end{align}
The three mappings exhibit fundamentally different decay behaviors. The reciprocal decays polynomially, meaning its sensitivity decreases slowly and all samples in the group remain distinguishable in raw reward space regardless of their $\delta$ value. The exponential decays much faster, causing samples with moderate-to-large $\delta$ to receive near-identical rewards and become indistinguishable. The linear has a hard cutoff at $\delta = 1/k$, beyond which sensitivity is exactly zero and all such samples are mapped to identical rewards.

\textbf{Absence of a hard reward dead zone.} We formally verify that the reciprocal mapping maintains distinct rewards for any pair of non-identical $\delta$ values. For $r(\delta) = \frac{1}{1+k\delta}$ with $k > 0$, since $(1+k\delta)^2 > 0$ for all $\delta \geq 0$, \textbf{the derivative $|r'_{\text{rec}}(\delta)| = k/(1+k\delta)^2 > 0$ for all $\delta \geq 0$}. Because $r(\delta)$ is strictly monotonically decreasing, $\delta_i \neq \delta_j \implies r(\delta_i) \neq r(\delta_j)$, so any generation group with $\sigma_\delta > 0$ yields $\sigma_r > 0$ regardless of $\bar{\delta}$.

\textbf{Positive lower bound on reward dispersion.} Beyond the qualitative guarantee above, we derive a quantitative lower bound on $\sigma_r$. Let $D = \max_{i,j}|r(\delta_i) - r(\delta_j)|$ be the maximum reward difference in the group. Since $\mu_r$ lies between $\min_i r_i$ and $\max_i r_i$, at least one term satisfies $(r_i - \mu_r)^2 \geq D^2/4$, giving:
\begin{equation}
    \sigma_r^2 = \frac{1}{G}\sum_i (r_i - \mu_r)^2 \geq \frac{D^2}{4G}
    \implies \sigma_r \geq \frac{D}{2\sqrt{G}}
\end{equation}
To bound $D$ from below, we apply the mean value theorem: for any $\delta_i, \delta_j$, there exists $\xi \in [\delta_i, \delta_j]$ such that:
\begin{equation}
    |r(\delta_i) - r(\delta_j)| = |r'(\xi)| \cdot |\delta_i - \delta_j|
\end{equation}
Since $|r'_{\text{rec}}(\delta)| = k/(1+k\delta)^2$ is monotonically decreasing and $\xi \leq \delta_{\max}$:
\begin{equation}
    |r'(\xi)| \geq |r'(\delta_{\max})| = \frac{k}{(1+k\delta_{\max})^2}
\end{equation}
Combining the above:
\begin{equation}
    \sigma_r \geq \frac{k}{2\sqrt{G}\,(1+k\delta_{\max})^2} \cdot \max_{i \neq j} |\delta_i - \delta_j|
\end{equation}

This lower bound confirms that \textbf{the reciprocal mapping maintains adequate sensitivity at both small $\delta$ (where $|r'| \approx k$) and moderate $\delta$ (where $|r'| = k/(1+k\delta)^2$ remains positive)}, providing effective discrimination across the full $\delta$ range encountered during GRPO training.

\textbf{Failure modes of alternative mappings.} Applying the same mean value theorem-based derivation to the alternative mappings yields analogous bounds that reveal their limitations. For the \textbf{exponential} mapping $r_{\text{exp}}(\delta) = e^{-k\delta}$, since $|r'_{\text{exp}}(\xi)| = ke^{-k\xi}$ is monotonically decreasing, $|r'_{\text{exp}}(\xi)| \geq k \cdot e^{-k\delta_{\max}}$ for $\xi \leq \delta_{\max}$, the analogous bound is:
\begin{equation}
    \sigma_r \geq \frac{k \cdot e^{-k\delta_{\max}}}{2\sqrt{G}} \cdot \max_{i \neq j} |\delta_i - \delta_j|
\end{equation}
The factor $e^{-k\delta_{\max}}$ decays exponentially, making this bound vanishingly small even at moderate $\delta_{\max}$ (e.g., with $k\delta_{\max} = 5$, the factor is $\approx 0.007$). Practically, effective raw-reward saturation ($|r'| < 0.01$) begins at $\delta \approx \ln(100k)/k$, beyond which all samples become effectively indistinguishable.

For the \textbf{linear} mapping $r_{\text{lin}}(\delta) = \max(0, 1-k\delta)$, when $\delta_{\max} > 1/k$, the derivative $|r'_{\text{lin}}(\xi)| = 0$ for $\xi > 1/k$, so the same bound degenerates to:
\begin{equation}
    \sigma_r \geq 0
\end{equation}
which is trivially true and provides no guarantee. When $\min_i \delta_i > 1/k$, all rewards are exactly zero and $\sigma_r = 0$.

In contrast, the reciprocal's bound has only polynomial decay $(1+k\delta_{\max})^2$ in the denominator, ensuring a strictly positive lower bound for any finite $\delta_{\max}$. Among the three mappings considered and under the rollout ranges studied, the reciprocal is the most robust: it decays polynomially rather than exponentially, avoiding both the hard dead zone of the linear mapping and the rapid effective saturation of the exponential mapping.

\subsubsection{Why naive reward $r=-\delta$ fails in practice}
\label{sec:naive_reward_analysis}

While the naive mapping $r(\delta) = -\delta$ does not suffer from dead-zone issues, it introduces two problems under GRPO's advantage normalization $\hat{A}_i{=}(r_i{-}\mu_r)/\sigma_r$: (1)~as a linear function, it produces uniform advantage spacing regardless of candidate quality, whereas the convex reciprocal concentrates discrimination among near-optimal candidates; (2)~its unbounded range allows a single outlier to dominate $\sigma_r$, compressing all near-optimal advantages to near-identical values.

\begin{center}
\small
\begin{tabular}{lcc}
\toprule
& $r = -\delta$ & Reciprocal $r = 1/(1+20\delta)$ \\
\midrule
Top-2 gap ($\delta$=0.001 vs 0.01) & $0.009$ & $0.147$ \\
Bottom-2 gap ($\delta$=0.5 vs 2.0) & $1.500$ & $0.067$ \\
Normalized $\hat{A}_1 - \hat{A}_2$ & $0.012$ & $0.379$ \\
\bottomrule
\end{tabular}
\end{center}
In a realistic mixed-quality group ($\delta \in \{0.001, 0.01, 0.1, 0.5, 2.0\}$), the reciprocal achieves $32\times$ larger normalized advantage separation between the best two candidates, because it simultaneously bounds outlier rewards near zero (preventing $\sigma_r$ inflation) and nonlinearly stretches near-oracle differences. This effect cannot be replicated by any affine transformation of $r{=}-\delta$.

\begin{table*}[h!]
\centering
\footnotesize
\caption{Summary of the time-series characteristic analysis tools.}
\label{tab:ts_statistics_summary_full}
 \vspace{-6pt}
\begin{tabular}{p{2cm} p{3cm} p{7.5cm}}
\toprule
\textbf{Feature Group} & \textbf{Key features} & \textbf{Description} \\
\midrule

\multirow{2}{*}{Stationarity $\phi_{\text{sta}}$}
& ADF test result & Tests whether the series contains a unit root, i.e., non-stationary time series. \\
& KPSS test result & Tests whether a series is stationary around its “trend” or “mean”. \\

\midrule
\multirow{3}{*}{Noise $\phi_{\text{noise}}$}
& Coefficient of variation & Captures relative variability compared to the mean level. \\
& Noise proportion & Estimates how much variance is attributed to irregular components. \\
& Noise intensity & Reflects predictability difficulty due to random fluctuations. \\

\midrule
\multirow{4}{*}{Trend $\phi_{\text{trend}}$}
& Trend slope & Measures long-term directional movement over time. \\
& R$^2$ (Goodness of fit)& Indicates how well a linear trend explains the series. \\
& Trend type  & Identifies whether a clear increasing or decreasing pattern exists. \\
& Trend strength & Quantifies the dominance of trend relative to fluctuations. \\

\midrule
\multirow{3}{*}{Seasonality $\phi_{\text{sea}}$}
& Period length & Indicates repeating temporal cycle length. \\
& Seasonal variance ratio & Measures contribution of periodic components. \\
& Seasonality strength & Assesses whether periodic patterns are dominant. \\

\midrule
\multirow{4}{*}{Autocorrelation $\phi_{\text{ac}}$}
& Significant ACF/PACF lags & The number of lags outside the confidence interval in the ACF (Autocorrelation Function) and Partial ACF. \\
& ACF/PACF coefficients  & Autocorrelation coefficients at each lag. \\
& Autocorrelation judgment & Evaluates overall temporal dependency strength. \\

\midrule
\multirow{6}{*}{Statistics $\phi_{\text{stat}}$}
& Mean / Median & Describe central tendency and distribution symmetry. \\
& Standard deviation & Measures overall dispersion of values. \\
& Minimum / Maximum & Define the range of observed values. \\
& Quantiles (Q1, Q2, Q3) & Describe distribution spread and skewness. \\
& Percentiles & Capture tail behavior of the distribution. \\
& Central tendency consistency & Assesses symmetry between mean and median. \\
\midrule
\multirow{5}{*}{IQR Outliers $\phi_{\text{out}}$}
& Lower threshold & Defines lower bound for outlier detection. \\
& Upper threshold & Defines upper bound for outlier detection. \\
& Lower outlier count & Counts extreme low values. \\
& Upper outlier count & Counts extreme high values. \\
& Outlier ratio & Reflects prevalence of rare extreme events. \\

\midrule
\multirow{4}{*}{Distribution $\phi_{\text{dist}}$}
& Normality test & Assesses deviation from Gaussian assumptions. \\
& Skewness & Measures asymmetry of the distribution. \\
& Kurtosis & Captures tail heaviness and peak sharpness. \\
& Distribution type & Summarizes overall distribution shape. \\

\bottomrule
\end{tabular}
\end{table*}

\subsection{Derivation of Closed-Form Oracle Weights}
\label{app:oracle_derivation}

Given $N$ base forecasters, let $\hat{\mathbf{y}}_i \in \mathbb{R}^T$ denote the prediction of model $i$ and $\mathbf{y} \in \mathbb{R}^T$ the ground truth, where $T$ is the prediction horizon. We define the per-model error as $\mathbf{e}_i = \hat{\mathbf{y}}_i - \mathbf{y}$.

The ensemble prediction under weights $\mathbf{w}$ is $\hat{\mathbf{y}} = \sum_{i=1}^M w_i \hat{\mathbf{y}}_i$. Since $\sum_{i=1}^N w_i = 1$, we can write $\mathbf{y} = \sum_{i=1}^M w_i \mathbf{y}$, and the ensemble error becomes:
\begin{equation}
    \mathbf{e}_{\text{ens}} = \hat{\mathbf{y}} - \mathbf{y} = \sum_{i=1}^M w_i \hat{\mathbf{y}}_i - \sum_{i=1}^M w_i \mathbf{y} = \sum_{i=1}^M w_i (\hat{\mathbf{y}}_i - \mathbf{y}) = \sum_{i=1}^M w_i \mathbf{e}_i
\end{equation}
The ensemble MSE is:
\begin{align}
    \mathcal{L}(\mathbf{w}) &= \frac{1}{T} \sum_{t=1}^{T} \left(\sum_{i=1}^M w_i \, e_{i,t}\right)^2 \notag \\
    &= \frac{1}{T} \sum_{t=1}^{T} \sum_{i=1}^M \sum_{j=1}^M w_i \, w_j \, e_{i,t} \, e_{j,t} \notag \\
    &= \sum_{i=1}^M \sum_{j=1}^M w_i \, w_j \underbrace{\frac{1}{T}\sum_{t=1}^{T} e_{i,t} \, e_{j,t}}_{G_{ij}} \notag \\
    &= \mathbf{w}^\top \mathbf{G} \mathbf{w}
\end{align}
where $\mathbf{G} \in \mathbb{R}^{M \times M}$ with $G_{ij} = \frac{1}{T}\sum_{t=1}^{T} e_{i,t} \, e_{j,t}$. Thus, minimizing the ensemble MSE is equivalent to minimizing the quadratic form $\mathbf{w}^\top \mathbf{G} \mathbf{w}$.

We minimize $\mathbf{w}^\top \mathbf{G} \mathbf{w}$ subject to the constraint $\sum_{i=1}^M w_i = 1$. Introducing a Lagrange multiplier $\lambda$:
\begin{equation}
    \mathcal{J}(\mathbf{w}, \lambda) = \mathbf{w}^\top \mathbf{G} \mathbf{w} - \lambda (\mathbf{1}^\top \mathbf{w} - 1)
\end{equation}

Taking the derivative with respect to $\mathbf{w}$ and setting it to zero:
\begin{equation}
    \frac{\partial \mathcal{J}}{\partial \mathbf{w}} = 2\mathbf{G}\mathbf{w} - \lambda \mathbf{1} = \mathbf{0} \quad \Longrightarrow \quad \mathbf{w} = \frac{\lambda}{2} \mathbf{G}^{-1} \mathbf{1}
\end{equation}

Substituting into the constraint $\mathbf{1}^\top \mathbf{w} = 1$:
\begin{equation}
    \mathbf{1}^\top \left(\frac{\lambda}{2} \mathbf{G}^{-1} \mathbf{1}\right) = 1 \quad \Longrightarrow \quad \frac{\lambda}{2} = \frac{1}{\mathbf{1}^\top \mathbf{G}^{-1} \mathbf{1}}
\end{equation}

Substituting back:
\begin{equation}
    \mathbf{w}^* = \frac{\mathbf{G}^{-1} \mathbf{1}}{\mathbf{1}^\top \mathbf{G}^{-1} \mathbf{1}}
\end{equation}
This is computed independently for each sample (each sample has its own $\mathbf{G}$).

Since per-sample $\mathbf{G}$ may be ill-conditioned, we regularize $\mathbf{G} \leftarrow \mathbf{G} + \epsilon \mathbf{I}$ (with $\epsilon = 10^{-10}$). We then solve the simplex-constrained QP $\mathbf{w}^* = \arg\min_{\mathbf{w} \geq 0,\, \mathbf{1}^\top\mathbf{w}=1} \mathbf{w}^\top \mathbf{G} \mathbf{w}$ to obtain the exact non-negative optimal weights. The closed-form solution above serves as the initial point.

\subsection{Diverse Weight Selection Procedure}
\label{app:diverse_weight}

As discussed in the main text, since relying solely on the single optimum $\mathbf{w}^*$ as supervision may provide insufficient supervision signals and degrade fine-tuning performance, we generate $K'{-}1{=}9$ diverse yet high-quality weight vectors per sample. We sample candidates via a Dirichlet distribution centered on $\mathbf{w}^*$ and select $K$ vectors through a multi-objective greedy procedure that minimizes:
\begin{equation}
    s(\mathbf{w}) = \frac{\mathcal{L}(\mathbf{w}) - \mathcal{L}^*}{\mathcal{L}^*} - d_{\min}(\mathbf{w}, \mathcal{S}) - H(\mathbf{w})
\end{equation}
where $\mathcal{L}(\mathbf{w})$ is the ensemble MSE under weight $\mathbf{w}$, $\mathcal{L}^*$ is the optimal MSE, $d_{\min}(\mathbf{w}, \mathcal{S}) = \min_{\mathbf{w}' \in \mathcal{S}} \|\mathbf{w} - \mathbf{w}'\|_2$ is the minimum $\ell_2$ distance to the already-selected set $\mathcal{S}$, and $H(\mathbf{w}) = -\sum_i w_i \log w_i$ encourages smooth non-sparse distributions.

This diverse supervision enables the LLM to learn a distribution of reasonable weight configurations rather than overfitting to a single solution, improving generalization to unseen time series patterns. For SFT, it exposes the model to multiple valid solutions per sample, preventing memorization of a single weight pattern; for GRPO, it provides row-wise oracle references (Section~\ref{sec:training}) that enable fine-grained per-row reward computation, yielding richer and more informative gradient signals than a single-row comparison.

\subsection{SFT Training Data Example}
\label{sec:sft_example}

We provide a representative SFT training sample below to illustrate the input--output format. Each sample consists of a system prompt, a structured user input, and the model's chain-of-thought response with ensemble weights.

\smallskip
\noindent\textbf{System prompt:}
\begin{quote}
\small\texttt{You are a time series ensemble expert. Generate a weight table (K' rows $\times$ N candidate models (tools)) for ensemble. Each row: all weights non-negative and sum to 1.}
\end{quote}

\noindent\textbf{User input} (structured prompt with dataset context, series analysis, tool descriptions, and RAG-retrieved similar samples):
\begin{quote}
\small\texttt{Assign ensemble weights to the forecasting tools based on the information below.}

\smallskip
\small\texttt{Dataset: Exchange (daily exchange rates of 8 countries). Task: predict the next 96 values from the historical data.}

\smallskip
\small\texttt{Historical series analysis:}

\small\texttt{[stationarity] ADF p-value: 0.69, ADF stationary: False, KPSS stationary: False, Overall judgment: Non-stationary}

\small\texttt{[trend] Trend slope: 0.0113, R\textsuperscript{2}: 93\%, Trend type: Upward trend, Trend strength: Strong}

\small\texttt{[noise] Noise ratio: 18\%, Noise strength: Low noise}

\small\texttt{[autocorrelation] Dominant ACF lag: 1, Dominant ACF value: 0.972, Judgment: Significant autocorrelation exists}

\small\texttt{[distribution] Normal: False, Skewness: -0.089, Kurtosis: -1.53, Type: Left-skewed (non-normal)}

\small\texttt{[seasonality] ... \quad [statistics] ... \quad [outliers] ...}

\smallskip
\small\texttt{Available forecasting tools:}

\small\texttt{TimeXer: decoder-only Transformer with patch tokenization and causal masking.}

\small\texttt{PDF: FFT-based periodicity decoupling with parallel Transformer and conv branches.}

\small\texttt{CARD: channel-dependent Transformer with cross-channel attention and token blending.}

\small\texttt{LSINet: lightweight linear model with sparse temporal connections via Bernoulli interaction.}

\smallskip
\small\texttt{Similar sample analysis: [stationarity] Non-stationary, [trend] Strong upward, [noise] Low ...}

\small\texttt{Optimal ensemble weights(\%) on the most similar time series:}

\small\texttt{CARD,LSINet,PDF,TimeXer}

\small\texttt{4,33,35,28 | 6,43,25,26 | 5,39,25,31 | ...}
\end{quote}

\noindent\textbf{Model output} (chain-of-thought reasoning followed by weight predictions):
\begin{quote}
\small\texttt{<think>}

\small\texttt{Key patterns: non-stationary series, pronounced upward drift, clean signal with little noise, extremely strong short-lag persistence, no outliers, left-skewed distribution.}

\small\texttt{Tool match: PDF aligns strongly since FFT periodicity decoupling fits the series; LSINet contributes through strong trend favoring linear decomposition; CARD and TimeXer have limited relevance.}

\small\texttt{Reference: similar-series ensembles lean toward LSINet with moderate PDF, but current characteristics better suit PDF.}

\small\texttt{Conclusion: prioritize PDF, hedge with LSINet, token allocation to CARD and TimeXer.}

\small\texttt{</think>}

\small\texttt{Optimal ensemble weights(\%) on the historical time series:}

\small\texttt{CARD,LSINet,PDF,TimeXer}

\small\texttt{5,12,71,12 | 13,40,33,14 | 19,18,48,15 | 15,29,42,14 | ...}
\end{quote}

\subsection{More discussions about Interpretability Analysis and GRPO Reasoning Improvement}
\label{sec:interpretability}

\subsubsection{Chain-of-thought interpretability}

Both REATS-SFT and REATS-GRPO generate structured chain-of-thought reasoning before producing ensemble weights. The reasoning follows a consistent four-step structure: (1) summarize key series patterns, (2) match patterns to tool capabilities, (3) reference similar historical cases from RAG, and (4) state the allocation conclusion. This structure enables users to understand \emph{why} specific tools receive higher weights, making the ensemble decision transparent and auditable.

We illustrate with a Weather test sample (non-stationary, downward trend, low noise, ACF lag-1 = 0.987):

\smallskip
\noindent\textbf{REATS-SFT:}
\begin{quote}
\small\texttt{Key patterns: non-stationary, marked downward trajectory, very little noise, extremely strong short-lag persistence.}

\small\texttt{Tool match: PDF fits best because FFT periodicity decoupling aligns with series characteristics; CARD can help modestly through attention exploiting autocorrelation.}

\small\texttt{Conclusion: allocate heavily to PDF, keep moderate weight on CARD.}

\small\texttt{Weights: CARD=32\%, LSINet=5\%, PDF=56\%, TimeXer=7\%}
\end{quote}

\noindent\textbf{REATS-GRPO:}
\begin{quote}
\small\texttt{Key patterns: non-stationary, marked downward trajectory, very little noise, extremely strong short-lag persistence.}

\small\texttt{Tool match: CARD is well-suited here as attention can exploit autocorrelation structure; PDF adds complementary value via FFT periodicity decoupling.}

\small\texttt{Conclusion: prioritize CARD given its fit, hedge with PDF, LSINet.}

\small\texttt{Weights: CARD=53\%, LSINet=18\%, PDF=24\%, TimeXer=5\%}
\end{quote}

Both outputs provide clear, interpretable reasoning that links observed series properties to specific tool capabilities, demonstrating the inherent interpretability of the REATS framework regardless of the training stage.

\subsubsection{GRPO reasoning improvement over SFT}

Comparing the two outputs above reveals that GRPO refines the feature-to-tool mapping through RL exploration. The oracle weights for this sample are: CARD=49\%, PDF=38\%, TimeXer=7\%, LSINet=6\%.

The dominant characteristic of this series is its extremely strong autocorrelation (0.987), which favors attention-based models that can directly exploit lag dependencies. GRPO correctly identifies this and allocates CARD as the primary tool (53\% vs.\ oracle 49\%). SFT instead defaults to PDF (56\%), which targets periodicity, a less relevant property for this autocorrelation-dominated series. This shows that GRPO, through reward-driven exploration, learns more accurate reasoning pathways that better connect series characteristics with appropriate tool capabilities.

\begin{table*}[t]
    \centering
    \caption{Ensemble forecasting results (MAE $\downarrow$). Dataset abbreviations: Exch=Exchange, H1/H2=ETTh1/ETTh2, M1/M2=ETTm1/ETTm2, Wea=Weather, Elec=Electricity, Traf=Traffic.}
    \label{tab:main_res_mae_combined}
     \vspace{-6pt}
    \begin{minipage}[t]{0.49\textwidth}
    \vspace{0pt}
    \centering
    {\small \textbf{(a) Foundation Model Candidates}}\\[2pt]
    \resizebox{\textwidth}{!}{
    \begin{tabular}{l|cccccccc|c}
        \toprule
        \textbf{Method} & \textbf{Exch} & \textbf{H1} & \textbf{H2} & \textbf{M1} & \textbf{M2} & \textbf{Wea} & \textbf{Elec} & \textbf{Traf} & \textbf{Avg} \\
        \midrule
        \multicolumn{10}{l}{\textit{Individual Models}} \\
    MOIRAI & .2810 & .3476 & .3588 & .2662 & .3957 & .0584 & .5800 & .2695 & .3196 \\
        MOMENT & .3845 & .3428 & .3990 & .2516 & .3140 & .0922 & .7105 & .8640 & .4198 \\
        TimeMoE & .3341 & .3199 & .3238 & .2311 & .2731 & .0586 & .4489 & .1894 & .2724 \\
        TimesFM & .2984 & .3117 & .3423 & .2506 & .3575 & .0431 & .5040 & .1932 & .2876 \\
        \midrule
        \multicolumn{10}{l}{\textit{Traditional Ensemble}} \\
        Ens$_{\text{avg}}$ & .3026 & .3060 & .3338 & .2300 & .3046 & .0602 & .5145 & .3160 & .2960 \\
        Ens$_{\text{rand}}$ & .3071 & .3114 & .3386 & .2341 & .3112 & .0608 & .5248 & .3295 & .3022 \\
        InvMSE$_{\text{tr}}$ & .2950 & .3057 & .3320 & .2292 & .2930 & .0569 & .4961 & .1973 & .2756 \\
        InvMSE$_{\text{val}}$ & .3016 & .3057 & .3323 & .2292 & .2928 & .0479 & .4958 & .1958 & .2751 \\
        RLMC & .2979 & .3045 & .3272 & .2229 & .3016 & .0409 & .4488 & .1885 & .2666 \\
        OptW$_{\text{tr}}$ & .2833 & .3042 & .3240 & .2250 & .2748 & .0540 & .4544 & .1967 & .2645 \\
        OptW$_{\text{val}}$ & .3071 & .3049 & .3262 & .2249 & .2747 & .0455 & .4541 & .1968 & .2668 \\
        \midrule
        \multicolumn{10}{l}{\textit{LLM-based Ensemble}} \\
        GPT-5.2 & .2728 & .2915 & .3193 & .2194 & .2850 & .0492 & .4741 & .2244 & .2670 \\
        Codex & .2701 & .2894 & .3189 & .2167 & .2783 & .0514 & .4728 & .2217 & .2649 \\
        GPT-5.5 & .2756 & .2954 & .3234 & .2220 & .2916 & .0566 & .4822 & .2420 & .2736 \\
        DeepSeek-V3.2 & .2631 & .2868 & .3182 & .2155 & .2783 & .0498 & .4739 & .2408 & .2658 \\
        Grok-4 & .2780 & .2950 & .3249 & .2224 & .2909 & .0592 & .4941 & .2605 & .2781 \\
        \midrule
        \multicolumn{10}{l}{\textit{REATS (Ours)}} \\
        REATS-SFT & .2545 & .2815 & .3125 & .2116 & .2681 & .0391 & .4521 & .1968 & .2520 \\
        \textbf{REATS-GRPO} & \textbf{.2464} & \textbf{.2751} & \textbf{.3031} & \textbf{.2071} & \textbf{.2597} & \textbf{.0367} & \textbf{.4397} & \textbf{.1839} & \textbf{.2440} \\
        \bottomrule
    \end{tabular}
    }
    \end{minipage}
    \hfill
    \begin{minipage}[t]{0.49\textwidth}
    \vspace{0pt}
    \centering
    {\small \textbf{(b) Small Model Candidates}}\\[2pt]
    \resizebox{\textwidth}{!}{
    \begin{tabular}{l|cccccccc|c}
        \toprule
        \textbf{Method} & \textbf{Exch} & \textbf{H1} & \textbf{H2} & \textbf{M1} & \textbf{M2} & \textbf{Wea} & \textbf{Elec} & \textbf{Traf} & \textbf{Avg} \\
        \midrule
        \multicolumn{10}{l}{\textit{Individual Models}} \\
      CARD & .2849 & .3087 & .3278 & .2289 & .2535 & .0608 & .4929 & .1708 & .2660 \\
        LSINet & .4962 & .2992 & .3137 & .2084 & .1993 & .1165 & .3730 & 1.1956 & .4002 \\
        PDF & .5316 & .2958 & .3183 & .2022 & .1877 & .0919 & .2810 & 1.1898 & .3873 \\
        TimeXer & .9606 & 1.0635 & 1.1316 & 1.0863 & 1.1301 & .0827 & 1.0080 & 1.1635 & .9533 \\
        \midrule
        \multicolumn{10}{l}{\textit{Traditional Ensemble}} \\
        Ens$_{\text{avg}}$ & .4306 & .3842 & .4135 & .3330 & .3456 & .0821 & .4291 & .7684 & .3983 \\
        Ens$_{\text{rand}}$ & .4637 & .4125 & .4423 & .3570 & .3695 & .0840 & .4544 & .8007 & .4230 \\
        InvMSE$_{\text{tr}}$ & .3143 & .2955 & .3164 & .2050 & .1961 & .0830 & .3061 & .1965 & .2391 \\
        InvMSE$_{\text{val}}$ & .4237 & .3053 & .3493 & .2169 & .1961 & .0811 & .3169 & .1923 & .2602 \\
        RLMC & .4439 & .3005 & .4371 & .2277 & .1993 & .0765 & .2888 & \textbf{.1707} & .2681 \\
        OptW$_{\text{tr}}$ & .2824 & .2996 & .3163 & .2055 & .1930 & .0821 & .2812 & .2197 & .2350 \\
        OptW$_{\text{val}}$ & .3658 & .3050 & .3280 & .2258 & .1912 & .0810 & .2860 & .2195 & .2503 \\
        \midrule
        \multicolumn{10}{l}{\textit{LLM-based Ensemble}} \\
        GPT-5.2 & .3745 & .3286 & .3518 & .2568 & .2563 & .0634 & .3439 & .4405 & .3020 \\
        Codex & .3572 & .3146 & .3422 & .2417 & .2438 & .0636 & .3332 & .3826 & .2849 \\
        GPT-5.5 & .3839 & .3347 & .3570 & .2635 & .2620 & .0686 & .3551 & .5058 & .3163 \\
        DeepSeek-V3.2 & .3581 & .3109 & .3371 & .2395 & .2386 & .0615 & .3214 & .3742 & .2802 \\
        Grok-4 & .3837 & .3307 & .3604 & .2653 & .2679 & .0750 & .3529 & .5550 & .3239 \\
        \midrule
        \multicolumn{10}{l}{\textit{REATS (Ours)}} \\
        REATS-SFT & .3028 & .2916 & .3144 & .2087 & .2019 & .0472 & .2925 & .2316 & .2363 \\
        \textbf{REATS-GRPO} & \textbf{.2802} & \textbf{.2823} & \textbf{.3071} & \textbf{.1964} & \textbf{.1865} & \textbf{.0406} & \textbf{.2800} & .2117 & \textbf{.2231} \\
        \bottomrule
    \end{tabular}
    }
    \end{minipage}
\end{table*}

\begin{table*}[t]
    \centering
    \caption{Generalization study across tool groups (MAE $\downarrow$). The model is trained on small model candidates (CARD, LSINet, PDF, TimeXer). (a) Evaluated on \textbf{foundation model} candidates (all 4 unseen). (b) Evaluated on \textbf{mixed model} candidates (2--3 unseen per dataset). \textbf{Bold} in mapping tables indicates unseen models not present during training. Dataset abbreviations: Exch=Exchange, H1/H2=ETTh1/ETTh2, M1/M2=ETTm1/ETTm2, Wea=Weather, Elec=Electricity, Traf=Traffic.}
    \label{tab:ood_combined_mae}
     \vspace{-6pt}
    \begin{minipage}[t]{0.49\textwidth}
    \vspace{0pt}
    \centering
    {\small \textbf{(a) OOD to Foundation Models}}\\[2pt]
    \resizebox{\textwidth}{!}{
    \begin{tabular}{l|cccccccc|c}
        \toprule
        \textbf{Method} & \textbf{Exch} & \textbf{H1} & \textbf{H2} & \textbf{M1} & \textbf{M2} & \textbf{Wea} & \textbf{Elec} & \textbf{Traf} & \textbf{Avg} \\
        \midrule
        \multicolumn{10}{l}{\textit{Individual Models}} \\
        Model 1 & .3845 & .3428 & .3990 & .2516 & .3140 & .0922 & .7105 & .8640 & .4198 \\
        Model 2 & .3341 & .3169 & .3283 & .2304 & .2461 & .0786 & \textbf{.4383} & .1943 & .2709 \\
        Model 3 & .2984 & .3117 & .3423 & .2506 & .2731 & .0586 & .5764 & .3245 & .3045 \\
        Model 4 & .2810 & .3476 & .3588 & .2662 & .3957 & .0431 & .5800 & .1894 & .3077 \\
        \midrule
        \multicolumn{10}{l}{\textit{Traditional Ensemble}} \\
        Ens$_{\text{avg}}$ & .3025 & .3053 & .3349 & .2305 & .2734 & .0652 & .5310 & .3408 & .2980 \\
        Ens$_{\text{rand}}$ & .3073 & .3106 & .3395 & .2343 & .2806 & .0656 & .5409 & .3521 & .3039 \\
        InvMSE$_{\text{tr}}$ & .2950 & .3050 & .3332 & .2298 & .2614 & .0603 & .5089 & .2109 & .2756 \\
        InvMSE$_{\text{val}}$ & .3016 & .3051 & .3337 & .2298 & .2615 & .0452 & .5086 & .2087 & .2743 \\
        RLMC & .2859 & .3055 & .3368 & .2228 & .2461 & .0730 & \textbf{.4383} & \textbf{.1839} & .2615 \\
        OptW$_{\text{tr}}$ & .2833 & .3036 & .3271 & .2257 & .2441 & .0579 & .4466 & .1980 & .2608 \\
        OptW$_{\text{val}}$ & .3071 & .3052 & .3312 & .2259 & .2447 & .0466 & .4466 & .1988 & .2633 \\
        \midrule
        \multicolumn{10}{l}{\textit{LLM-based Ensemble}} \\
        GPT-5.2 & .2744 & .2908 & .3220 & .2204 & .2436 & .0535 & .4840 & .2507 & .2674 \\
        Codex & .2695 & .2873 & .3212 & .2171 & .2409 & .0560 & .4839 & .2365 & .2641 \\
        GPT-5.5 & .2815 & .2976 & .3287 & .2255 & .2566 & .0585 & .4985 & .2712 & .2773 \\
        DeepSeek-V3.2 & .2667 & .2863 & .3193 & .2165 & .2336 & .0539 & .4739 & .2290 & .2599 \\
        Grok-4 & .2827 & .2955 & .3280 & .2229 & .2565 & .0637 & .5066 & .2899 & .2807 \\
        \midrule
        \multicolumn{10}{l}{\textit{REATS (Ours)}} \\
        \textbf{REATS-GRPO} & \textbf{.2482} & \textbf{.2760} & \textbf{.3103} & \textbf{.2139} & \textbf{.2280} & \textbf{.0382} & .4390 & .2711 & \textbf{.2531} \\
        \bottomrule
    \end{tabular}
    }
    \end{minipage}
    \hfill
    \begin{minipage}[t]{0.49\textwidth}
    \vspace{0pt}
    \centering
    {\small \textbf{(b) OOD to Mixed Models}}\\[2pt]
    \resizebox{\textwidth}{!}{
    \begin{tabular}{l|cccccccc|c}
        \toprule
        \textbf{Method} & \textbf{Exch} & \textbf{H1} & \textbf{H2} & \textbf{M1} & \textbf{M2} & \textbf{Wea} & \textbf{Elec} & \textbf{Traf} & \textbf{Avg} \\
        \midrule
        \multicolumn{10}{l}{\textit{Individual Models}} \\
        Model 1 & .2849 & \textbf{.2958} & 1.1093 & \textbf{.2084} & 1.1312 & .0608 & 1.0157 & \textbf{.1708} & .5346 \\
        Model 2 & .2810 & 1.0929 & \textbf{.3283} & .2516 & \textbf{.1877} & .1017 & .7105 & .8640 & .4772 \\
        Model 3 & .3845 & 1.0635 & 1.1564 & 1.1205 & 1.1374 & .0922 & \textbf{.2810} & 1.1882 & .8030 \\
        Model 4 & .9946 & .3169 & 1.1316 & 1.0863 & .3575 & .0786 & 1.0292 & 1.1745 & .7711 \\
        \midrule
        \multicolumn{10}{l}{\textit{Traditional Ensemble}} \\
        Ens$_{\text{avg}}$ & .3813 & .5373 & .7387 & .5149 & .5370 & .0775 & .6200 & .7260 & .5166 \\
        Ens$_{\text{rand}}$ & .4070 & .5713 & .7783 & .5486 & .5732 & .0797 & .6494 & .7494 & .5446 \\
        InvMSE$_{\text{tr}}$ & .2851 & .3066 & .3910 & .2172 & .2104 & .0774 & .3284 & .2068 & .2529 \\
        InvMSE$_{\text{val}}$ & .3716 & .3394 & .4822 & .2424 & .2088 & .0756 & .3560 & .2019 & .2847 \\
        RLMC & .3846 & .3161 & .3285 & .2467 & .1893 & .0741 & .2869 & \textbf{.1708} & .2446 \\
        OptW$_{\text{tr}}$ & .2784 & .3087 & .3494 & .2193 & .2074 & .0803 & .2911 & .2150 & .2437 \\
        OptW$_{\text{val}}$ & .3070 & .3186 & .3506 & .2514 & .2073 & .0726 & .2937 & .2146 & .2520 \\
        \midrule
        \multicolumn{10}{l}{\textit{LLM-based Ensemble}} \\
        GPT-5.2 & .3038 & .4679 & .6045 & .4102 & .4087 & .0654 & .4573 & .3766 & .3868 \\
        Codex & .2931 & .4431 & .5938 & .3572 & .3796 & .0637 & .4308 & .3502 & .3639 \\
        GPT-5.5 & .3263 & .4952 & .6270 & .4163 & .4429 & .0702 & .4732 & .4430 & .4117 \\
        DeepSeek-V3.2 & .2952 & .4528 & .5989 & .3299 & .3961 & .0542 & .4101 & .3426 & .3600 \\
        Grok-4 & .3101 & .4725 & .6149 & .3899 & .4458 & .0679 & .4998 & .4525 & .4067 \\
        \midrule
        \multicolumn{10}{l}{\textit{REATS (Ours)}} \\
        \textbf{REATS-GRPO} & \textbf{.2524} & .3095 & .3844 & .2085 & .1948 & \textbf{.0394} & .2866 & .2074 & \textbf{.2354} \\
        \bottomrule
    \end{tabular}
    }
    \end{minipage}

    \vspace{6pt}
    \begin{minipage}[t]{0.49\textwidth}
    \centering
    \tiny
    {\small \textbf{Model Mapping for (a)}}\\[2pt]
    \begin{tabular}{l|cccc}
        \toprule
        Dataset & Model 1 & Model 2 & Model 3 & Model 4 \\
        \midrule
        Exchange & \textbf{MOMENT} & \textbf{TimeMoE} & \textbf{TimesFM} & \textbf{MOIRAI} \\
        ETTh1 & \textbf{MOMENT} & \textbf{Sundial} & \textbf{TimesFM} & \textbf{MOIRAI} \\
        ETTh2 & \textbf{MOMENT} & \textbf{Sundial} & \textbf{TimesFM} & \textbf{MOIRAI} \\
        ETTm1 & \textbf{MOMENT} & \textbf{Sundial} & \textbf{TimesFM} & \textbf{MOIRAI} \\
        ETTm2 & \textbf{MOMENT} & \textbf{Timer} & \textbf{TimeMoE} & \textbf{MOIRAI} \\
        Weather & \textbf{MOMENT} & \textbf{Timer} & \textbf{TimeMoE} & \textbf{TimesFM} \\
        Electricity & \textbf{MOMENT} & \textbf{Sundial} & \textbf{Timer} & \textbf{MOIRAI} \\
        Traffic & \textbf{MOMENT} & \textbf{Sundial} & \textbf{Timer} & \textbf{TimeMoE} \\
        \bottomrule
    \end{tabular}
    \end{minipage}
    \hfill
    \begin{minipage}[t]{0.49\textwidth}
    \centering
    \tiny
    {\small \textbf{Model Mapping for (b)}}\\[2pt]
    \begin{tabular}{l|cccc}
        \toprule
        Dataset & Model 1 & Model 2 & Model 3 & Model 4 \\
        \midrule
        Exchange & CARD & \textbf{MOIRAI} & \textbf{MOMENT} & \textbf{ModernTCN} \\
        ETTh1 & PDF & \textbf{TimeMixer} & TimeXer & \textbf{Timer} \\
        ETTh2 & \textbf{PatchTST} & \textbf{Sundial} & \textbf{TimeMixer} & TimeXer \\
        ETTm1 & LSINet & \textbf{MOMENT} & \textbf{PatchTST} & TimeXer \\
        ETTm2 & \textbf{ModernTCN} & PDF & \textbf{PatchTST} & \textbf{TimesFM} \\
        Weather & CARD & \textbf{MLF} & \textbf{MOMENT} & \textbf{Timer} \\
        Electricity & \textbf{DLinear} & \textbf{MOMENT} & PDF & \textbf{TimeMixer} \\
        Traffic & CARD & \textbf{MOMENT} & \textbf{SEMixer} & \textbf{TimeMixer} \\
        \bottomrule
    \end{tabular}
    \end{minipage}
\end{table*}

\begin{table*}[t]
    \centering
    \caption{Generalization study within tool groups (MSE $\downarrow$). (a) Trained on \textbf{foundation model} candidates (MOIRAI, MOMENT, TimeMoE, TimesFM), 1--2 unseen per dataset. (b) Trained on \textbf{small model} candidates (CARD, LSINet, PDF, TimeXer), 1--3 unseen per dataset. \textbf{Bold} in mapping tables indicates unseen models not present during training. The MAE results are shown in Table~\ref{tab:ood_within_combined_mae}.}
    \label{tab:ood_within_combined}
    \vspace{-6pt}
    \begin{minipage}[t]{0.49\textwidth}
    \vspace{0pt}
    \centering
    {\small \textbf{(a) OOD within Foundation Models}}\\[2pt]
    \resizebox{\textwidth}{!}{
    \begin{tabular}{l|cccccccc|c}
        \toprule
        \textbf{Method} & \textbf{Exch} & \textbf{H1} & \textbf{H2} & \textbf{M1} & \textbf{M2} & \textbf{Wea} & \textbf{Elec} & \textbf{Traf} & \textbf{Avg} \\
        \midrule
        \multicolumn{10}{l}{\textit{Individual Models}} \\
        Model 1 & .3037 & .1966 & .2765 & .1124 & .1472 & .5683 & .7635 & .9884 & .4196 \\
        Model 2 & .2355 & .1742 & .2022 & .0987 & .1229 & .4397 & .3740 & .1800 & .2284 \\
        Model 3 & .2502 & .1712 & .2102 & .1112 & .2862 & .0752 & .5569 & .0831 & .2180 \\
        Model 4 & .1675 & .2161 & .2413 & .1273 & .3164 & \textbf{.0039} & .5908 & .1489 & .2265 \\
        \midrule
        \multicolumn{10}{l}{\textit{Traditional Ensemble}} \\
        Ens$_{\text{avg}}$ & .2123 & .1618 & .2070 & .0946 & .1585 & .1670 & .4765 & .1835 & .2076 \\
        Ens$_{\text{rand}}$ & .2177 & .1674 & .2119 & .0981 & .1707 & .1837 & .4957 & .2175 & .2203 \\
        InvMSE$_{\text{tr}}$ & .1989 & .1615 & .2052 & .0943 & .1339 & .1083 & .4485 & .0996 & .1813 \\
        InvMSE$_{\text{val}}$ & .2129 & .1615 & .2057 & .0943 & .1337 & .0072 & .4481 & .0979 & .1702 \\
        RLMC & .1898 & .1619 & .2011 & .0902 & .1209 & .2903 & .3740 & .0827 & .1888 \\
        OptW$_{\text{tr}}$ & .1707 & .1608 & .2006 & .0930 & .1183 & .0761 & .3797 & .0864 & .1607 \\
        OptW$_{\text{val}}$ & .1989 & .1619 & .2007 & .0927 & .1191 & .0109 & .3797 & .0863 & .1563 \\
        \midrule
        \multicolumn{10}{l}{\textit{LLM-based Ensemble}} \\
        GPT-5.2 & .1760 & .1493 & .1924 & .0884 & .1281 & .0473 & .4160 & .1114 & .1636 \\
        Codex & .1696 & .1451 & .1928 & .0861 & .1215 & .1146 & .4168 & .1052 & .1689 \\
        GPT-5.5 & .1940 & .1553 & .2003 & .0912 & .1390 & .0435 & .4351 & .1369 & .1744 \\
        DeepSeek-V3.2 & .1690 & .1464 & .1904 & .0861 & .1173 & .0538 & .4033 & .1051 & .1589 \\
        Grok-4 & .1929 & .1531 & .2005 & .0902 & .1351 & .1915 & .4443 & .1465 & .1943 \\
        \midrule
        \multicolumn{10}{l}{\textit{REATS (Ours)}} \\
        \textbf{REATS-GRPO} & \textbf{.1447} & \textbf{.1341} & \textbf{.1803} & \textbf{.0827} & \textbf{.1059} & .0030 & \textbf{.3648} & \textbf{.0810} & \textbf{.1371} \\
        \bottomrule
    \end{tabular}
    }
    \end{minipage}
    \hfill
    \begin{minipage}[t]{0.49\textwidth}
    \vspace{0pt}
    \centering
    {\small \textbf{(b) OOD within Small Models}}\\[2pt]
    \resizebox{\textwidth}{!}{
    \begin{tabular}{l|cccccccc|c}
        \toprule
        \textbf{Method} & \textbf{Exch} & \textbf{H1} & \textbf{H2} & \textbf{M1} & \textbf{M2} & \textbf{Wea} & \textbf{Elec} & \textbf{Traf} & \textbf{Avg} \\
        \midrule
        \multicolumn{10}{l}{\textit{Individual Models}} \\
        Model 1 & .1690 & .1647 & 2.0318 & .0815 & .1147 & .4372 & 1.7633 & \textbf{.0667} & .6036 \\
        Model 2 & 1.5364 & \textbf{.1558} & \textbf{.1825} & 1.9358 & 1.9402 & 1.2352 & 1.7247 & 2.1738 & 1.3605 \\
        Model 3 & .5319 & 2.0278 & 1.8930 & \textbf{.0748} & 1.9540 & .1663 & .1472 & 2.1670 & 1.1202 \\
        Model 4 & .5978 & 1.9515 & 1.9351 & 2.0260 & \textbf{.0791} & .4429 & 1.7951 & 2.1059 & 1.3667 \\
        \midrule
        \multicolumn{10}{l}{\textit{Traditional Ensemble}} \\
        Ens$_{\text{avg}}$ & .3268 & .4752 & .8083 & .4193 & .4189 & .1445 & .7556 & .8227 & .5214 \\
        Ens$_{\text{rand}}$ & .4027 & .5965 & .9526 & .5429 & .5359 & .2355 & .8761 & .9823 & .6406 \\
        InvMSE$_{\text{tr}}$ & .1903 & .1608 & .2229 & .0763 & .0843 & .1697 & .1818 & .0742 & .1450 \\
        InvMSE$_{\text{val}}$ & .2975 & .1907 & .8023 & .1208 & .0842 & .1857 & .2085 & .0728 & .2453 \\
        RLMC & .2547 & .1653 & .8372 & .0814 & 	.1129 & .1713 & .1473 & .0667 & .2253 \\
        OptW$_{\text{tr}}$ & \textbf{.1586} & .1634 & .1921 & .0832 & .0836 & .1727 & .1578 & .0851 & .1371 \\
        OptW$_{\text{val}}$ & .4002 & .1672 & .8066 & .0857 & .0840 & .1857 & .1585 & .0850 & .2466 \\
        \midrule
        \multicolumn{10}{l}{\textit{LLM-based Ensemble}} \\
        GPT-5.2 & .2719 & .2756 & .5583 & .2252 & .3101 & .0651 & .4118 & .2476 & .2957 \\
        Codex & .2625 & .2736 & .4768 & .1837 & .2936 & .0737 & .3771 & .2212 & .2703 \\
        GPT-5.5 & .2771 & .3144 & .6262 & .2890 & .3151 & .0787 & .4311 & .2994 & .3289 \\
        DeepSeek-V3.2 & .2659 & .4610 & .5092 & .2402 & .5704 & .1260 & .4059 & .2403 & .3524 \\
        Grok-4 & .2763 & .3221 & .5934 & .2237 & .3041 & .1214 & .4528 & .3253 & .3274 \\
        \midrule
        \multicolumn{10}{l}{\textit{REATS (Ours)}} \\
        \textbf{REATS-GRPO} & .1606 & \textbf{.1524} & .2324 & .0755 & .0981 & \textbf{.0446} & .1555 & .0812 & \textbf{.1250} \\
        \bottomrule
    \end{tabular}
    }
    \end{minipage}

    \vspace{6pt}
    \begin{minipage}[t]{0.49\textwidth}
    \centering
      \setlength{\tabcolsep}{6pt}
    \tiny
    {\small \textbf{Model Mapping for (a)}}\\[2pt]
    \begin{tabular}{l|cccc}
        \toprule
        Dataset & Model 1 & Model 2 & Model 3 & Model 4 \\
        \midrule
        Exchange & MOMENT & \textbf{Sundial} & \textbf{Timer} & MOIRAI \\
        ETTh1 & MOMENT & \textbf{Sundial} & TimesFM & MOIRAI \\
        ETTh2 & MOMENT & \textbf{Sundial} & \textbf{Timer} & MOIRAI \\
        ETTm1 & MOMENT & \textbf{Sundial} & TimesFM & MOIRAI \\
        ETTm2 & \textbf{Sundial} & \textbf{Timer} & TimesFM & MOIRAI \\
        Weather & MOMENT & \textbf{Timer} & TimeMoE & TimesFM \\
        Electricity & MOMENT & \textbf{Sundial} & \textbf{Timer} & MOIRAI \\
        Traffic & MOMENT & \textbf{Timer} & TimeMoE & MOIRAI \\
        \bottomrule
    \end{tabular}
    \end{minipage}
    \hfill
    \begin{minipage}[t]{0.49\textwidth}
    \centering
      \setlength{\tabcolsep}{4.5pt}
    \tiny
    {\small\textbf{Model Mapping for (b)}}\\[2pt]
    \begin{tabular}{l|cccc}
        \toprule
        Dataset & Model 1 & Model 2 & Model 3 & Model 4 \\
        \midrule
        Exchange & CARD & \textbf{DLinear} & LSINet & \textbf{SEMixer} \\
        ETTh1 & CARD & LSINet & \textbf{MLF} & \textbf{TimeMixer} \\
        ETTh2 & \textbf{ModernTCN} & PDF & \textbf{PatchTST} & TimeXer \\
        ETTm1 & LSINet & \textbf{MLF} & PDF & \textbf{PatchTST} \\
        ETTm2 & CARD & \textbf{ModernTCN} & \textbf{PatchTST} & \textbf{SEMixer} \\
        Weather & CARD & LSINet & \textbf{PatchTST} & TimeXer \\
        Electricity & \textbf{MLF} & \textbf{ModernTCN} & PDF & \textbf{PatchTST} \\
        Traffic & CARD & PDF & \textbf{SEMixer} & \textbf{TimeMixer} \\
        \bottomrule
    \end{tabular}
    \end{minipage}
\end{table*}

\begin{table*}[t]
    \centering
    \caption{Generalization study within tool groups (MAE $\downarrow$). (a) Trained on \textbf{foundation model} candidates (MOIRAI, MOMENT, TimeMoE, TimesFM), 1--2 unseen per dataset. (b) Trained on \textbf{small model} candidates (CARD, LSINet, PDF, TimeXer), 1--3 unseen per dataset. \textbf{Bold} in mapping tables indicates unseen models not present during training. Dataset abbreviations: Exch=Exchange, H1/H2=ETTh1/ETTh2, M1/M2=ETTm1/ETTm2, Wea=Weather, Elec=Electricity, Traf=Traffic.}
    \label{tab:ood_within_combined_mae}
    \vspace{-6pt}
    \begin{minipage}[t]{0.49\textwidth}
    \vspace{0pt}
    \centering
    {\small\textbf{(a) OOD within Foundation Models}}\\[2pt]
    \resizebox{\textwidth}{!}{
    \begin{tabular}{l|cccccccc|c}
        \toprule
        \textbf{Method} & \textbf{Exch} & \textbf{H1} & \textbf{H2} & \textbf{M1} & \textbf{M2} & \textbf{Wea} & \textbf{Elec} & \textbf{Traf} & \textbf{Avg} \\
        \midrule
        \multicolumn{10}{l}{\textit{Individual Models}} \\
        Model 1 & .3845 & .3428 & .3990 & .2516 & .2756 & .0922 & .7105 & .8640 & .4150 \\
        Model 2 & .3351 & .3169 & .3283 & .2304 & .2461 & .0786 & \textbf{.4383} & .3245 & .2873 \\
        Model 3 & .3433 & .3117 & .3392 & .2506 & .3575 & .0586 & .5764 & \textbf{.1894} & .3033 \\
        Model 4 & .2810 & .3476 & .3588 & .2662 & .3957 & .0431 & .5800 & .2695 & .3177 \\
        \midrule
        \multicolumn{10}{l}{\textit{Traditional Ensemble}} \\
        Ens$_{\text{avg}}$ & .3162 & .3053 & .3349 & .2305 & .2843 & .0652 & .5310 & .3530 & .3026 \\
        Ens$_{\text{rand}}$ & .3204 & .3106 & .3393 & .2343 & .2915 & .0656 & .5409 & .3657 & .3085 \\
        InvMSE$_{\text{tr}}$ & .3057 & .3050 & .3331 & .2298 & .2621 & .0603 & .5089 & .2294 & .2793 \\
        InvMSE$_{\text{val}}$ & .3167 & .3051 & .3337 & .2298 & .2620 & .0452 & .5086 & .2263 & .2784 \\
        RLMC & .3005 & .3055 & .3281 & .2228 & .2445 & .0730 & \textbf{.4383} & .1895 & .2628 \\
        OptW$_{\text{tr}}$ & .2836 & .3036 & .3274 & .2257 & .2444 & .0579 & .4466 & .2045 & .2617 \\
        OptW$_{\text{val}}$ & .3060 & .3052 & .3279 & .2259 & .2459 & .0466 & .4466 & .2039 & .2635 \\
        \midrule
        \multicolumn{10}{l}{\textit{LLM-based Ensemble}} \\
        GPT-5.2 & .2836 & .2916 & .3207 & .2210 & .2540 & .0503 & .4846 & .2518 & .2697 \\
        Codex & .2763 & .2871 & .3208 & .2170 & .2463 & .0579 & .4848 & .2445 & .2668 \\
        GPT-5.5 & .2982 & .2979 & .3282 & .2252 & .2643 & .0497 & .4986 & .2859 & .2810 \\
        DeepSeek-V3.2 & .2734 & .2870 & .3181 & .2161 & .2418 & .0491 & .4733 & .2362 & .2619 \\
        Grok-4 & .2972 & .2956 & .3287 & .2235 & .2616 & .0653 & .5066 & .3043 & .2853 \\
        \midrule
        \multicolumn{10}{l}{\textit{REATS (Ours)}} \\
        \textbf{REATS-GRPO} & \textbf{.2529} & \textbf{.2742} & \textbf{.3094} & \textbf{.2109} & \textbf{.2292} & \textbf{.0365} & .4415 & .1940 & \textbf{.2436} \\
        \bottomrule
    \end{tabular}
    }
    \end{minipage}
    \hfill
    \begin{minipage}[t]{0.49\textwidth}
    \vspace{0pt}
    \centering
    {\small\textbf{(b) OOD within Small Models}}\\[2pt]
    \resizebox{\textwidth}{!}{
    \begin{tabular}{l|cccccccc|c}
        \toprule
        \textbf{Method} & \textbf{Exch} & \textbf{H1} & \textbf{H2} & \textbf{M1} & \textbf{M2} & \textbf{Wea} & \textbf{Elec} & \textbf{Traf} & \textbf{Avg} \\
        \midrule
        \multicolumn{10}{l}{\textit{Individual Models}} \\
        Model 1 & .2849 & .3087 & 1.1625 & .2084 & .2535 & .0608 & 1.0294 & \textbf{.1708} & .4349 \\
        Model 2 & .9776 & .2992 & \textbf{.3183} & 1.1029 & 1.1312 & .1165 & 1.0166 & 1.1898 & .7690 \\
        Model 3 & .4962 & 1.1200 & 1.1093 & \textbf{.2022} & 1.1374 & .0615 & \textbf{.2810} & 1.1882 & .6995 \\
        Model 4 & .5167 & 1.0929 & 1.1316 & 1.1205 & \textbf{.1929} & .0827 & 1.0367 & 1.1745 & .7936 \\
        \midrule
        \multicolumn{10}{l}{\textit{Traditional Ensemble}} \\
        Ens$_{\text{avg}}$ & .4416 & .5454 & .7421 & .5210 & .5272 & .0742 & .6723 & .7608 & .5356 \\
        Ens$_{\text{rand}}$ & .4714 & .5817 & .7822 & .5522 & .5589 & .0764 & .7085 & .7948 & .5658 \\
        InvMSE$_{\text{tr}}$ & .3167 & .3059 & .3744 & .2045 & .2079 & .0764 & .3173 & .1952 & .2498 \\
        InvMSE$_{\text{val}}$ & .4134 & .3378 & .7392 & .2742 & .2077 & .0773 & .3426 & .1915 & .3230 \\
        RLMC & .3750 & .3087 & .7556 & .2083 & .2512 & .0759 & .2812 & .1708 & .2960 \\
        OptW$_{\text{tr}}$ & \textbf{.2831} & .3089 & .3395 & .2198 & .2110 & .0759 & .2925 & .2189 & .2437 \\
        OptW$_{\text{val}}$ & .4428 & .3132 & .7408 & .2228 & .2118 & .0772 & .2931 & .2188 & .3151 \\
        \midrule
        \multicolumn{10}{l}{\textit{LLM-based Ensemble}} \\
        GPT-5.2 & .3987 & .4037 & .6049 & .3758 & .4470 & .0549 & .4900 & .4051 & .3975 \\
        Codex & .3884 & .4035 & .5614 & .3396 & .4304 & .0548 & .4664 & .3817 & .3783 \\
        GPT-5.5 & .4070 & .4350 & .6479 & .4200 & .4503 & .0588 & .4986 & .4449 & .4203 \\
        DeepSeek-V3.2 & .3879 & .5183 & .5636 & .3739 & .5916 & .0626 & .4660 & .3716 & .4169 \\
        Grok-4 & .4042 & .4438 & .6304 & .3741 & .4388 & .0645 & .5119 & .4652 & .4166 \\
        \midrule
        \multicolumn{10}{l}{\textit{REATS (Ours)}} \\
        \textbf{REATS-GRPO} & .2847 & \textbf{.2952} & .3669 & .2025 & .2167 & \textbf{.0453} & .2900 & .2111 & \textbf{.2391} \\
        \bottomrule
    \end{tabular}
    }
    \end{minipage}

    \vspace{6pt}
    \begin{minipage}[t]{0.49\textwidth}
    \centering
      \setlength{\tabcolsep}{6pt}
    \tiny
    {\small\textbf{Model Mapping for (a)}}\\[2pt]
    \begin{tabular}{l|cccc}
        \toprule
        Dataset & Model 1 & Model 2 & Model 3 & Model 4 \\
        \midrule
        Exchange & MOMENT & \textbf{Sundial} & \textbf{Timer} & MOIRAI \\
        ETTh1 & MOMENT & \textbf{Sundial} & TimesFM & MOIRAI \\
        ETTh2 & MOMENT & \textbf{Sundial} & \textbf{Timer} & MOIRAI \\
        ETTm1 & MOMENT & \textbf{Sundial} & TimesFM & MOIRAI \\
        ETTm2 & \textbf{Sundial} & \textbf{Timer} & TimesFM & MOIRAI \\
        Weather & MOMENT & \textbf{Timer} & TimeMoE & TimesFM \\
        Electricity & MOMENT & \textbf{Sundial} & \textbf{Timer} & MOIRAI \\
        Traffic & MOMENT & \textbf{Timer} & TimeMoE & MOIRAI \\
        \bottomrule
    \end{tabular}
    \end{minipage}
    \hfill
    \begin{minipage}[t]{0.49\textwidth}
    \centering
      \setlength{\tabcolsep}{4.5pt}
    \tiny
    {\small\textbf{Model Mapping for (b)}}\\[2pt]
    \begin{tabular}{l|cccc}
        \toprule
        Dataset & Model 1 & Model 2 & Model 3 & Model 4 \\
        \midrule
        Exchange & CARD & \textbf{DLinear} & LSINet & \textbf{SEMixer} \\
        ETTh1 & CARD & LSINet & \textbf{MLF} & \textbf{TimeMixer} \\
        ETTh2 & \textbf{ModernTCN} & PDF & \textbf{PatchTST} & TimeXer \\
        ETTm1 & LSINet & \textbf{MLF} & PDF & \textbf{PatchTST} \\
        ETTm2 & CARD & \textbf{ModernTCN} & \textbf{PatchTST} & \textbf{SEMixer} \\
        Weather & CARD & LSINet & \textbf{PatchTST} & TimeXer \\
        Electricity & \textbf{MLF} & \textbf{ModernTCN} & PDF & \textbf{PatchTST} \\
        Traffic & CARD & PDF & \textbf{SEMixer} & \textbf{TimeMixer} \\
        \bottomrule
    \end{tabular}
    \end{minipage}
\end{table*}

\begin{table*}[t]
    \centering
    \caption{Scalability analysis across different numbers of candidate models (MSE $\downarrow$). All candidates are foundation models. (a) N=6, (b) N=8, (c) N=2. Dataset abbreviations: Exch=Exchange, H1/H2=ETTh1/ETTh2, M1/M2=ETTm1/ETTm2, Wea=Weather, Elec=Electricity, Traf=Traffic.}
    \label{tab:scalability_zeroshot_combined}
    \vspace{-6pt}
    \begin{minipage}[t]{0.49\textwidth}
    \vspace{0pt}
    \centering
    {\small\textbf{(a) N=6 (Chronos, MOIRAI, MOMENT, Sundial, Timer, TimesFM)}}\\[2pt]
    \resizebox{\textwidth}{!}{
    \begin{tabular}{l|cccccccc|c}
        \toprule
        \textbf{Method} & \textbf{Exch} & \textbf{H1} & \textbf{H2} & \textbf{M1} & \textbf{M2} & \textbf{Wea} & \textbf{Elec} & \textbf{Traf} & \textbf{Avg} \\
        \midrule
        \multicolumn{10}{l}{\textit{Individual Models}} \\
        Chronos & .1772 & .1796 & .2299 & .1053 & .1925 & .0162 & .6863 & .2638 & .2314 \\
        MOIRAI & .1675 & .2161 & .2413 & .1273 & .3164 & .0269 & .5908 & .1489 & .2294 \\
        MOMENT & .3037 & .1966 & .2765 & .1124 & .1707 & .5683 & .7635 & .9884 & .4225 \\
        Sundial & .2355 & .1742 & .2022 & .0987 & .1472 & .1833 & \textbf{.3740} & .0893 & .1881 \\
        Timer & .2502 & .1746 & .2102 & .1039 & .1229 & .4397 & .5569 & .1800 & .2548 \\
        TimesFM & .1845 & .1712 & .2212 & .1112 & .2862 & \textbf{.0039} & .4770 & .0864 & .1927 \\
        \midrule
        \multicolumn{10}{l}{\textit{Traditional Ensemble}} \\
        Ens$_{\text{avg}}$ & .1919 & .1595 & .2027 & .0920 & .1473 & .0972 & .4765 & .1501 & .1896 \\
        Ens$_{\text{rand}}$ & .1957 & .1630 & .2066 & .0947 & .1558 & .1104 & .4905 & .1703 & .1984 \\
        InvMSE$_{\text{val}}$ & .1926 & .1594 & .2020 & .0919 & .1326 & .0074 & .4539 & .0987 & .1673 \\
        RLMC & .1710 & .1598 & .2032 & .0908 & .1155 & .0247 & .4609 & \textbf{.0860} & .1640 \\
        InvMSE$_{\text{tr}}$ & .1829 & .1594 & .2017 & .0918 & .1326 & .0708 & .4541 & .0997 & .1741 \\
        OptW$_{\text{tr}}$ & .1649 & .1584 & .1971 & .0909 & .1162 & .0777 & .3828 & .0886 & .1596 \\
        OptW$_{\text{val}}$ & .1939 & .1595 & .2025 & .0908 & .1176 & .0164 & .3823 & .0884 & .1564 \\
        \midrule
        \multicolumn{10}{l}{\textit{LLM-based Ensemble}} \\
        GPT-5.2 & .1643 & .1468 & .1905 & .0867 & .1340 & .0288 & .4231 & .1059 & .1600 \\
        Codex & .1654 & .1468 & .1908 & .0858 & .1260 & .0579 & .4343 & .1151 & .1653 \\
        GPT-5.5 & .1676 & .1489 & .1922 & .0876 & .1345 & .0350 & .4246 & .1067 & .1621 \\
        DeepSeek-V3.2 & .1660 & .1465 & .1900 & .0850 & .1283 & .0377 & .4285 & .1160 & .1622 \\
        Grok-4 & .1665 & .1483 & .1919 & .0864 & .1279 & .0562 & .4368 & .1163 & .1663 \\
        \midrule
        \multicolumn{10}{l}{\textit{REATS (Ours)}} \\
        REATS-SFT & \textbf{.1468} & \textbf{.1346} & \textbf{.1751} & \textbf{.0809} & \textbf{.1077} & .0065 & .3769 & .0862 & \textbf{.1393} \\
        \bottomrule
    \end{tabular}
    }
    \end{minipage}
    \hfill
    \begin{minipage}[t]{0.49\textwidth}
    \vspace{0pt}
    \centering
    {\small \textbf{(b) N=8 (Chronos, MOIRAI, MOMENT, Sundial, TimeMoE, Timer, TimerXL, TimesFM)}}\\[2pt]
    \resizebox{\textwidth}{!}{
    \begin{tabular}{l|cccccccc|c}
        \toprule
        \textbf{Method} & \textbf{Exch} & \textbf{H1} & \textbf{H2} & \textbf{M1} & \textbf{M2} & \textbf{Wea} & \textbf{Elec} & \textbf{Traf} & \textbf{Avg} \\
        \midrule
        \multicolumn{10}{l}{\textit{Individual Models}} \\
        Chronos & .1772 & .1796 & .2299 & .1053 & .1925 & .0162 & .6863 & .2638 & .2314 \\
        MOIRAI & .1675 & .2161 & .2413 & .1273 & .3164 & .0269 & .5908 & .1489 & .2294 \\
        MOMENT & .3037 & .1966 & .2765 & .1124 & .1707 & .5683 & .7635 & .9884 & .4225 \\
        Sundial & .2355 & .1742 & .2022 & .0987 & .1472 & .1833 & .3740 & .0893 & .1881 \\
        TimeMoE & .2359 & .1768 & .1996 & .0997 & .1416 & .0752 & .3902 & .0831 & .1753 \\
        Timer & .2502 & .1746 & .2102 & .1039 & .1229 & .4397 & .5569 & .1800 & .2548 \\
        TimerXL & .2744 & .1782 & .2138 & .1028 & .1121 & .4343 & .4104 & .1305 & .2321 \\
        TimesFM & .1845 & .1712 & .2212 & .1112 & .2862 & \textbf{.0039} & .4770 & .0864 & .1927 \\
        \midrule
        \multicolumn{10}{l}{\textit{Traditional Ensemble}} \\
        Ens$_{\text{avg}}$ & .2040 & .1607 & .1995 & .0919 & .1324 & .1205 & .4373 & .1257 & .1840 \\
        Ens$_{\text{rand}}$ & .2067 & .1629 & .2023 & .0936 & .1384 & .1306 & .4475 & .1389 & .1901 \\
        InvMSE$_{\text{val}}$ & .2046 & .1611 & .1990 & .0921 & .1213 & .0082 & .4181 & .0916 & .1620 \\
        RLMC & .1686 & .1612 & .1997 & .0922 & .1079 & .0257 & .3727 & \textbf{.0778} & .1507 \\
        InvMSE$_{\text{tr}}$ & .1940 & .1610 & .1989 & .0921 & .1214 & .0921 & .4184 & .0926 & .1713 \\
        OptW$_{\text{tr}}$ & .1843 & .1590 & .1954 & .0915 & .1075 & .0716 & .3810 & .0838 & .1593 \\
        OptW$_{\text{val}}$ & .2033 & .1603 & .1976 & .0915 & .1082 & .0200 & .3807 & .0836 & .1556 \\
        \midrule
        \multicolumn{10}{l}{\textit{LLM-based Ensemble}} \\
        GPT-5.2 & .1830 & .1512 & .1892 & .0874 & .1183 & .0960 & .3977 & .0957 & .1648 \\
        Codex & .1844 & .1514 & .1901 & .0875 & .1176 & .0922 & .4067 & .1036 & .1667 \\
        GPT-5.5 & .1859 & .1541 & .1924 & .0893 & .1220 & .1011 & .4045 & .1001 & .1687 \\
        DeepSeek-V3.2 & .1785 & .1487 & .1864 & .0864 & .1142 & .0874 & .3954 & .0996 & .1621 \\
        Grok-4 & .1814 & .1515 & .1908 & .0869 & .1183 & .0592 & .4046 & .1008 & .1617 \\
        \midrule
        \multicolumn{10}{l}{\textit{REATS (Ours)}} \\
        REATS-SFT & \textbf{.1622} & \textbf{.1390} & \textbf{.1767} & \textbf{.0832} & \textbf{.1041} & .0139 & \textbf{.3706} & .0832 & \textbf{.1416} \\
        \bottomrule
    \end{tabular}
    }
    \end{minipage}
    \par\vspace{12pt}
    \centering
    \begin{minipage}[t]{0.55\textwidth}
    \vspace{0pt}
    \centering
    {\small\textbf{(c) N=2 (MOMENT, TimerXL)}}\\[2pt]
    \resizebox{\textwidth}{!}{
    \begin{tabular}{l|cccccccc|c}
        \toprule
        \textbf{Method} & \textbf{Exch} & \textbf{H1} & \textbf{H2} & \textbf{M1} & \textbf{M2} & \textbf{Wea} & \textbf{Elec} & \textbf{Traf} & \textbf{Avg} \\
        \midrule
        \multicolumn{10}{l}{\textit{Individual Models}} \\
        MOMENT & .3037 & .1966 & .2765 & .1124 & .1707 & .5683 & .7635 & .9884 & .4225 \\
        TimerXL & .2744 & .1782 & .2138 & .1028 & .1121 & \textbf{.4343} & .4104 & .1305 & .2321 \\
        \midrule
        \multicolumn{10}{l}{\textit{Traditional Ensemble}} \\
        Ens$_{\text{avg}}$ & .2870 & .1827 & .2226 & .1062 & .1223 & .4809 & .4887 & .3302 & .2776 \\
        Ens$_{\text{rand}}$ & .2877 & .1842 & .2301 & .1066 & .1286 & .4883 & .5215 & .4061 & .2941 \\
        InvMSE$_{\text{val}}$ & .2866 & .1823 & .2203 & .1060 & .1171 & .4735 & .4454 & .1357 & .2459 \\
        RLMC & .2745 & .1791 & .2114 & .1028 & .1116 & \textbf{.4343} & .4104 & .1305 & .2318 \\
        InvMSE$_{\text{tr}}$ & .2863 & .1823 & .2192 & .1060 & .1173 & .4766 & .4463 & .1363 & .2463 \\
        OptW$_{\text{tr}}$ & .2749 & .1782 & .2117 & .1029 & .1111 & .4393 & .4094 & .1298 & .2322 \\
        OptW$_{\text{val}}$ & .2752 & .1783 & .2125 & .1029 & .1111 & .4363 & .4094 & .1298 & .2319 \\
        \midrule
        \multicolumn{10}{l}{\textit{LLM-based Ensemble}} \\
        GPT-5.2 & .2780 & .1772 & .2068 & .1038 & .1109 & .4431 & .4193 & .1558 & .2369 \\
        Codex & .2807 & .1791 & .2120 & .1046 & .1125 & .4585 & .4287 & .1568 & .2416 \\
        GPT-5.5 & .2812 & .1795 & .2155 & .1049 & .1155 & .4487 & .4452 & .2125 & .2504 \\
        DeepSeek-V3.2 & .2786 & .1776 & .2072 & .1039 & .1097 & .4467 & .4128 & .1399 & .2346 \\
        Grok-4 & .2796 & .1788 & .2119 & .1046 & .1142 & .4402 & .4326 & .1801 & .2428 \\
        \midrule
        \multicolumn{10}{l}{\textit{REATS (Ours)}} \\
        REATS-SFT & \textbf{.2735} & \textbf{.1746} & \textbf{.2018} & \textbf{.1027} & \textbf{.1091} & .4356 & \textbf{.3990} & \textbf{.1275} & \textbf{.2280} \\
        \bottomrule
    \end{tabular}
    }
    \end{minipage}
\end{table*}

\begin{table*}[t]
    \centering
    \caption{Scalability study with foundation model candidates (MAE $\downarrow$). We evaluate REATS with varying numbers of candidate models: (a) N=6, (b) N=8, and (c) N=2. Dataset abbreviations: Exch=Exchange, H1/H2=ETTh1/ETTh2, M1/M2=ETTm1/ETTm2, Wea=Weather, Elec=Electricity, Traf=Traffic.}
    \label{tab:scalability_zeroshot_combined_mae}
    \vspace{-6pt}
    \begin{minipage}[t]{0.49\textwidth}
    \vspace{0pt}
    \centering
    {\small \textbf{(a) N=6 (Chronos, MOIRAI, MOMENT, Sundial, Timer, TimesFM)}}\\[2pt]
    \resizebox{\textwidth}{!}{
    \begin{tabular}{l|cccccccc|c}
        \toprule
        \textbf{Method} & \textbf{Exch} & \textbf{H1} & \textbf{H2} & \textbf{M1} & \textbf{M2} & \textbf{Wea} & \textbf{Elec} & \textbf{Traf} & \textbf{Avg} \\
        \midrule
        \multicolumn{10}{l}{\textit{Individual Models}} \\
        Chronos & .2986 & .3165 & .3541 & .2391 & .3095 & .0413 & .6375 & .3693 & .3207 \\
        MOIRAI & .2810 & .3476 & .3588 & .2662 & .3957 & .0584 & .5800 & .2695 & .3196 \\
        MOMENT & .3845 & .3428 & .3990 & .2516 & .3140 & .0922 & .7105 & .8640 & .4198 \\
        Sundial & .3351 & .3169 & .3283 & .2304 & .2756 & .0682 & \textbf{.4383} & .1943 & .2734 \\
        Timer & .3433 & .3169 & .3392 & .2378 & .2461 & .0786 & .5764 & .3245 & .3078 \\
        TimesFM & .2984 & .3117 & .3423 & .2506 & .3575 & .0431 & .5040 & \textbf{.1932} & .2876 \\
        \midrule
        \multicolumn{10}{l}{\textit{Traditional Ensemble}} \\
        Ens$_{\text{avg}}$ & .3022 & .3021 & .3303 & .2260 & .2776 & .0597 & .5269 & .3087 & .2917 \\
        Ens$_{\text{rand}}$ & .3052 & .3056 & .3339 & .2292 & .2838 & .0601 & .5345 & .3185 & .2963 \\
        InvMSE$_{\text{val}}$ & .3026 & .3020 & .3295 & .2256 & .2636 & .0435 & .5092 & .2209 & .2746 \\
        RLMC & .2893 & .3026 & .3308 & .2236 & .2405 & .0503 & .5057 & .1950 & .2672 \\
        InvMSE$_{\text{tr}}$ & .2956 & .3019 & .3292 & .2255 & .2637 & .0563 & .5093 & .2227 & .2755 \\
        OptW$_{\text{tr}}$ & .2818 & .3004 & .3247 & .2227 & .2424 & .0571 & .4498 & .2032 & .2603 \\
        OptW$_{\text{val}}$ & .3050 & .3022 & .3292 & .2228 & .2444 & .0440 & .4491 & .2029 & .2625 \\
        \midrule
        \multicolumn{10}{l}{\textit{LLM-based Ensemble}} \\
        GPT-5.2 & .2757 & .2879 & .3182 & .2182 & .2610 & .0480 & .4875 & .2408 & .2672 \\
        Codex & .2762 & .2879 & .3188 & .2166 & .2538 & .0519 & .4966 & .2571 & .2699 \\
        GPT-5.5 & .2780 & .2905 & .3198 & .2198 & .2622 & .0494 & .4879 & .2424 & .2687 \\
        DeepSeek-V3.2 & .2764 & .2872 & .3185 & .2154 & .2548 & .0479 & .4905 & .2507 & .2677 \\
        Grok-4 & .2780 & .2897 & .3199 & .2175 & .2560 & .0515 & .4981 & .2573 & .2710 \\
        \midrule
        \multicolumn{10}{l}{\textit{REATS (Ours)}} \\
        REATS-SFT & \textbf{.2572} & \textbf{.2739} & \textbf{.3041} & \textbf{.2088} & \textbf{.2331} & \textbf{.0373} & .4491 & .2013 & \textbf{.2456} \\
        \bottomrule
    \end{tabular}
    }
    \end{minipage}
    \hfill
    \begin{minipage}[t]{0.49\textwidth}
    \vspace{0pt}
    \centering
   {\small \textbf{(b) N=8 (Chronos, MOIRAI, MOMENT, Sundial, TimeMoE, Timer, TimerXL, TimesFM)}}\\[2pt]
    \resizebox{\textwidth}{!}{
    \begin{tabular}{l|cccccccc|c}
        \toprule
        \textbf{Method} & \textbf{Exch} & \textbf{H1} & \textbf{H2} & \textbf{M1} & \textbf{M2} & \textbf{Wea} & \textbf{Elec} & \textbf{Traf} & \textbf{Avg} \\
        \midrule
        \multicolumn{10}{l}{\textit{Individual Models}} \\
        Chronos & .2986 & .3165 & .3541 & .2391 & .3095 & \textbf{.0413} & .6375 & .3693 & .3207 \\
        MOIRAI & .2810 & .3476 & .3588 & .2662 & .3957 & .0584 & .5800 & .2695 & .3196 \\
        MOMENT & .3845 & .3428 & .3990 & .2516 & .3140 & .0922 & .7105 & .8640 & .4198 \\
        Sundial & .3351 & .3169 & .3283 & .2304 & .2756 & .0682 & .4383 & .1943 & .2734 \\
        TimeMoE & .3341 & .3199 & .3238 & .2311 & .2731 & .0586 & .4489 & .1894 & .2724 \\
        Timer & .3433 & .3169 & .3392 & .2378 & .2461 & .0786 & .5764 & .3245 & .3078 \\
        TimerXL & .3600 & .3199 & .3399 & .2364 & .2354 & .0822 & .4512 & .2408 & .2832 \\
        TimesFM & .2984 & .3117 & .3423 & .2506 & .3575 & .0431 & .5040 & .1932 & .2876 \\
        \midrule
        \multicolumn{10}{l}{\textit{Traditional Ensemble}} \\
        Ens$_{\text{avg}}$ & .3106 & .3033 & .3272 & .2249 & .2632 & .0620 & .4959 & .2738 & .2826 \\
        Ens$_{\text{rand}}$ & .3127 & .3055 & .3298 & .2269 & .2679 & .0623 & .5018 & .2813 & .2860 \\
        InvMSE$_{\text{val}}$ & .3110 & .3037 & .3266 & .2248 & .2510 & .0441 & .4790 & .2073 & .2684 \\
        RLMC & .2852 & .3038 & .3271 & .2251 & .2314 & .0539 & \textbf{.4381} & \textbf{.1802} & .2556 \\
        InvMSE$_{\text{tr}}$ & .3032 & .3035 & .3264 & .2247 & .2511 & .0589 & .4793 & .2090 & .2695 \\
        OptW$_{\text{tr}}$ & .2954 & .3010 & .3227 & .2228 & .2330 & .0562 & .4477 & .1972 & .2595 \\
        OptW$_{\text{val}}$ & .3105 & .3028 & .3239 & .2228 & .2339 & .0453 & .4472 & .1970 & .2604 \\
        \midrule
        \multicolumn{10}{l}{\textit{LLM-based Ensemble}} \\
        GPT-5.2 & .2913 & .2929 & .3170 & .2183 & .2462 & .0578 & .4633 & .2211 & .2635 \\
        Codex & .2920 & .2927 & .3181 & .2183 & .2457 & .0568 & .4721 & .2379 & .2667 \\
        GPT-5.5 & .2938 & .2960 & .3200 & .2211 & .2504 & .0592 & .4692 & .2280 & .2672 \\
        DeepSeek-V3.2 & .2872 & .2896 & .3146 & .2162 & .2399 & .0551 & .4615 & .2273 & .2614 \\
        Grok-4 & .2899 & .2928 & .3185 & .2175 & .2464 & .0532 & .4692 & .2307 & .2648 \\
        \midrule
        \multicolumn{10}{l}{\textit{REATS (Ours)}} \\
        REATS-SFT & \textbf{.2734} & \textbf{.2788} & \textbf{.3052} & \textbf{.2110} & \textbf{.2281} & .0414 & .4435 & .1979 & \textbf{.2474} \\
        \bottomrule
    \end{tabular}
    }
    \end{minipage}
    \par\vspace{12pt}
    \centering
    \begin{minipage}[t]{0.55\textwidth}
    \vspace{0pt}
    \centering
    {\small\textbf{(c) N=2 (MOMENT, TimerXL)}}\\[2pt]
    \resizebox{\textwidth}{!}{
    \begin{tabular}{l|cccccccc|c}
        \toprule
        \textbf{Method} & \textbf{Exch} & \textbf{H1} & \textbf{H2} & \textbf{M1} & \textbf{M2} & \textbf{Wea} & \textbf{Elec} & \textbf{Traf} & \textbf{Avg} \\
        \midrule
        \multicolumn{10}{l}{\textit{Individual Models}} \\
        MOMENT & .3845 & .3428 & .3990 & .2516 & .3140 & .0922 & .7105 & .8640 & .4198 \\
        TimerXL & .3600 & .3199 & .3399 & .2364 & .2354 & .0822 & \textbf{.4512} & .2408 & .2832 \\
        \midrule
        \multicolumn{10}{l}{\textit{Traditional Ensemble}} \\
        Ens$_{\text{avg}}$ & .3710 & .3270 & .3517 & .2421 & .2552 & .0867 & .5356 & .4873 & .3321 \\
        Ens$_{\text{rand}}$ & .3714 & .3284 & .3576 & .2427 & .2617 & .0869 & .5504 & .5056 & .3381 \\
        InvMSE$_{\text{val}}$ & .3706 & .3265 & .3493 & .2418 & .2471 & .0862 & .4970 & .2649 & .2979 \\
        RLMC & .3601 & .3219 & .3393 & .2364 & .2351 & .0822 & .4512 & \textbf{.2408} & .2834 \\
        InvMSE$_{\text{tr}}$ & .3704 & .3265 & .3481 & .2417 & .2475 & .0864 & .4978 & .2664 & .2981 \\
        OptW$_{\text{tr}}$ & .3604 & .3201 & .3393 & .2366 & .2349 & .0829 & .4530 & .2454 & .2841 \\
        OptW$_{\text{val}}$ & .3606 & .3203 & .3406 & .2366 & .2349 & .0825 & .4527 & .2454 & .2842 \\
        \midrule
        \multicolumn{10}{l}{\textit{LLM-based Ensemble}} \\
        GPT-5.2 & .3631 & .3207 & .3365 & .2382 & .2383 & .0833 & .4752 & .2999 & .2944 \\
        Codex & .3657 & .3229 & .3410 & .2396 & .2407 & .0846 & .4836 & .3026 & .2976 \\
        GPT-5.5 & .3656 & .3234 & .3444 & .2400 & .2453 & .0842 & .4972 & .3598 & .3075 \\
        DeepSeek-V3.2 & .3633 & .3211 & .3359 & .2384 & .2368 & .0834 & .4690 & .2706 & .2898 \\
        Grok-4 & .3644 & .3222 & .3411 & .2395 & .2434 & .0833 & .4861 & .3310 & .3014 \\
        \midrule
        \multicolumn{10}{l}{\textit{REATS (Ours)}} \\
        REATS-SFT & \textbf{.3590} & \textbf{.3179} & \textbf{.3319} & \textbf{.2363} & \textbf{.2349} & \textbf{.0820} & .4546 & .2482 & \textbf{.2831} \\
        \bottomrule
    \end{tabular}
    }
    \end{minipage}
\end{table*}

\end{document}